\PassOptionsToPackage{table}{xcolor}
\documentclass[sigconf]{acmart}

\usepackage{dblfloatfix}
\usepackage{graphicx}
\usepackage{float}
\usepackage{placeins}
\usepackage{amsmath}
\usepackage{booktabs}
\usepackage{tabularx}
\usepackage{multirow} 
\usepackage{subcaption}
\usepackage{array}
\usepackage{listings}
\usepackage{enumitem}
\usepackage[most]{tcolorbox}
\lstdefinestyle{ccocrv2prompt}{
  basicstyle=\ttfamily\footnotesize,
  breaklines=true,
  breakatwhitespace=false,
  columns=fullflexible,
  keepspaces=true,
  showstringspaces=false
}
\tcbset{
promptbox/.style={
  enhanced,
  breakable,
  colback=gray!3,
  colframe=gray!45,
  fonttitle=\bfseries,
  boxrule=0.5pt,
  arc=1mm,
  left=1mm,
  right=1mm,
  top=1mm,
  bottom=1mm
}
}
\newcolumntype{C}[1]{>{\centering\arraybackslash}m{#1}}
\usepackage[T1]{fontenc}

\usepackage[utf8]{inputenc}

\usepackage{microtype}

\usepackage{makecell}
\usepackage{pifont}

\AtBeginDocument{%
  }

\copyrightyear{2018}
\acmYear{2018}
\setcopyright{acmlicensed}
\acmDOI{XXXXXXX.XXXXXXX}
\acmConference[Conference acronym 'XX]{Make sure to enter the correct
  conference title from your rights confirmation email}{June 03--05,
  2018}{Woodstock, NY}
\acmISBN{978-1-4503-XXXX-X/2018/06}

\def\method{\textsc{CC-OCR v2}}

\begin{document}

\title{\method{}: Fine-Grained Attribution of LMM Failures in Real-World Visual Document Understanding}





\author{Chunyi Peng}
\affiliation{
  \institution{Northeastern University}
  \city{Shenyang}
  \country{China}
}
\email{hm.cypeng@stumail.neu.edu.cn}

\author{Zhipeng Xu}
\authornote{\ \ indicates project leader.}
\affiliation{
  \institution{Alibaba Group}
  \city{Hangzhou}
  \country{China}
}
\email{shifeng.xzp@alibaba-inc.com}

\author{Yuqi Xiong}
\affiliation{
  \institution{Northeastern University}
  \city{Shenyang}
  \country{China}
}
\email{xiongyuqi@mails.neu.edu.cn}

\author{Zulong Chen}
\authornotemark[1]
\affiliation{
  \institution{Alibaba Group}
  \city{Hangzhou}
  \country{China}
}
\email{zulong.czl@alibaba-inc.com}

\author{Junhao Ji}
\affiliation{
  \institution{Alibaba Group}
  \city{Hangzhou}
  \country{China}
}
\email{ jijunhao.jjh@alibaba-inc.com }

\author{Qing Liu}
\affiliation{
  \institution{Alibaba Group}
  \city{Hangzhou}
  \country{China}
}
\email{yuqing.lq@alibaba-inc.com}

\author{Zhenghao Liu}
\authornote{ \ \ indicates corresponding author.}
\affiliation{
  \institution{Northeastern University}
  \city{Shenyang}
  \country{China}
}
\email{liuzhenghao@mail.neu.edu.cn}

\author{Zubao Qin}
\affiliation{
  \institution{Alibaba Group}
  \city{Hangzhou}
  \country{China}
}
\email{qinzubao.qzb@alibaba-inc.com}

\author{Bing Zhao}
\affiliation{
  \institution{ Alibaba Group}
  \city{Hangzhou}
  \country{China}
}
\email{xiongdao@alibaba-inc.com}

\author{Jianqiang Wan}
\affiliation{
  \institution{Qwen Team, Alibaba Group}
  \city{Hangzhou}
  \country{China}
}
\email{wjq264216@alibaba-inc.com}

\author{Jun Tang}
\affiliation{
  \institution{Qwen Team, Alibaba Group}
  \city{Hangzhou}
  \country{China}
}
\email{xixing.tj@alibaba-inc.com}

\author{Zhibo Yang}
\affiliation{
  \institution{Qwen Team, Alibaba Group}
  \city{Hangzhou}
  \country{China}
}
\email{zhibo.yzb@alibaba-inc.com}

\author{Shuai Bai}
\affiliation{
  \institution{Qwen Team, Alibaba Group}
  \city{Hangzhou}
  \country{China}
}
\email{baishuai.bs@alibaba-inc.com}

\author{Dayiheng Liu}
\affiliation{
  \institution{Qwen Team, Alibaba Group}
  \city{Hangzhou}
  \country{China}
}
\email{liudayiheng.ldyh@alibaba-inc.com}

\author{Maosong Sun}
\affiliation{
  \institution{Tsinghua University}
  \city{Beijing}
  \country{China}
}
\email{sms@tsinghua.edu.cn}
\renewcommand{\shortauthors}{Chunyi Peng et al.}


\begin{abstract}
Recent Large Multimodal Models (LMMs) have achieved remarkable progress on OCR-centric document understanding and processing tasks. Existing benchmarks primarily evaluate LMMs across diverse tasks to reflect practical document-processing workflows or analyze how document characteristics influence model performance. However, they provide limited insight into the reliability of LMMs under real-world document acquisition conditions, where factors such as lighting, screen displays, imaging quality, and capture methods can substantially affect performance.
To bridge this gap, we present CC-OCR v2, a comprehensive benchmark for attributing LMM failures in real-world document processing. CC-OCR v2 provides a unified evaluation framework covering five core document-processing capabilities through 16 subtasks, 74 application scenarios, and 7,093 samples. The benchmark spans five evaluation tracks, ten document categories, and 32 languages in the recognition suite. Beyond task-level evaluation, each sample is annotated with ten fine-grained document factors, enabling systematic attribution of model failures across document types and acquisition conditions.
Extensive experiments on 17 representative LMMs reveal substantial performance variation across tasks, document categories, and real-world conditions. Moreover, models with comparable overall accuracy often exhibit fundamentally different failure patterns under specific document factors, highlighting a significant gap between benchmark-level performance and reliable deployment in practical applications.
The dataset and evaluation toolkit are publicly available at \url{https://github.com/eioss/CC-OCR-V2}.
\end{abstract}





\maketitle

\section{Introduction}
Modern Optical Character Recognition (OCR) has evolved beyond raw text transcription toward transforming visual documents into machine-readable representations~\cite{cui2021document,liu2024ocrbench,yang2025cc,ouyang2025omnidocbench} that can support downstream workflows~\cite{molina2024fetch,wang2025document}. 
As document-centered applications have grown more complex, OCR-centric systems are increasingly expected to preserve document structure, connect extracted content to visual evidence, and produce outputs that can be directly used in subsequent processing~\cite{huang2022layoutlmv3, kim2022ocr, subramani2020survey}. 
Recent Large Multimodal Models (LMMs)~\cite{bai2025qwen3,yu2025minicpm,wang2025internvl3} have significantly advanced this paradigm by enabling end-to-end document understanding within a unified architecture, reducing reliance on task-specific OCR pipelines~\cite{sun2025visrag,xiong2026lang2act,wang2025vrag}. As these models are increasingly adopted in practical document workflows, reliable benchmarking of their visual document understanding capabilities has become essential for assessing their robustness and deployment readiness.

To provide a more comprehensive assessment of LMMs for document understanding, recent research has substantially expanded OCR-centric benchmarks. Existing efforts have broadened the range of evaluated tasks and application scenarios while extending document and language coverage to more diverse layouts and acquisition settings~\cite{liu2024ocrbench,fu2024ocrbench,yang2025cc}. More recent benchmarks further incorporate physically captured documents together with synthetic or semi-synthetic perturbations to expose models to challenging conditions that are difficult to collect at scale~\cite{wang2025wilddoc}. These advances have significantly improved the breadth and realism of document evaluation.
Nevertheless, several important limitations remain. First, broader task coverage does not necessarily reflect the fundamental operations that recur in real-world document-processing workflows, as benchmark tasks are often inherited from existing datasets rather than derived from practical applications~\cite{wang2023vrdu,van2023document,wang2025document,liu2026mitigating}. Second, document types and languages encountered in real deployments remain unevenly represented, limiting the ability to evaluate model generalization across diverse documents and writing systems. Finally, most existing benchmarks summarize performance using task-level or benchmark-level metrics, overlooking the heterogeneous document factors that give rise to model failures~\cite{fu2024ocrbench,yang2025cc,ouyang2025omnidocbench,zhou2026real5}. As a result, they provide limited insight into how specific document characteristics and acquisition conditions contribute to performance degradation in practical applications~\cite{liu2024mmbench,zhang2025lmms,du2025docptbench}.

\begin{figure}[!t]        
\centering
\includegraphics[width=0.47\textwidth]{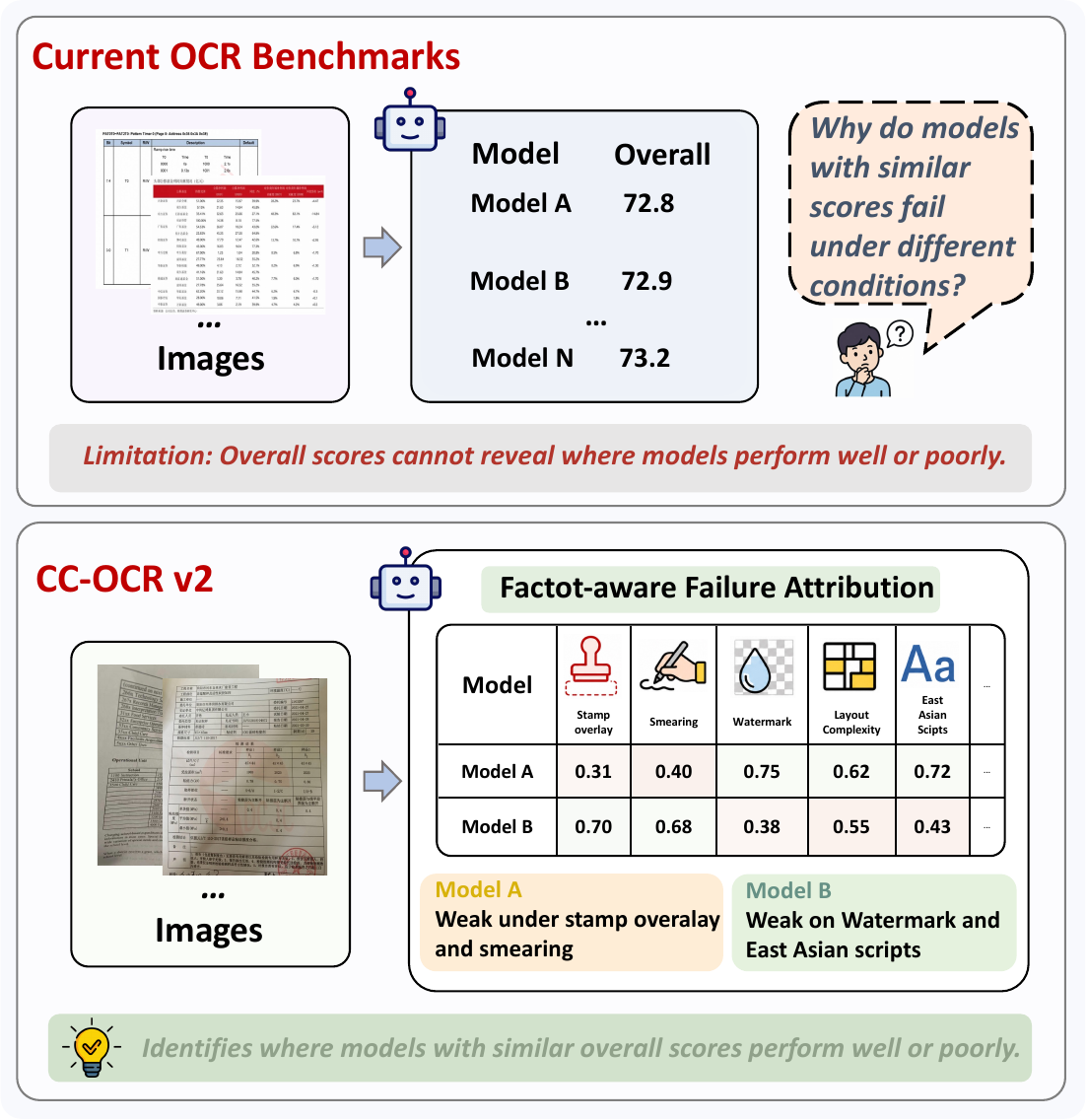}
\caption{\label{fig:intro}Comparison between current OCR benchmarks and CC-OCR v2. CC-OCR v2 is a comprehensive and diagnostic benchmark for realistic document processing.}
\label{fig:introduction}
\end{figure}

To address these limitations, we introduce \method{}, a fine-grained benchmark for attributing LMM failures in real-world OCR-centric document processing. As illustrated in Figure~\ref{fig:intro}, \method{} builds upon existing OCR benchmarks while reorganizing evaluation around practical document-processing workflows and introducing factor-aware failure attribution. The benchmark is organized around five core document-processing operations that recur in real applications, covering 16 subtasks and 74 application scenarios. It contains 7,093 samples spanning ten document types, a 32-language recognition suite, and challenging documents collected from production environments. Each sample is further annotated with ten fine-grained document factors that may influence LMM performance, including writing style, structural layout, acquisition conditions, and visual quality. These sample-level annotations enable systematic analysis of model behavior across tasks, document types, and document conditions. By leveraging these factors, \method{} identifies the document characteristics most strongly associated with each model's performance degradation, providing interpretable insights into LMM failures in OCR tasks.

Extensive experiments on 17 representative LMMs using \method{} show substantial performance variation across practical document-processing operations, with model rankings shifting considerably across different document types and acquisition conditions. Even the strongest models exhibit notable degradation on structurally complex, visually degraded, and noisy inputs, while multilingual evaluation reveals persistent performance gaps across diverse writing systems and low-resource languages. Fine-grained analysis further demonstrates that models with comparable aggregate scores can exhibit substantially different failure profiles. These findings highlight the necessity of diagnostic OCR benchmarks that move beyond aggregate performance evaluation toward systematic failure attribution under real-world conditions. By providing such analysis, \method{} offers a more informative foundation for model selection, training data construction, and capability improvement, ultimately facilitating the reliable deployment of LMMs in practical document workflows.

\begin{figure*}[t]
    \centering
    \includegraphics[width=\linewidth]{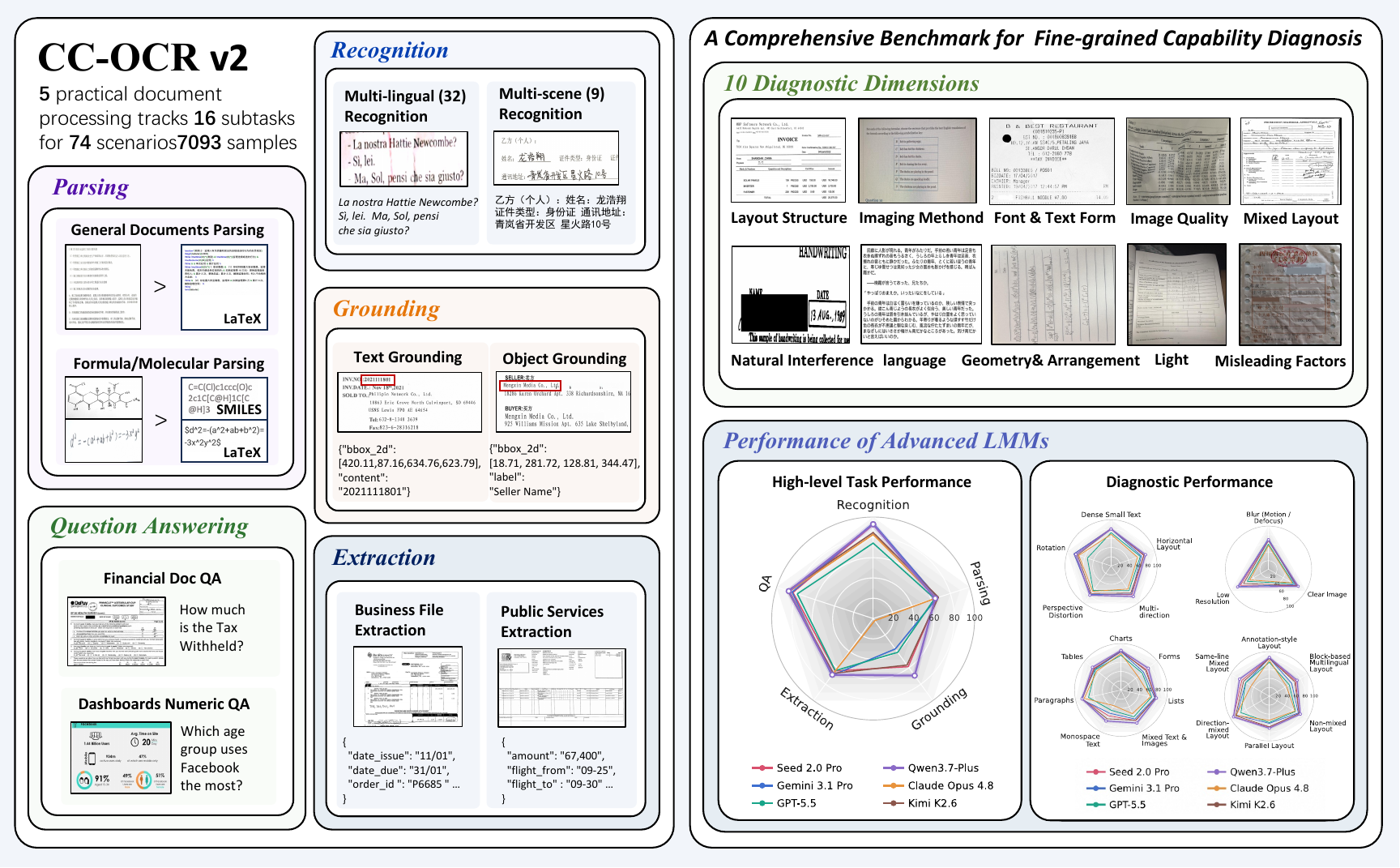}
    \caption{Overview of our \method{} benchmark.}
    \label{fig:main_figure}
\end{figure*}

\section{\method{}: Diagnosing LMM Failures in Real-World Document Processing}
In this section, we introduce the design and construction of \method{}, a diagnostic benchmark for evaluating and analyzing LMM behaviors in real-world document processing. 

As illustrated in Figure~\ref{fig:main_figure}, \method{} is organized around five fundamental OCR-centric operations commonly encountered in practical document workflows: \textit{Recognition}, \textit{Parsing}, \textit{Grounding}, \textit{Extraction}, and \textit{Question Answering}. \textit{Recognition} evaluates the ability to faithfully transcribe textual content across diverse scripts, layouts, and visual conditions. \textit{Parsing} extends recognition by assessing whether models can recover document structures, including reading order and structural relationships, and transform document content into structured formats such as LaTeX, HTML, or SMILES. \textit{Grounding} measures the capability to associate textual or semantic targets with their corresponding spatial regions in documents. Extraction evaluates the conversion of document content into schema-constrained representations, while \textit{Question Answering} assesses whether models can retrieve, integrate, and reason over document information to answer task-specific queries.

Together, these five operations cover 16 subtasks and 74 application scenarios, spanning the complete workflow of real-world document processing, from content recognition and structural understanding to information localization, extraction, and utilization. Beyond aggregate performance measurement, \method{} further provides fine-grained annotations to diagnose model behaviors and characterize failure patterns under diverse document conditions. We describe the data curation pipeline and document composition in Sec.~\ref{sec:data}, introduce the diagnostic annotations in Sec.~\ref{sec:diagnose}, and finally compare \method{} with existing OCR-centric benchmarks.

\begin{table*}[t]
\centering
\small
\setlength{\tabcolsep}{5pt}
\renewcommand{\arraystretch}{1.10}

\caption{
Comparison of representative OCR-centric benchmarks in benchmark statistics, document coverage, and analysis granularity. R, P, G, E, and Q denote Recognition, Parsing, Grounding, Extraction, and Question Answering, respectively. 
}
\label{tab:benchmark_comparison}

\begin{tabularx}{
    \textwidth
}{
    @{}
    l
    c
    c
    r
    c
    >{\raggedright\arraybackslash}X
    @{}
}
\toprule

\multirow{2}{*}{\textbf{Benchmark}}
& \multicolumn{3}{c}{\textbf{Benchmark Statistics}}
& \multicolumn{2}{c}{\textbf{Diagnostic Analysis}}
\\

\cmidrule(lr){2-4}
\cmidrule(lr){5-6}

& \textbf{Tasks}
& \textbf{\#Languages}
& \textbf{\#Scale}
& \textbf{\#Doc. Categories}
& \textbf{Analysis Granularity}
\\

\midrule

OCRBench~\cite{liu2024ocrbench}
& R/E/Q
& 2
& 1,000
& N/A
& 3 task categories
\\

OCRBench v2~\cite{fu2024ocrbench}
& R/E/Q
& 2
& 10,000
& N/A
& 3 task categories
\\

olmOCR-Bench~\cite{poznanski2025olmocr}
& P
& 1
& 1,403
& N/A
& 5 output-property classes and 7,010 unit tests
\\

OmniDocBench~\cite{ouyang2025omnidocbench}
& P
& 2
& 981
& 9
& 15 parsing attribute labels
\\

CC-OCR~\cite{yang2025cc}
& R/P/E
& 10
& 7,058
& N/A
& 3 task categories and 39 subsets
\\

\midrule

\textbf{CC-OCR v2}
& R/P/G/E/Q
& 32
& 7,093
&10
& 5 task categories, 16 subtasks, and 10 diagnostic dimensions
\\

\bottomrule 
\end{tabularx}
\end{table*}

\subsection{Data Curation and Benchmark Construction}
\label{sec:data}
\method{} is built upon CC-OCR, but is not obtained by simply appending new samples to the original benchmark. Instead, we systematically reconstruct the benchmark through data refinement, targeted expansion, and rigorous quality control to better reflect LMMs' capability in real-world document-processing scenarios.

We first conduct a comprehensive review of the original benchmark and remove instances that are unsuitable for evaluating practical document-processing capabilities. Specifically, we exclude (1) samples outside the document domain (e.g., traffic signs, video advertisements, and internet memes), which are not representative of document-processing workflows; (2) samples with annotation inconsistencies, identified through manual inspection of cases where multiple frontier LMMs consistently disagreed with the reference annotations while producing highly consistent predictions; and (3) overly simple samples that no longer provide meaningful discriminative power, as they can be reliably solved even by lightweight on-device LMMs.
To improve coverage, we further collect documents from public corpora and web sources to strengthen underrepresented tasks, document categories, and languages. \method{} is additionally enriched with representative failure cases and long-tail examples collected from production environments, capturing realistic challenges that are rarely reflected in conventional public benchmarks. All newly added and revised samples undergo task-specific annotation, multi-stage verification, disagreement adjudication, and subsequent manual inspection with model-assisted quality control to ensure annotation accuracy.

The resulting benchmark contains 7,093 evaluation samples, which is comparable with CC-OCR benchmark. Beyond task labels, each sample is annotated with document category and language metadata to facilitate fine-grained analysis across diverse evaluation conditions. \method{} covers ten document categories commonly encountered in real-world applications (Sec.~\ref{sec:diagnose}), while its multilingual recognition suite spans 32 languages covering diverse writing systems and resource conditions. Rather than reproducing the natural distribution of any particular production environment, the benchmark is designed to maximize coverage and diagnostic value by intentionally retaining representative challenging and failure-prone cases.

\subsection{Fine-grained Document Factors for Failure Attribution in Document Processing}\label{sec:diagnose}
Beyond reconstructing CC-OCR, \method{} introduces fine-grained document annotations to enable systematic attribution of LMM failures under diverse real-world document conditions.

To characterize the sources of performance variation, we manually annotate each sample with \textbf{ten document factors}, including \textit{Imaging Method}, \textit{Image Quality}, and \textit{Lighting and Screen}, which characterize document acquisition processes and visual presentation conditions. \textit{Natural Interference} and \textit{Adversarial or Misleading Factors} capture visual elements that may obstruct or mislead model predictions. \textit{Font and Text Form} describes textual characteristics, while \textit{Mixed Layout}, \textit{Geometry and Arrangement}, and \textit{Layout Structure} characterize the spatial organization of document content. Since \method{} provides broad multilingual coverage, it uses the \textit{Language} factor to help benchmark LMMs in different language settings. Since multiple factors may co-occur within the same document, directly comparing category-level performance can confound the effects of correlated document characteristics. We therefore estimate the contribution of each factor while controlling for task, document category, and other document factors, enabling more comprehensive benchmarking and analysis of LMM behaviors. 

\subsection{Benchmark Comparison}
\label{sec:comparison}
Table~\ref{tab:benchmark_comparison} compares \method{} with representative OCR-centric benchmarks in terms of benchmark statistics and diagnostic analysis. The scale denotes the number of evaluation queries, whereas analysis granularity describes how benchmark results can be decomposed beyond an overall score.

OCRBench~\cite{liu2024ocrbench} and OCRBench v2~\cite{fu2024ocrbench} both cover recognition, extraction, and question answering in two languages, with 1,000 and 10,000 evaluation queries, respectively. Their results are primarily reported over three task categories. olmOCR-Bench~\cite{poznanski2025olmocr} focuses exclusively on document parsing, containing 1,403 evaluation queries in one language and decomposing performance into five output-property classes through 7,010 unit tests. OmniDocBench~\cite{ouyang2025omnidocbench} also focuses on parsing, with 981 samples in two languages across nine document categories, and provides 15 parsing attribute labels for fine-grained analysis. CC-OCR~\cite{yang2025cc} covers recognition, parsing, and extraction over 7,058 samples in ten languages, with results organized into three task categories and 39 predefined subsets. However, these subsets do not constitute a unified sample-level diagnostic taxonomy that can be consistently applied across tasks.

In comparison, \method{} is the only benchmark in Table~\ref{tab:benchmark_comparison} that jointly covers all five OCR-centric document-processing operations: Recognition, Parsing, Grounding, Extraction, and Question Answering. It contains 7,093 samples spanning 32 languages and ten real-world document categories. Its evaluation results can be examined at three complementary levels: five task categories, 16 subtasks, and ten sample-level diagnostic dimensions.

\begin{table*}[ht]
\centering
\caption{Overall performance of representative LMMs across five OCR-centric document-processing operations.}
\label{tab:benchmark_results}
\begin{tabularx}{\textwidth}{l *{6}{>{\centering\arraybackslash}X}}
\toprule
\multirow{2}{*}{\textbf{Model}}
& \multicolumn{5}{c}{\textbf{Document Processing Tasks}}
& \multirow{2}{*}{\textbf{Average}} \\
\cmidrule(lr){2-6}
& Recognition
& Parsing
& Grounding
& Extraction
& QA
& \\
\midrule

\multicolumn{7}{l}{\textit{\textbf{On-Device LMMs}}} \\
\midrule
\raisebox{-0.2\height}{\includegraphics[height=1.2em]{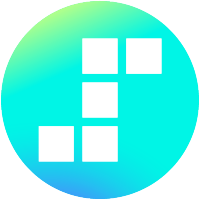}}\,Step3-VL-10B
& 40.95 & 29.36 & 3.37 & 56.50 & 74.60 & 40.96 \\

\raisebox{-0.2\height}{\includegraphics[height=1.2em]{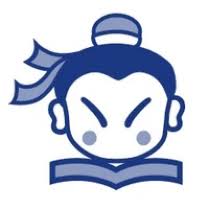}}\,InternVL3.5-8B
& 46.36 & 53.99 & 11.37 & 54.75 & 75.04 & 48.30 \\

\raisebox{-0.2\height}{\includegraphics[height=1.2em]{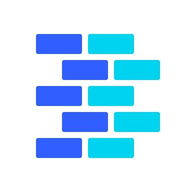}}\,MiniCPM-o 4.5-8B
& 61.09 & 50.91 & 13.75 & 50.68 & 83.41 & 51.97 \\

\raisebox{-0.2\height}{\includegraphics[height=1.2em]{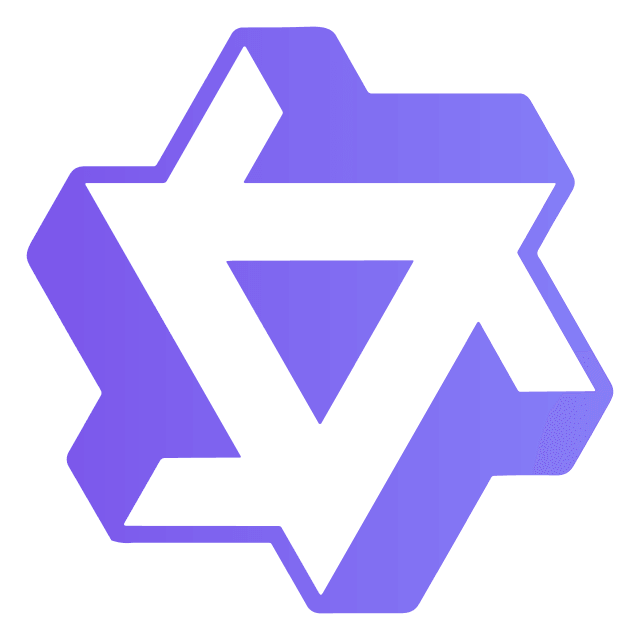}}\,Qwen3.5-9B
& 83.89 & 58.62 & 47.06 & 62.55 & 84.64 & 67.35 \\

\midrule
\multicolumn{7}{l}{\textit{\textbf{On-Server LMMs}}} \\
\midrule
\raisebox{-0.2\height}{\includegraphics[height=1.2em]{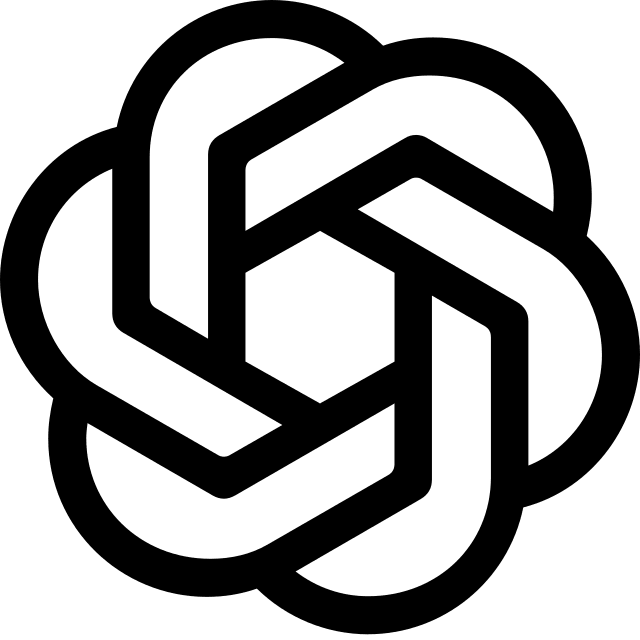}}\,GPT-5.4
& 72.35 & 62.59 & 11.76 & 57.94 & 78.82 & 56.69 \\

\raisebox{-0.2\height}{\includegraphics[height=1.2em]{logo/qwen.png}}\,Qwen-VL-Max
& 81.77 & 54.92 & 3.70 & 64.40 & 83.33 & 57.62 \\

\raisebox{-0.2\height}{\includegraphics[height=1.2em]{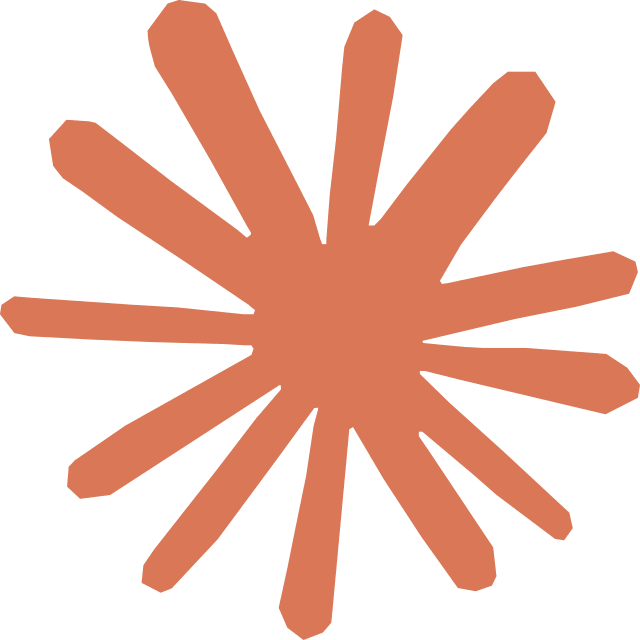}}\,Claude Opus 4.6
& 80.74 & 58.69 & 7.69 & 63.00 & 84.09 & 58.84 \\

\raisebox{-0.2\height}{\includegraphics[height=1.2em]{logo/claude.png}}\,Claude Sonnet 4.6
& 82.90 & 61.84 & 7.87 & 63.02 & 83.52 & 59.83 \\

\raisebox{-0.2\height}{\includegraphics[height=1.2em]{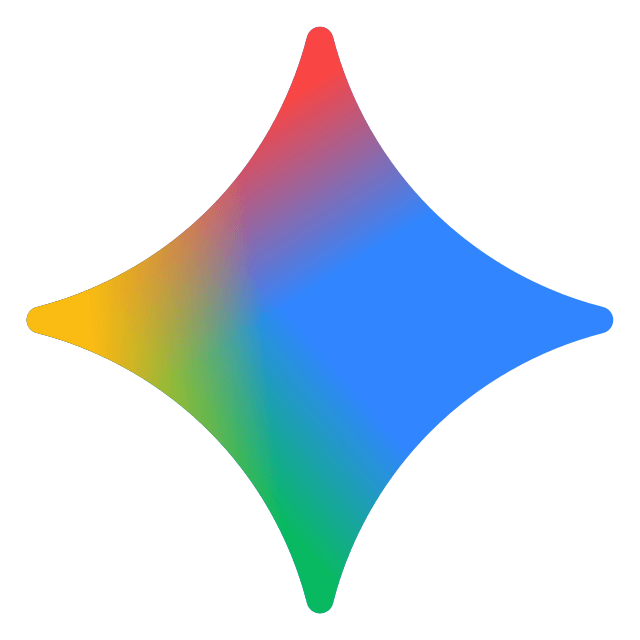}}\,Gemini 3.1 Flash
& \underline{93.61} & 64.67 & 7.33 & 62.17 & 79.41 & 61.44 \\

\raisebox{-0.2\height}{\includegraphics[height=1.2em]{logo/openai.png}}\,GPT-5.5
& 74.30 & 63.67 & 41.15 & 63.46 & 78.62 & 64.24 \\

\raisebox{-0.2\height}{\includegraphics[height=1.2em]{logo/claude.png}}\,Claude Opus 4.8
& 82.89 & 65.39 & 2.59 & 67.10 & 86.75 & 65.39 \\

\raisebox{-0.2\height}{\includegraphics[height=1.2em]{logo/gemini.png}}\,Gemini 3.1 Pro
& \textbf{93.99} & 66.51 & 37.07 & \textbf{69.38} & 83.98 & 70.19 \\

\raisebox{-0.2\height}{\includegraphics[height=1.2em]{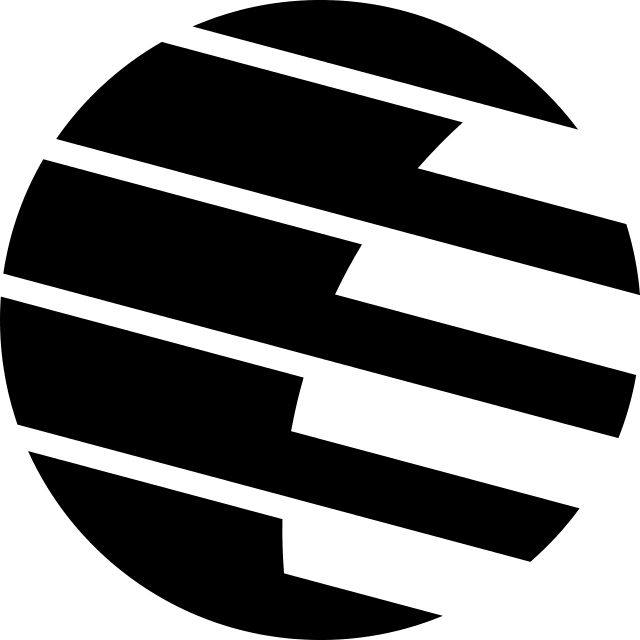}}\,Kimi K2.6
& 84.74 & \underline{67.61} & 56.70 & 66.73 & \underline{88.50} & 72.86 \\

\raisebox{-0.2\height}{\includegraphics[height=1.2em]{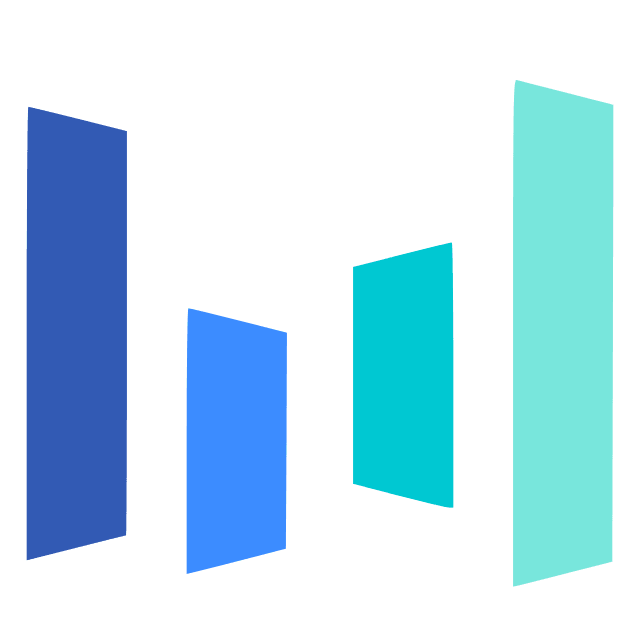}}\,Seed 2.0 Pro
& 92.48 & 66.78 & 58.67 & 63.80 & 83.11 & 72.97 \\

\raisebox{-0.2\height}{\includegraphics[height=1.2em]{logo/kimi.png}}\,Kimi K2.5
& 87.26 & \textbf{67.94} & 55.64 & 67.13 & \textbf{88.74} & 73.34 \\


\raisebox{-0.2\height}{\includegraphics[height=1.2em]{logo/qwen.png}}\,Qwen3.6-Plus
& 92.31 & 64.97 & \underline{69.37} & 68.48 & 87.34 & \underline{76.50} \\

\raisebox{-0.2\height}{\includegraphics[height=1.2em]{logo/qwen.png}}\,Qwen3.7-Plus
& 92.93 & 64.32 & \textbf{69.57} & \underline{68.53} & 87.49 & \textbf{76.57} \\

\bottomrule
\end{tabularx}
\end{table*}

\section{Experimental Methodology}
In this section, we describe the evaluated models, evaluation metrics, and implementation details adopted in our experiments.

\textbf{Evaluated Models.}
We evaluate 17 representative LMMs under both on-device and on-server settings. The on-device models include Qwen3.5-9B~\cite{bai2025qwen3}, MiniCPM-o 4.5-8B~\cite{yu2025minicpm}, InternVL3.5-8B~\cite{wang2025internvl3}, and Step3-VL-10B~\cite{huang2026step3}. The on-server models cover representative model families including GPT, Qwen, Claude, Gemini, Kimi, and Seed. Detailed model versions, identifiers, and deployment configurations are provided in Appendix~\ref{app:config}.

\textbf{Evaluation Metrics.}

Building on prior work, we employ task-specific metrics for the five document-processing tracks. Recognition is evaluated using full-text multi-set micro-F1~\cite{yang2025cc}. For parsing, we use normalized edit similarity~\cite{levenshtein1966binary} for general documents, formulas, and molecular structures, and TEDS~\cite{zhong2020image} for complex tables; information-board parsing combines
character-level micro-F1 with TEDS.

Grounding is evaluated using IoU following established visual grounding protocols~\cite{yu2016modeling}, with macro-averaged mIoU as the primary score and Acc@0.5 as an additional metric. Extraction uses normalized field-level micro-F1 following prior KIE work~\cite{kim2022ocr,yang2025cc}. For question answering, we follow document VQA evaluation~\cite{mathew2021docvqa}, using normalized Levenshtein similarity for long-form answers and containment-based exact matching for short answers. Full definitions and normalization procedures are provided in Appendix~\ref{app:metrics}

\textbf{Implementation Details.}
All models are evaluated using the same task-specific prompts and deterministic decoding with the temperature set to 0. On-server models are accessed through their official APIs, while on-device models are deployed using vLLM with FlashAttention acceleration. Full prompts and additional implementation details are provided in Appendix~\ref{app:prompt}.

\begin{table*}[t]
\centering
\caption{Performance of representative LMMs across ten real-world document categories.}
\label{tab:doc_category}
\resizebox{\textwidth}{!}{
\begin{tabular}{
ll
*{10}{c}
}
\toprule
\multirow{2}{*}{\textbf{Model}}
& \multicolumn{10}{c}{\textbf{Document Categories}}
& \multirow{2}{*}{\textbf{Average}} \\
\cmidrule(lr){2-11}
& Books
& Reports
& Marketing
& Legal
& Forms
& Receipts
& Medical
& Handwrit.
& Charts
& Screens
& \\
\midrule

\multicolumn{12}{l}{\textit{\textbf{On-Device LMMs}}} \\
\midrule
\raisebox{-0.2\height}{\includegraphics[height=1.2em]{logo/stepfun.png}}\,Step3-VL-10B
& 54.65 & 68.82 & 73.03 & 67.32 & 51.66 & 37.26 & 59.51 & 42.11 & 62.29 & 63.63 & 58.03 \\

\raisebox{-0.2\height}{\includegraphics[height=1.2em]{logo/shailab.png}}\,InternVL3.5-8B
& 62.68 & 81.45 & 74.19 & 72.17 & 53.34 & 39.70 & 51.69 & 55.43 & 65.52 & 65.04 & 62.12 \\

\raisebox{-0.2\height}{\includegraphics[height=1.2em]{logo/minicpm.png}}\,MiniCPM-o 4.5-8B
& 73.38 & 79.24 & 81.48 & 71.52 & 51.66 & 35.85 & 57.24 & 55.61 & 71.11 & 46.72 & 62.38 \\

\raisebox{-0.2\height}{\includegraphics[height=1.2em]{logo/qwen.png}}\,Qwen3.5-9B
& 82.66 & 79.46 & 82.05 & 78.52 & 68.27 & 61.80 & 78.62 & 63.72 & 77.84 & 72.47 & 74.54 \\

\midrule
\multicolumn{12}{l}{\textit{\textbf{On-Server LMMs}}} \\
\midrule
\raisebox{-0.2\height}{\includegraphics[height=1.2em]{logo/openai.png}}\,GPT-5.4
& 73.31 & 81.30 & 75.09 & 72.65 & 53.51 & 43.19 & 39.58 & 62.37 & 76.12 & 77.72 & 65.48 \\

\raisebox{-0.2\height}{\includegraphics[height=1.2em]{logo/qwen.png}}\,Qwen-VL-Max
& 80.54 & 81.25 & 78.91 & 71.97 & 53.37 & 39.10 & 42.18 & 63.72 & 77.06 & 71.21 & 65.93 \\

\raisebox{-0.2\height}{\includegraphics[height=1.2em]{logo/claude.png}}\,Claude Sonnet 4.6
& 80.09 & 81.10 & 78.53 & 73.76 & 54.57 & 42.57 & 52.72 & 61.60 & 77.84 & 79.42 & 68.22 \\

\raisebox{-0.2\height}{\includegraphics[height=1.2em]{logo/claude.png}}\,Claude Opus 4.6
& 79.32 & 82.66 & 79.60 & 75.97 & 54.71 & 41.46 & 58.76 & 57.08 & 74.76 & 79.27 & 68.36 \\

\raisebox{-0.2\height}{\includegraphics[height=1.2em]{logo/claude.png}}\,Claude Opus 4.8
& 82.06 & 84.68 & 84.04 & 75.47 & 52.27 & 42.48 & 37.59 & 67.27 & 80.37 & 84.49 & 69.10 \\

\raisebox{-0.2\height}{\includegraphics[height=1.2em]{logo/openai.png}}\,GPT-5.5
& 73.22 & 80.80 & 76.34 & 72.39 & 68.93 & 53.74 & 46.92 & 63.13 & 74.65 & \underline{86.02} & 69.61 \\

\raisebox{-0.2\height}{\includegraphics[height=1.2em]{logo/gemini.png}}\,Gemini 3.1 Flash
& \textbf{87.89} & 81.60 & 77.33 & 76.43 & 57.19 & 44.48 & 71.71 & 67.33 & 73.29 & 70.18 & 70.74 \\

\raisebox{-0.2\height}{\includegraphics[height=1.2em]{logo/kimi.png}}\,Kimi K2.5
& 86.61 & \textbf{86.26} & \textbf{87.05} & 77.05 & 55.99 & 44.30 & 56.52 & 71.06 & \textbf{85.88} & 80.59 & 73.13 \\

\raisebox{-0.2\height}{\includegraphics[height=1.2em]{logo/kimi.png}}\,Kimi K2.6
& 85.34 & \underline{85.42} & \underline{86.52} & 78.44 & 60.68 & 48.57 & 56.22 & 70.34 & \underline{84.54} & 78.19 & 73.43 \\

\raisebox{-0.2\height}{\includegraphics[height=1.2em]{logo/gemini.png}}\,Gemini 3.1 Pro
& \underline{87.77} & 83.05 & 81.61 & 78.66 & 66.94 & 58.12 & \underline{79.22} & \underline{72.57} & 80.38 & 84.87 & 77.32 \\

\raisebox{-0.2\height}{\includegraphics[height=1.2em]{logo/bytedance.png}}\,Seed 2.0 Pro
& 87.38 & 84.39 & 81.11 & 79.01 & 74.31 & 61.61 & 74.21 & \textbf{73.11} & 80.38 & 80.70 & 77.62 \\


\raisebox{-0.2\height}{\includegraphics[height=1.2em]{logo/qwen.png}}\,Qwen3.6-Plus
& 86.36 & 83.09 & 84.55 & \underline{81.71} & \underline{80.01} & \underline{67.92} & 73.89 & 67.25 & 83.30 & 82.20 & \underline{79.03} \\

\raisebox{-0.2\height}{\includegraphics[height=1.2em]{logo/qwen.png}}\,Qwen3.7-Plus
& 86.37 & 84.98 & 84.33 & \textbf{82.87} & \textbf{80.30} & \textbf{72.43} & \textbf{82.94} & 68.58 & 82.21 & \textbf{86.04} & \textbf{81.11} \\

\bottomrule
\end{tabular}
}
\end{table*}
\section{Evaluation Results}
This section presents a comprehensive evaluation of current LMMs using \method{}, providing a systematic analysis of their capability boundaries and failure patterns.

\subsection{Overall Performance}

As shown in Table~\ref{tab:benchmark_results}, we compare 17 representative LMMs across five OCR-centric document-processing operations on \method{}.

The results show that frontier on-server models achieve the strongest overall performance, with Qwen3.7-Plus and Qwen3.6-Plus obtaining nearly identical average scores of 76.57 and 76.50, respectively. These results indicate that \method{} remains sufficiently challenging to distinguish the capabilities of current advanced LMMs. Among on-device models, Qwen3.5-9B achieves substantially stronger performance than other compact models and even surpasses several on-server systems, demonstrating the rapid progress of efficient LMMs for document processing.

The five tasks in \method{} exhibit substantially different levels of difficulty. Recognition and Question Answering represent relatively well-established capabilities, whereas Parsing, Extraction, and Grounding remain considerably more challenging, as they require models to not only identify document content across diverse real-world formats but also preserve structural relationships and associate information with specific visual evidence. Notably, several systems that achieve strong performance in Recognition and Question Answering obtain substantially lower scores on Grounding, while the strongest Qwen models demonstrate more consistent visual localization capabilities. This gap indicates that accurate document reading alone does not necessarily imply the ability to link extracted information to precise and verifiable regions within the document.
No single model consistently achieves superior performance across all tasks. For example, Gemini 3.1 Pro obtains the best results in Recognition and Extraction, Kimi K2.5 leads in Parsing and Question Answering, while Qwen3.7-Plus achieves the strongest performance in Grounding as well as overall evaluation. Moreover, similar aggregate scores can conceal substantially different capability profiles. For instance, Seed 2.0 Pro and Kimi K2.6 achieve comparable overall performance, yet Seed 2.0 Pro is stronger in Recognition, whereas Kimi K2.6 exhibits advantages in Extraction and Question Answering. These observations suggest that aggregate metrics provide only a coarse characterization of LMM capabilities and fail to reveal model-specific strengths, limitations, and suitability for different document-processing scenarios.

\subsection{Document Processing Capability across Different Categories}
To investigate whether model capabilities remain consistent across practical document scenarios, we evaluate different LMMs across diverse document categories.

As shown in Table~\ref{tab:doc_category}, models generally achieve more reliable performance on books, reports, and marketing materials, where textual content and page layouts tend to be relatively regular. In contrast, performance degrades substantially on more challenging document categories, such as handwritten documents and receipts, where dense information, heterogeneous structures, and degraded or ambiguous visual content impose greater demands on visual perception and structural understanding.

Across different LMMs, performance also varies substantially across document categories. For example, Kimi K2.5 performs particularly well on reports, marketing materials, and charts, but exhibits considerably weaker performance on receipts. Gemini 3.1 Flash shows a similar pattern, achieving strong results on books while struggling with receipts. Qwen3.7-Plus demonstrates stronger competitiveness on structured and application-oriented categories, including legal documents, forms, receipts, and screens, whereas Seed 2.0 Pro shows a clear advantage on handwritten documents. Notably, these category-level gaps often exceed the differences suggested by aggregate scores, revealing that model rankings are sensitive to document categories. This instability is especially important for deployment, where the target document distribution may differ substantially from the benchmark average, indicating that high overall performance does not imply universal applicability, and different models may possess distinct capability profiles tailored to specific document scenarios.

\begin{figure*}[t]
    \centering
    \captionsetup[subfigure]{font=scriptsize,skip=2pt}

    \makebox[\textwidth][c]{%
    \begin{subfigure}[t]{0.245\textwidth}
        \centering
        \includegraphics[width=1.07\linewidth,trim=12pt 2pt 12pt 2pt,clip]{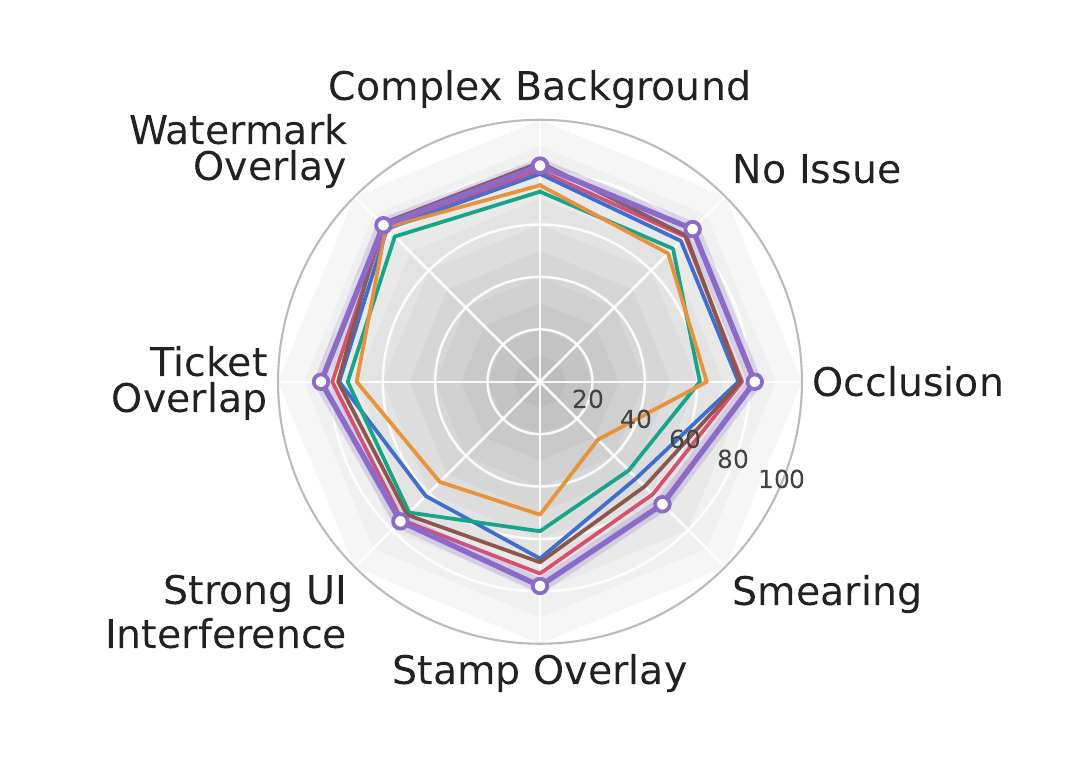}
        \caption{Natural Interference.}
        \label{fig:radar_a}
    \end{subfigure}%
    \hspace{-0.004\textwidth}%
    \begin{subfigure}[t]{0.245\textwidth}
        \centering
        \includegraphics[width=1.07\linewidth,trim=12pt 2pt 12pt 2pt,clip]{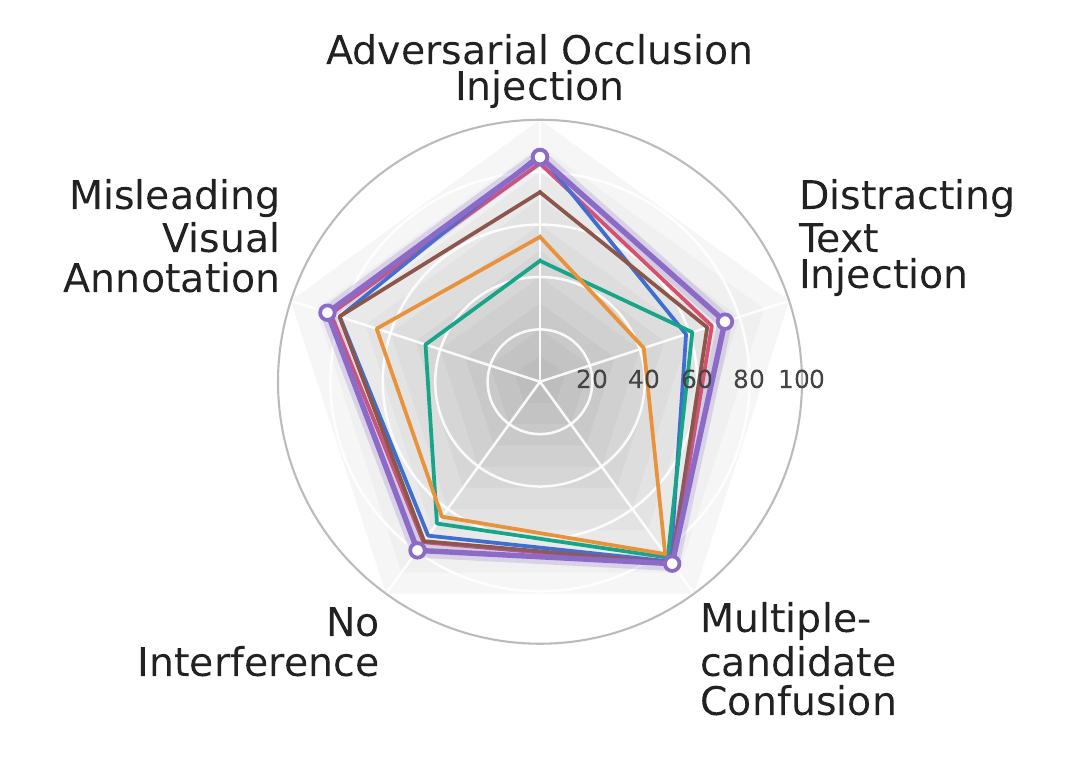}
        \caption{Adversarial Factors.}
        \label{fig:radar_b}
    \end{subfigure}%
    \hspace{-0.004\textwidth}%
    \begin{subfigure}[t]{0.245\textwidth}
        \centering
        \includegraphics[width=1.07\linewidth,trim=12pt 2pt 12pt 2pt,clip]{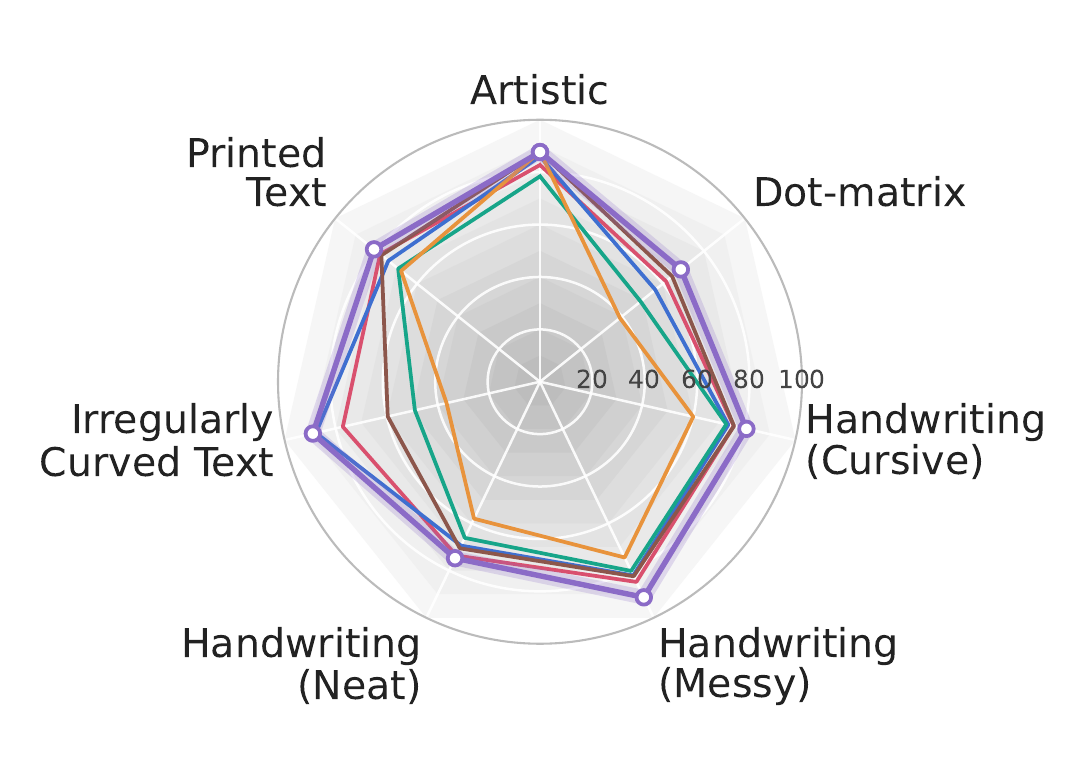}
        \caption{Font and Text Form.}
        \label{fig:radar_c}
    \end{subfigure}%
    \hspace{-0.004\textwidth}%
    \begin{subfigure}[t]{0.245\textwidth}
        \centering
        \includegraphics[width=1.07\linewidth,trim=12pt 2pt 12pt 2pt,clip]{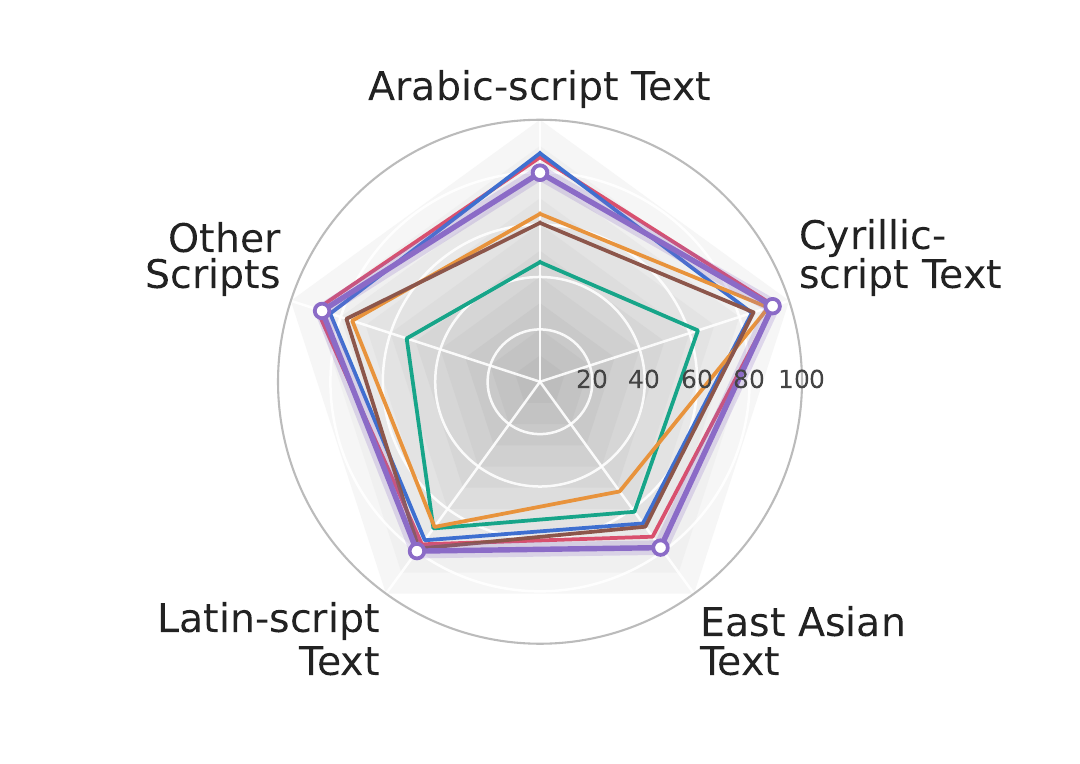}
        \caption{Language.}
        \label{fig:radar_d}
    \end{subfigure}%
    }

    \vspace{1pt}
    \includegraphics[width=0.72\textwidth]{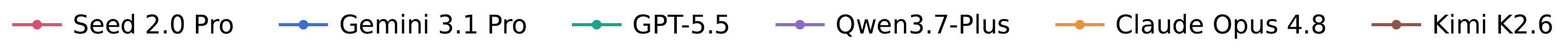}
    \vspace{-2pt}

    \caption{Fine-grained failure attribution of representative LMMs across four document dimensions in CC-OCR v2.}
    \label{fig:attribute}
\end{figure*}

\subsection{Fine-grained LMM Failure Attribution for Real-World Document Processing}
\label{sec:attribution}

In this section, we use the ten diagnostic dimensions defined in
Sec.~\ref{sec:diagnose} to analyze and attribute LMM failures under
diverse real-world document conditions. Figure~\ref{fig:attribute}
presents four representative dimensions spanning natural interference,
adversarial and misleading factors, text appearance, and language
variation, revealing distinct failure patterns across models.
Complete performance results across all ten dimensions are provided in
Appendix~\ref{app:attribute}, while their category-wise distributions
are reported in Appendix~\ref{app:statistics}.

Our analysis shows that models with similar
overall performance can exhibit substantially different fine-grained
robustness profiles, highlighting the necessity of diagnostic evaluation
beyond aggregate metrics.
As shown in Figures~\ref{fig:radar_a} and~\ref{fig:radar_b}, model robustness varies substantially across different interference scenarios. Qwen3.7-Plus demonstrates relatively consistent robustness across most natural interference categories, whereas Claude Opus 4.8 exhibits notable performance degradation under smearing and stamp overlays. Under adversarial perturbations, Gemini 3.1 Pro shows strong resilience against adversarial occlusion injection, while Seed 2.0 Pro and Qwen3.7-Plus remain robust under distracting text and misleading visual annotations. These results reveal fine-grained robustness differences among LMMs that are not captured by existing benchmarks that do not provide fine-grained attributions.

Figure~\ref{fig:radar_c} further analyzes text appearance, revealing substantial model-specific variations across different text rendering styles. Models achieve relatively comparable performance on artistic and neatly handwritten text, but exhibit larger discrepancies on more challenging forms, including dot-matrix text, irregularly curved text, and messy or cursive handwriting. Qwen3.7-Plus and Seed 2.0 Pro maintain relatively strong robustness across most text forms, whereas Claude Opus 4.8 shows greater sensitivity to dot-matrix and curved text. These results suggest that recognition capabilities do not generalize uniformly across different character renderings and spatial arrangements.

Figure~\ref{fig:radar_d} further reveals pronounced script-dependent performance differences. Gemini 3.1 Pro achieves particularly strong performance on Arabic-script text, while Seed 2.0 Pro and Qwen3.7-Plus exhibit more balanced robustness across diverse script groups. In contrast, GPT-5.5 experiences larger performance degradation on several non-Latin scripts, and Claude Opus 4.8 shows weaker performance on East Asian text. These findings demonstrate that multilingual document understanding cannot be adequately characterized by a single aggregate OCR score, as models exhibit distinct strengths and weaknesses across language scripts. Detailed results for all 32 languages are reported in Appendix~\ref{app:multilingual}.

\begin{figure*}
    \centering
    \includegraphics[width=\linewidth]{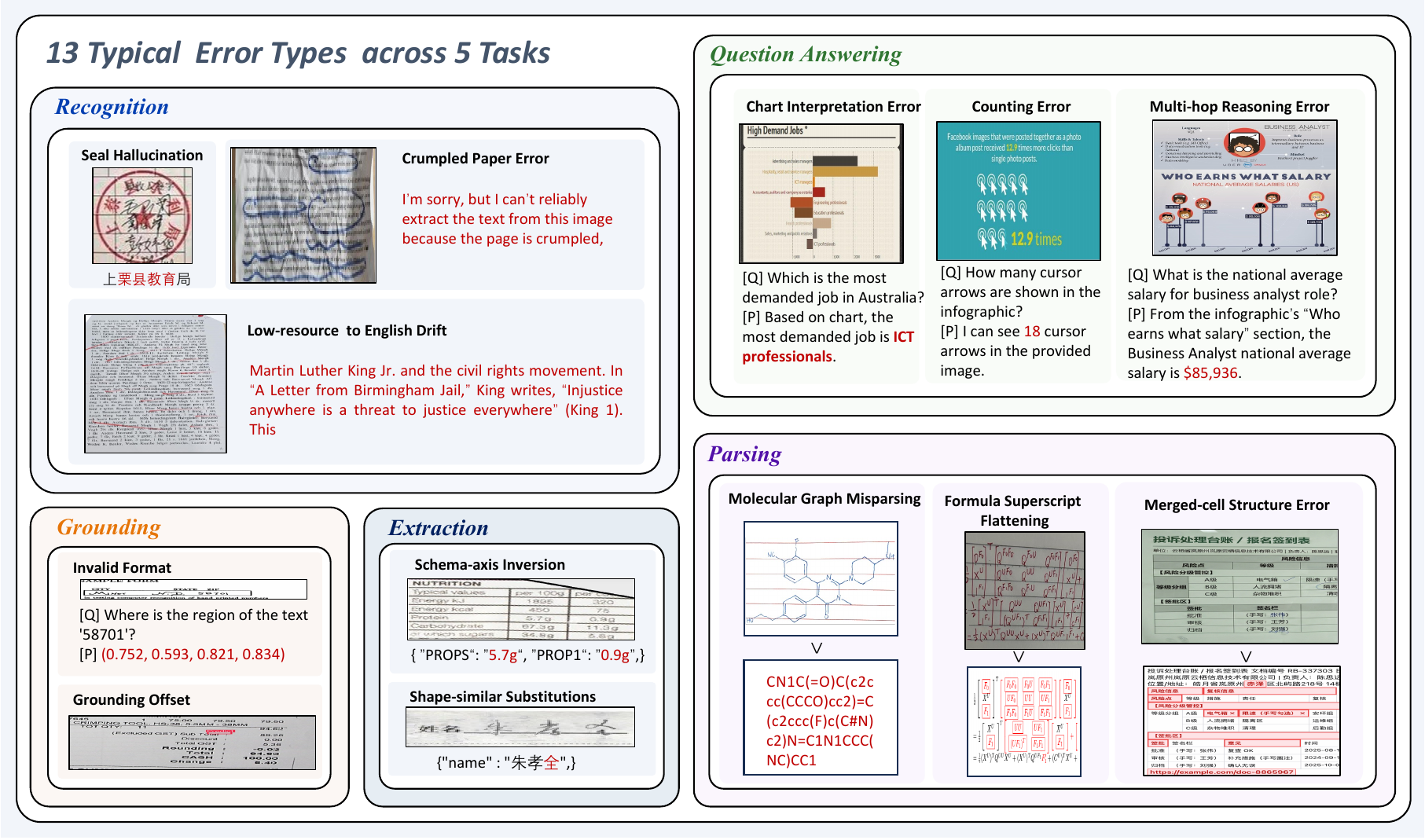}
\caption{Case studies. We present representative examples of 13 typical error types observed in LMMs across the five tracks.}
\label{fig:badcase}
\vspace{-0.1in}
\end{figure*}

\subsection{Case Studies of LMM Failures in \method{}}
Figure~\ref{fig:badcase} summarizes 13 representative failure types across the five visual document processing tracks, demonstrating that LMM errors emerge in different document understanding tasks.

Extraction and Grounding failures primarily arise from inaccurate mappings between recognized content and the required output representations. Schema-axis Inversion associates correctly identified values with incorrect fields, whereas Shape-similar Substitutions generate erroneous outputs due to visually similar characters. Similarly, Invalid Format occurs when models identify the target region but return coordinates in an incorrect format, while Grounding Offset produces valid but spatially misaligned bounding boxes. Together, these failures reveal persistent weaknesses in field-value association and precise spatial localization.

Question Answering introduces additional challenges beyond local recognition and extraction, requiring models to integrate evidence across visual regions and perform document-level reasoning. Overall, these failure cases cover multiple capability dimensions, including visual perception, structural reconstruction, spatial grounding, field association, and multi-step reasoning. Additional examples and model-wise comparisons are provided in Appendix~\ref{app:failure_analysis}.

In Recognition, Seal Hallucination reflects the tendency of models to hallucinate plausible but incorrect text when interpreting ambiguous seals, while Low-resource-to-English Drift shows that insufficient visual evidence can be overridden by dominant language priors. Severe deformation, blur, or perspective distortion further leads to Crumpled Paper Error, resulting in substantial omissions, substitutions, and failures to produce reliable transcriptions. These cases indicate that even basic content recognition remains vulnerable to visual ambiguity and distribution shifts.

Parsing failures reveal a fundamental gap between recognizing local elements and reconstructing their structural relationships within visual documents. Examples such as Molecular Graph Misparsing, Formula Superscript Flattening, and Merged-cell Structure Error demonstrate that accurate content recognition alone is insufficient for recovering the topological and hierarchical structures required for document understanding.

\section{Related Work}
Optical Character Recognition (OCR) has long been treated as a standalone text recognition task, with its outputs serving as input to downstream applications such as key information extraction and document question answering~\cite{subramani2020survey,molina2024fetch,wang2025document}.
While effective, downstream modules operate primarily on recognized text in such pipeline-based systems, making them susceptible to error propagation from inaccurate OCR results~\cite{zhang2020trie,shim2025revise,zhang2025ocr}.
To mitigate this limitation, recent research has shifted toward end-to-end modeling that performs downstream tasks directly on document images, enabling joint reasoning over textual content and visual layout while reducing reliance on intermediate OCR outputs~\cite{lee2023pix2struct,van2023document,yang2025cc}.

Recent advances in LMMs have substantially enhanced their document literacy, demonstrating strong potential across a wide range of document processing tasks~\cite{fu2025multimodal,ding2026deep}.
Prior work adapts LMMs to text-rich document images by enhancing their visual perception, enabling better understanding of dense text and complex layouts~\cite{ye2023ureader,zhang2025dockylin,liu2026textmonkey}. 
These methods typically remove redundant or less informative visual tokens~\cite{chen2024image,guo2025crop,yang2025visionzip}, integrate multi-scale visual features~\cite{park2024hierarchical,huang2025hires}, or introduce document-specific pretraining objectives~\cite{lv2023kosmos,peng2022ernie,bai2025qwen3}, thereby strengthening their ability to align textual content with visual layout. 
More recent efforts focus on enhancing the reasoning capabilities of LMMs over document content~\cite{yang2026realign,sun2025visrag,xiong2026lang2act}. 
They further introduce layout-aware reasoning that captures structural dependencies between textual content~\cite{mo2025doc,dong2026qianfan,xiong2026docr1}, or employ progressive zoom-in strategies that iteratively focus on relevant regions for fine-grained document understanding~\cite{wang2025vrag,su2025pixel}.

While recent advances have substantially improved LMMs for OCR-centric document processing, systematically evaluating their capabilities in real-world document understanding remains challenging~\cite{li2024survey,fu2024mme}. Existing benchmarks, such as OmniDocBench~\cite{ouyang2025omnidocbench} and olmOCR-Bench~\cite{poznanski2025olmocr}, primarily focus on document parsing and grounding over digitally rendered documents, where visual quality and layout structures are relatively well controlled. Real5-OmniDocBench~\cite{zhou2026real5} further improves realism by introducing physically captured document images; however, its evaluation scope remains limited to a narrow set of tasks, leaving many practical challenges in real-world document processing insufficiently explored~\cite{beyene2026survey,li2025readoc}. Beyond document parsing, OCRBench~\cite{liu2024ocrbench} and OCRBench v2~\cite{fu2024ocrbench} expand evaluation toward a broader range of OCR-centric capabilities. However, their task suites exhibit substantial overlap and remain limited in covering end-to-end document processing workflows encountered in real-world applications. CC-OCR~\cite{yang2025cc} further advances this direction by incorporating more realistic scenarios, diverse application domains, and multilingual settings. Nevertheless, its evaluation primarily emphasizes aggregate performance, offering limited diagnostic analysis of the underlying capabilities and failure modes of LMMs.

\section{Conclusion}
We present \method{}, a benchmark for attributing LMM failures in real-world document processing. \method{} unifies diverse tasks, documents, and languages, while providing fine-grained document factors that allow performance to be decomposed across observable input conditions. Evaluation of 17 representative LMMs reveals substantial variation across task requirements and document characteristics. 

Fine-grained attribution and error analysis further show that models with similar overall performance can fail for markedly different reasons, suggesting that continued improvements in benchmark scores may still conceal important weaknesses that affect practical deployment. By making such weaknesses explicit, \method{} can support more informed model selection and help direct data collection and model development. We hope \method{} will advance OCR-centric evaluation beyond model ranking toward a more explanatory understanding of model performance.

\bibliographystyle{ACM-Reference-Format}
\bibliography{custom}

\clearpage
\newpage
\appendix
\raggedbottom
\section{Appendix}
\subsection{License}
\label{app:license}

We strictly comply with the original licenses and terms of use associated with all data used to construct \method{}. The benchmark is built from three sources: samples inherited from CC-OCR, documents collected from public corpora and web sources, and challenging cases collected from real-world document-processing environments.

CC-OCR is released under the MIT License, and the inherited samples are included and redistributed in accordance with its original licensing terms. For documents collected from public corpora and web sources, we follow the licenses, copyright restrictions, and usage terms specified by their original providers. The original terms take precedence for all third-party content.

For documents collected from real-world document-processing environments, only samples authorized for public redistribution are included. Sensitive and personally identifiable information is removed before release. The detailed redaction and review procedure is described in Appendix~\ref{app:data_redaction}.

\subsection{Semantic Analysis of Benchmark Annotations}
\label{app:semantic_analysis}

\subsubsection{Document-Type Taxonomy}
\label{app:document_type_semantic}

We first examine whether the ten document-type annotations in \method{} are consistent with the multimodal semantic space. For each sample, we construct an image--question pair by combining the document image with its corresponding question. We then use Qwen3-VL-Embedding-2B~\cite{li2026qwen3} to encode each image--question pair into a multimodal embedding vector.

\begin{figure}[t]
    \centering
    \begin{subfigure}[t]{0.48\linewidth}
        \centering
        \includegraphics[width=\linewidth]{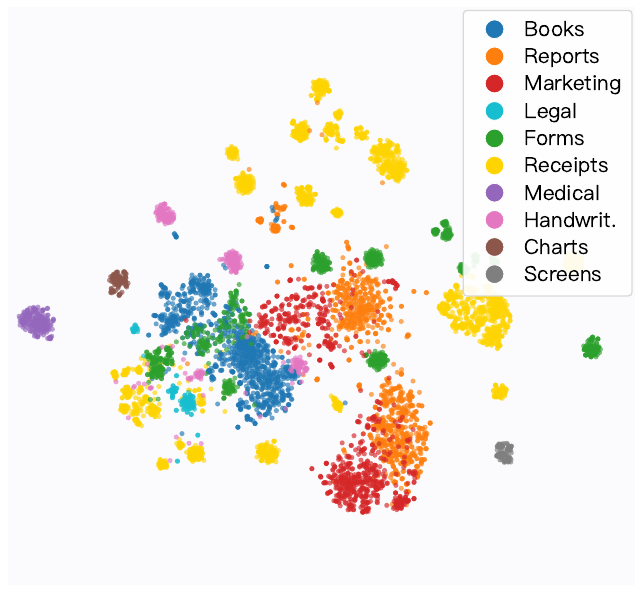}
        \caption{t-SNE visualization of image-question embeddings, colored by annotated document type.}
        \label{fig:doc_type_tsne}
    \end{subfigure}\hfill%
    \begin{subfigure}[t]{0.48\linewidth}
        \centering
        \includegraphics[width=\linewidth]{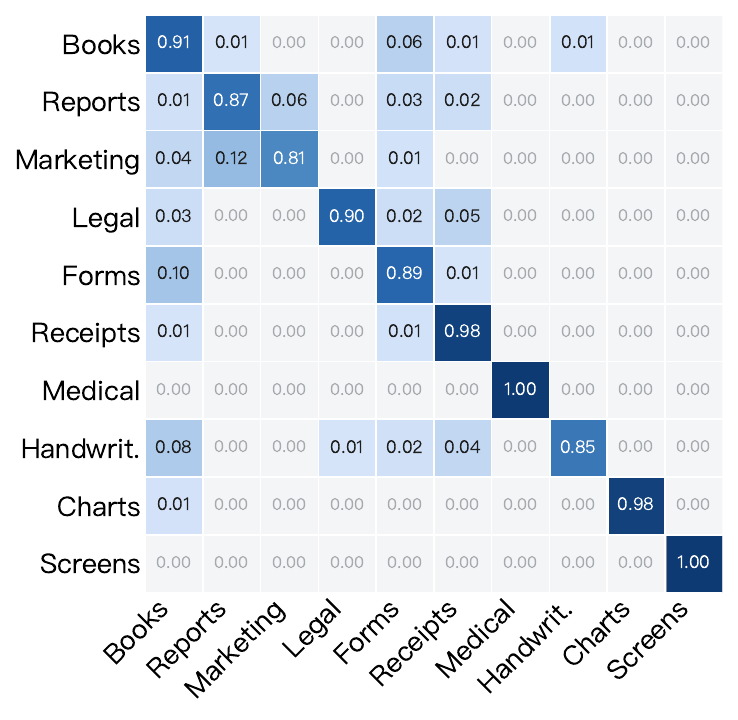}
        \caption{Row-normalized kNN document-type agreement matrix in the embedding space.}
        \label{fig:doc_type_knn}
    \end{subfigure}\hfill%
    \caption{Semantic analysis of the document-type taxonomy in CC-OCR V2. Each sample is embedded with both the document image and its associated question.}
    \label{fig:doc_semantic}
\end{figure}

Figure~\ref{fig:doc_semantic} presents two complementary views of the resulting embedding space. We apply t-SNE~\cite{van2008visualizing} to the multimodal embeddings for two-dimensional visualization and color each sample according to its annotated document type in Figure~\ref{fig:doc_type_tsne}. Several document types form compact and visually separable groups. Medical documents and screens are highly isolated, while receipts and charts also form clear local clusters. Books, reports, marketing materials, and forms occupy broader regions, reflecting their greater layout diversity and partial overlap in visual structure.

Figure~\ref{fig:doc_type_knn} provides a quantitative local-neighborhood analysis~\cite{cover1967nearest}. For each sample, we compute its top-$10$ nearest neighbors in the original high-dimensional embedding space using cosine similarity. We then calculate the proportion of neighboring samples belonging to each document type and average the results within each source document type. The resulting matrix is row-normalized, where each row denotes the annotated document type of the query sample and each column denotes the document type of its nearest neighbors.

The main off-diagonal entries correspond to visually or semantically adjacent categories. Marketing materials show noticeable overlap with reports, suggesting that these document types share similar page designs and visual layouts. Forms also exhibit overlap with books, potentially because both frequently contain text-heavy and structured regions. Handwritten documents overlap with books and receipts, reflecting that handwriting may occur across multiple document sources rather than constituting an entirely independent visual category.

\subsubsection{Diagnostic-Dimension Taxonomies}
\label{app:diagnostic_semantic}

\begin{figure}[t]
    \centering

    \captionsetup[subfigure]{
        font=tiny,
        skip=1pt,
        justification=centering
    }

    \begin{subfigure}[t]{0.49\linewidth}
        \centering
        \includegraphics[width=\linewidth]
        {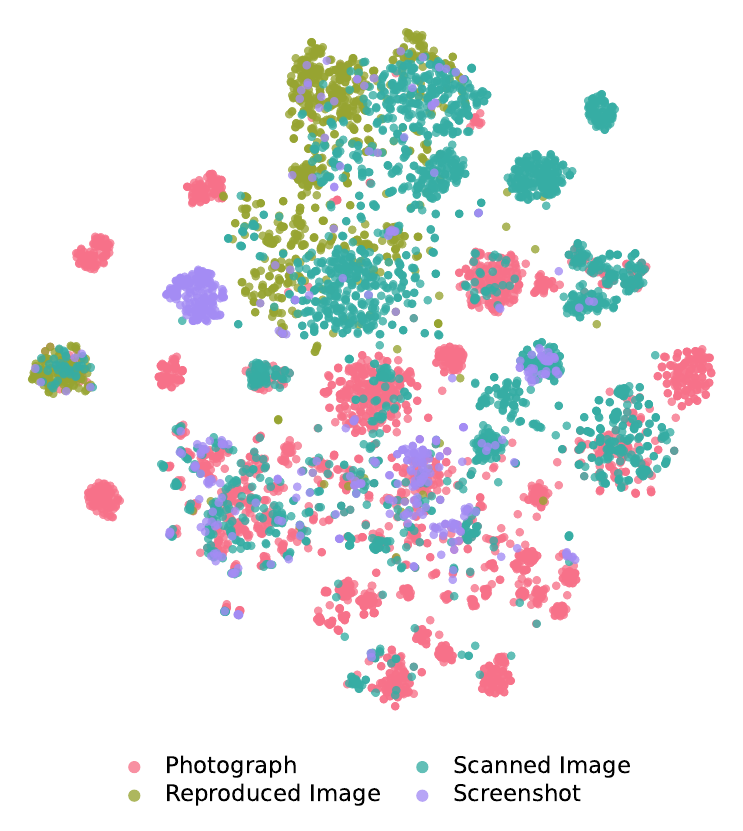}
        \caption{Imaging Method}
        \label{fig:diagnostic_tsne_imaging}
    \end{subfigure}
    \hfill
    \begin{subfigure}[t]{0.49\linewidth}
        \centering
        \includegraphics[width=\linewidth]
        {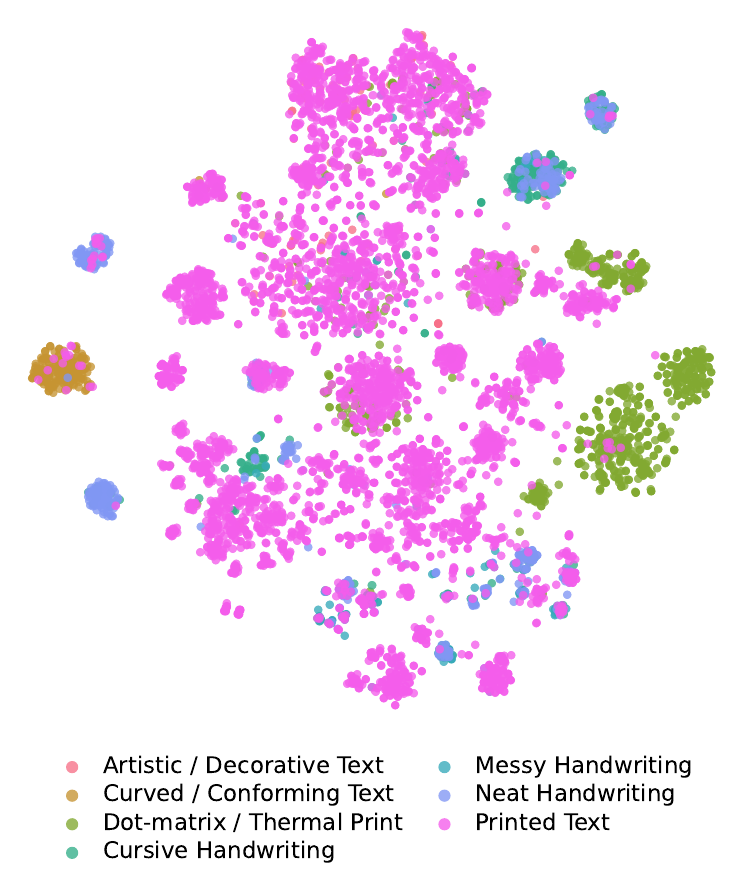}
        \caption{Font and Text Form}
        \label{fig:diagnostic_tsne_font}
    \end{subfigure}

    \vspace{2pt}

    \begin{subfigure}[t]{0.49\linewidth}
        \centering
        \includegraphics[width=\linewidth]
        {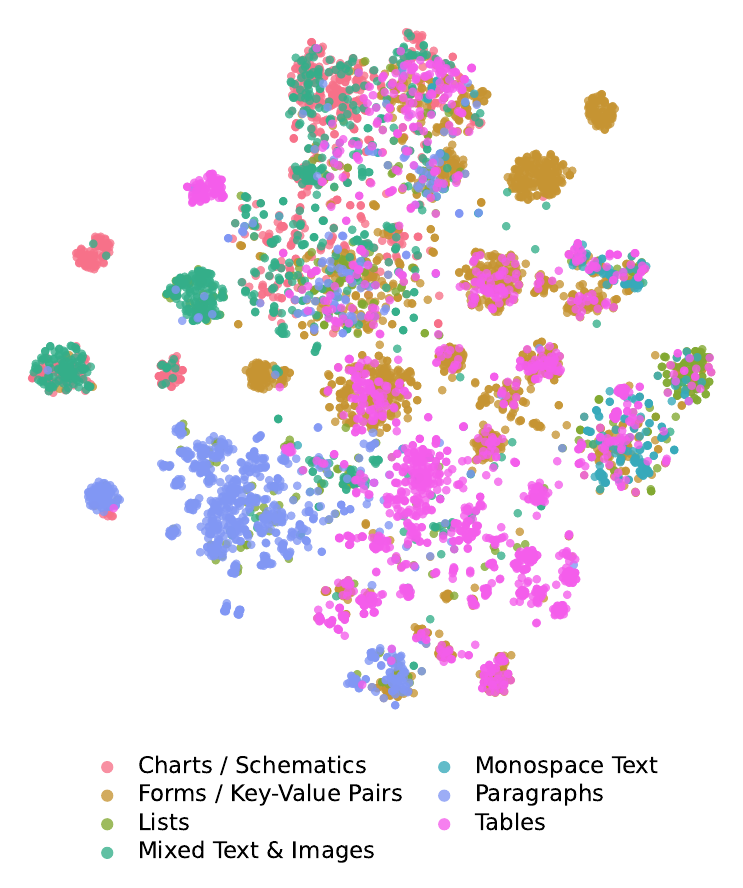}
        \caption{Layout Structure}
        \label{fig:diagnostic_tsne_layout}
    \end{subfigure}
    \hfill
    \begin{subfigure}[t]{0.49\linewidth}
        \centering
        \includegraphics[width=\linewidth]
        {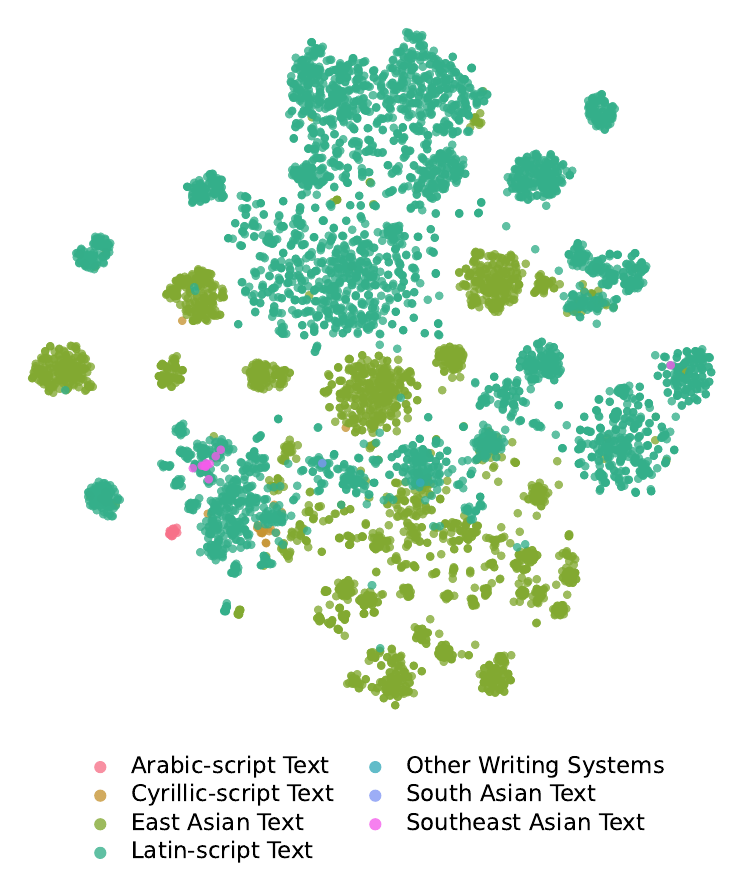}
        \caption{Language}
        \label{fig:diagnostic_tsne_language}
    \end{subfigure}

    \caption{
    Semantic organization of four representative diagnostic
    dimensions in the shared multimodal embedding space.
    All panels use the same image--question embeddings and identical
    t-SNE coordinates, with samples colored according to different
    diagnostic annotations.
    }
    \label{fig:diagnostic_tsne}
\end{figure}

We further investigate whether the categories defined under four representative diagnostic dimensions are reflected in the same multimodal semantic space. Specifically, we analyze Imaging Method, Font and Text Form, Layout Structure, and Language. All four visualizations in Figure~\ref{fig:diagnostic_tsne} use the same image--question embeddings and identical two-dimensional t-SNE coordinates as the document-type analysis. Only the category used to color each sample is changed, allowing direct comparison of how different diagnostic annotations organize the shared embedding space.

The four dimensions exhibit different levels of semantic separability. Imaging methods form relatively coherent local neighborhoods, although photographs, scans, and reproduced images partially overlap because they may contain similar document content. Within Font and Text Form, printed text occupies a broad portion of the embedding space, whereas several handwriting and specialized printing styles form more localized clusters. Layout Structure shows greater category overlap, indicating that documents with different structural organizations may nevertheless share similar visual and semantic content. For Language, Latin-script and East Asian documents dominate the embedding space, while several less frequent writing systems form smaller local groups.

To quantify these observations, Table~\ref{tab:diagnostic_knn} reports top-$10$ nearest-neighbor agreement computed in the original 2,048-dimensional embedding space rather than on the t-SNE coordinates. Imaging Method achieves the highest macro agreement, indicating relatively consistent local organization across its categories. Font and Text Form and Language obtain substantially higher micro than macro agreement because their largest categories, Printed Text and Latin-script Text, dominate the sample distribution. Layout Structure exhibits lower agreement, consistent with the greater visual and semantic overlap observed in the visualization.

\begin{table}[t]
    \centering
    \caption{
    Top-$10$ nearest-neighbor label agreement for four representative
    diagnostic dimensions. Neighbors are retrieved using cosine
    similarity in the original 2,048-dimensional multimodal embedding
    space.
    }
    \label{tab:diagnostic_knn}

    \small
    \setlength{\tabcolsep}{6pt}
    \renewcommand{\arraystretch}{1.08}

    \begin{tabular}{lcc}
        \toprule
        \textbf{Diagnostic Dimension}
        & \textbf{Micro}
        & \textbf{Macro} \\
        \midrule
        Imaging Method
        & 0.770
        & 0.724 \\
        Font and Text Form
        & 0.812
        & 0.574 \\
        Layout Structure
        & 0.631
        & 0.569 \\
        Language
        & 0.948
        & 0.501 \\
        \bottomrule
    \end{tabular}
\end{table}

\FloatBarrier

\subsection{Detailed Data Statistics of \method{}}
\label{app:statistics}

To support fine-grained diagnosis beyond task-level evaluation, we annotate all 7,093 samples in \method{} along ten dimensions covering visual quality, acquisition conditions, document structure, text form, spatial arrangement, and language. Each sample is assigned to one category within each dimension. Figure~\ref{fig:a2_all_distributions} shows the corresponding category distributions.

Several dimensions are dominated by common document conditions. Clear images account for 91.5\% of the benchmark, while 82.5\% of the samples contain no lighting or screen-related issue and 91.1\% contain no adversarial or misleading factor. Printed text, horizontal layouts, and non-mixed layouts account for 67.9\%, 65.6\%, and 73.6\%, respectively. Nevertheless, \method{} retains substantial diversity in document structure and acquisition conditions. Forms and tables together constitute 53.2\% of the samples, photographs and scans
account for 77.8\%, and more than half of the documents contain some form of natural interference, including masking, occlusion, watermarks, stamps, complex backgrounds, and overlapping content.

\begin{figure*}[t]
    \centering

    \begin{subfigure}[t]{0.48\textwidth}
        \centering
        \includegraphics[width=\linewidth]{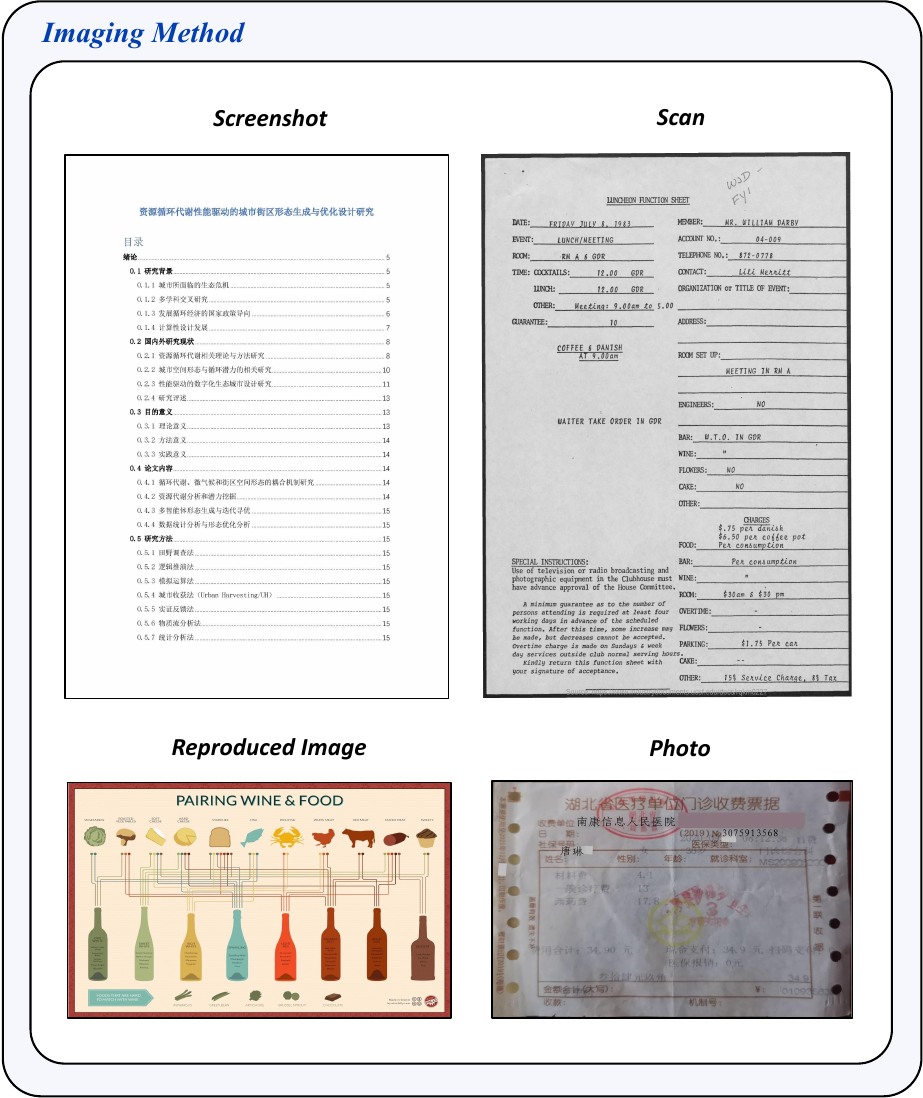}
        \caption{Imaging Method}
        \label{fig:a2_imaging_method}
    \end{subfigure}
    \hfill
    \begin{subfigure}[t]{0.48\textwidth}
        \centering
        \includegraphics[width=\linewidth]{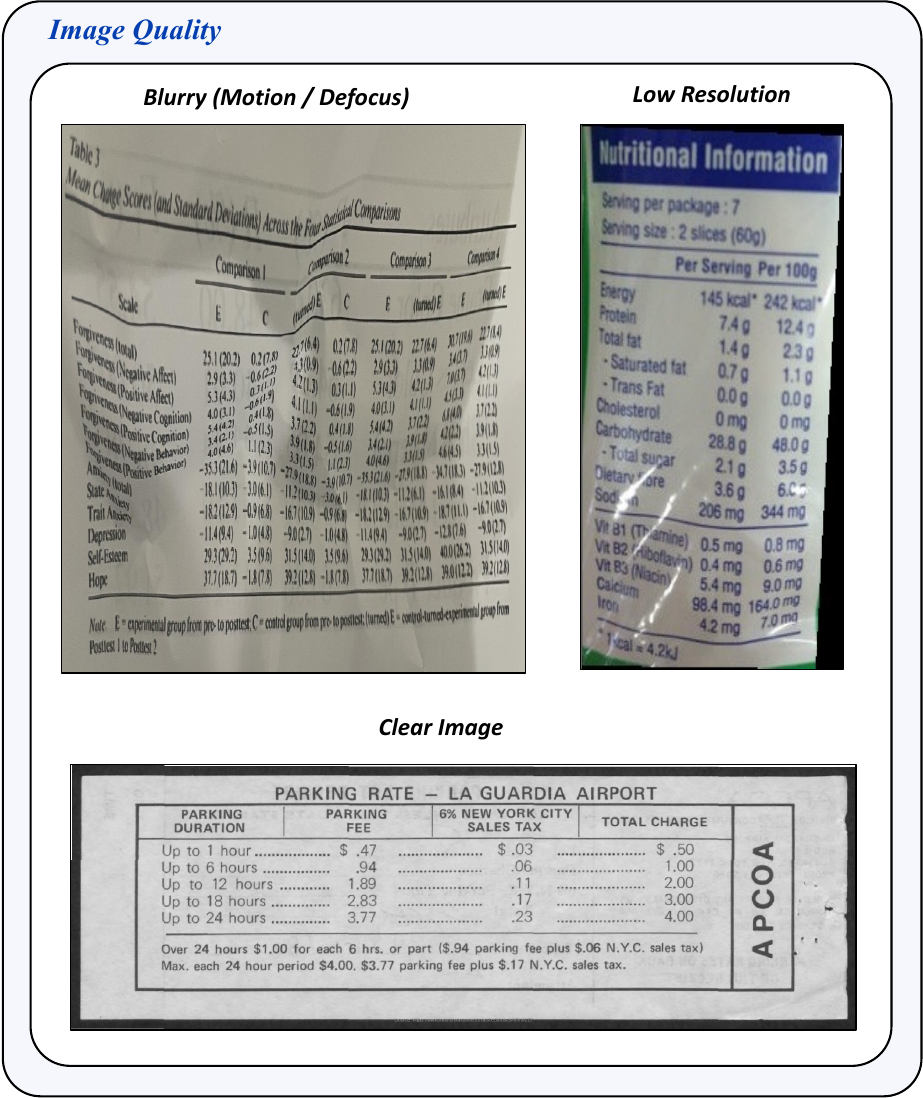}
        \caption{Image Quality}
        \label{fig:a2_image_quality}
    \end{subfigure}

    \vspace{0.8em}

    \begin{subfigure}[t]{0.48\textwidth}
        \centering
        \includegraphics[width=\linewidth]{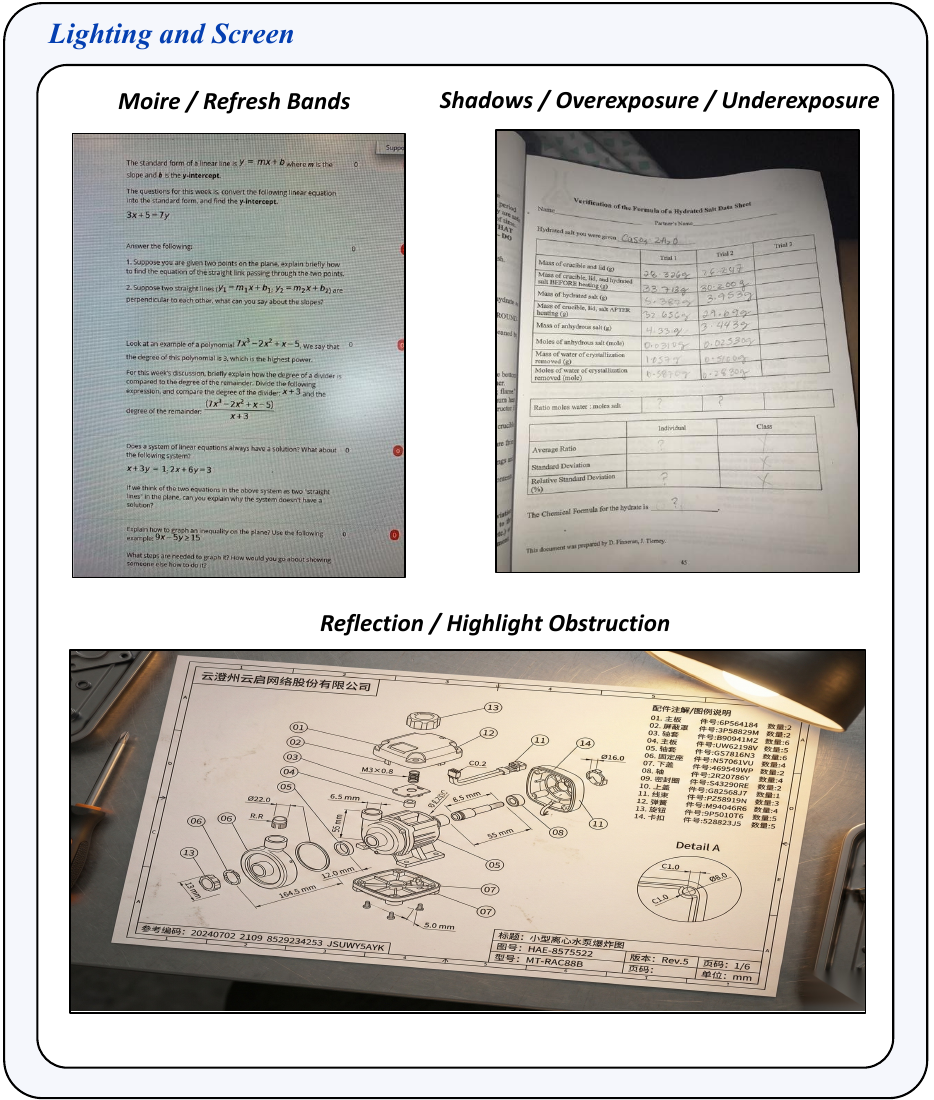}
        \caption{Lighting and Screen}
        \label{fig:a2_lighting_screen}
    \end{subfigure}
    \hfill
    \begin{subfigure}[t]{0.48\textwidth}
        \centering
        \includegraphics[width=\linewidth]{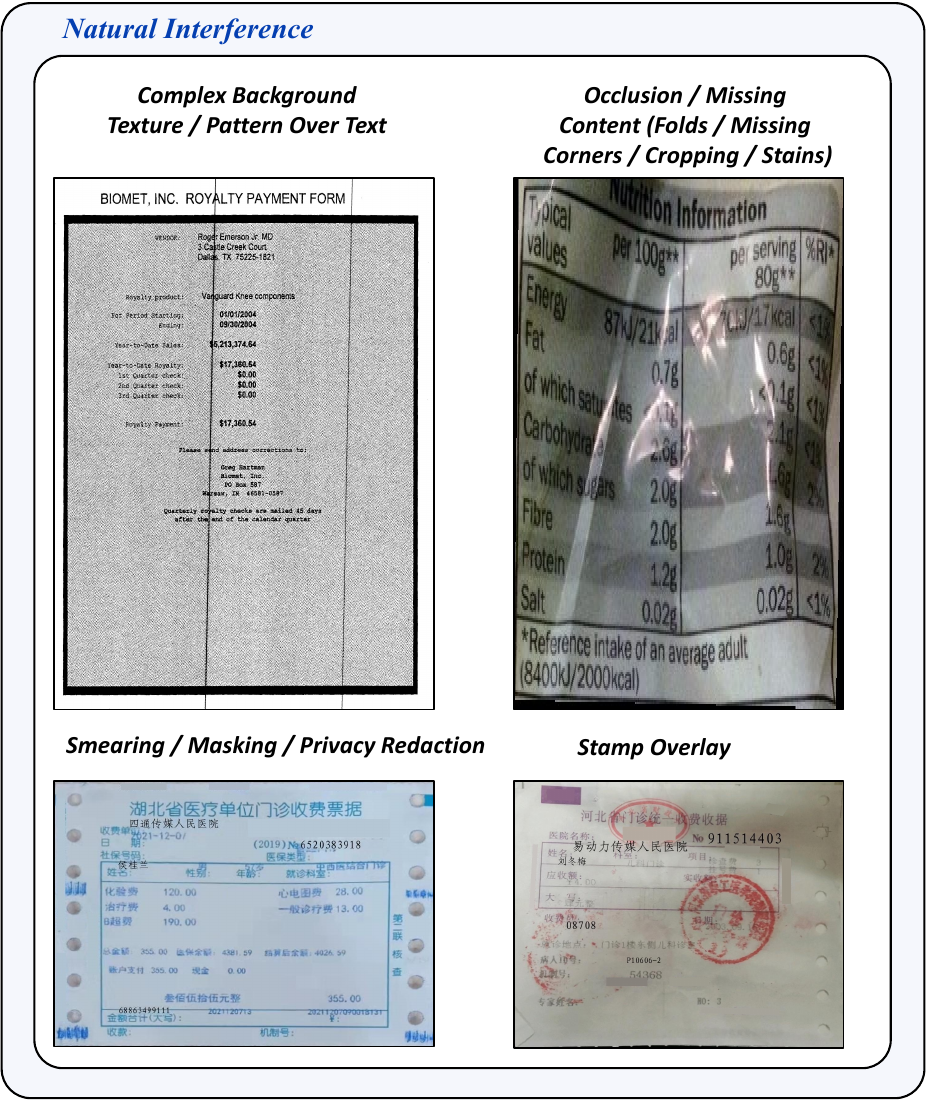}
        \caption{Natural Interference}
        \label{fig:a2_natural_interference}
    \end{subfigure}

    \caption{
    Representative examples of acquisition conditions and visual degradation dimensions in CC-OCR-v2. 
    These dimensions cover image sources, quality variations, lighting or screen artifacts, and natural visual interference.
    }
    \label{fig:a2_acquisition_quality_examples}
\end{figure*}

\begin{figure*}[t]
    \centering

    \begin{subfigure}[t]{0.48\textwidth}
        \centering
        \includegraphics[width=\linewidth]{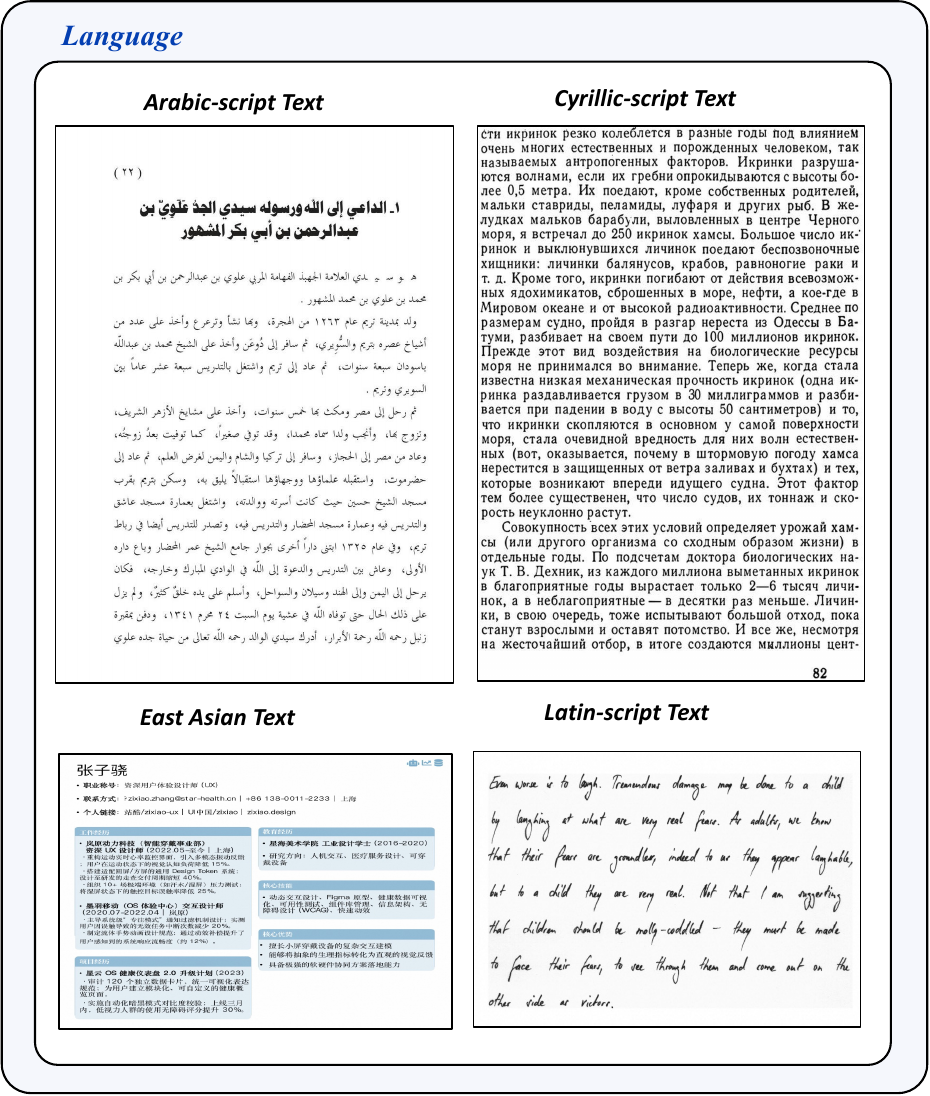}
        \caption{Language}
        \label{fig:a2_language}
    \end{subfigure}
    \hfill
    \begin{subfigure}[t]{0.48\textwidth}
        \centering
        \includegraphics[width=\linewidth]{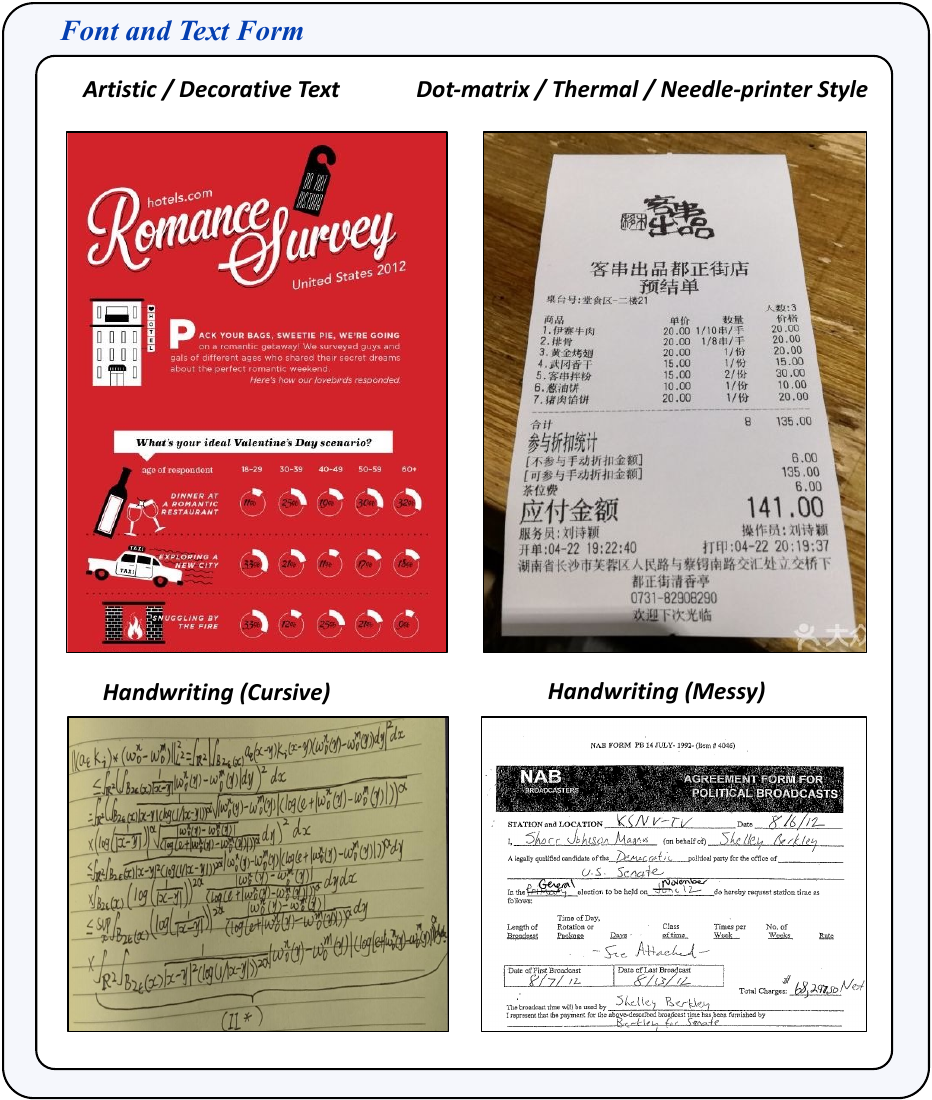}
        \caption{Font and Text Form}
        \label{fig:a2_font_text_form}
    \end{subfigure}

    \vspace{0.8em}

    \begin{subfigure}[t]{0.48\textwidth}
        \centering
        \includegraphics[width=\linewidth]{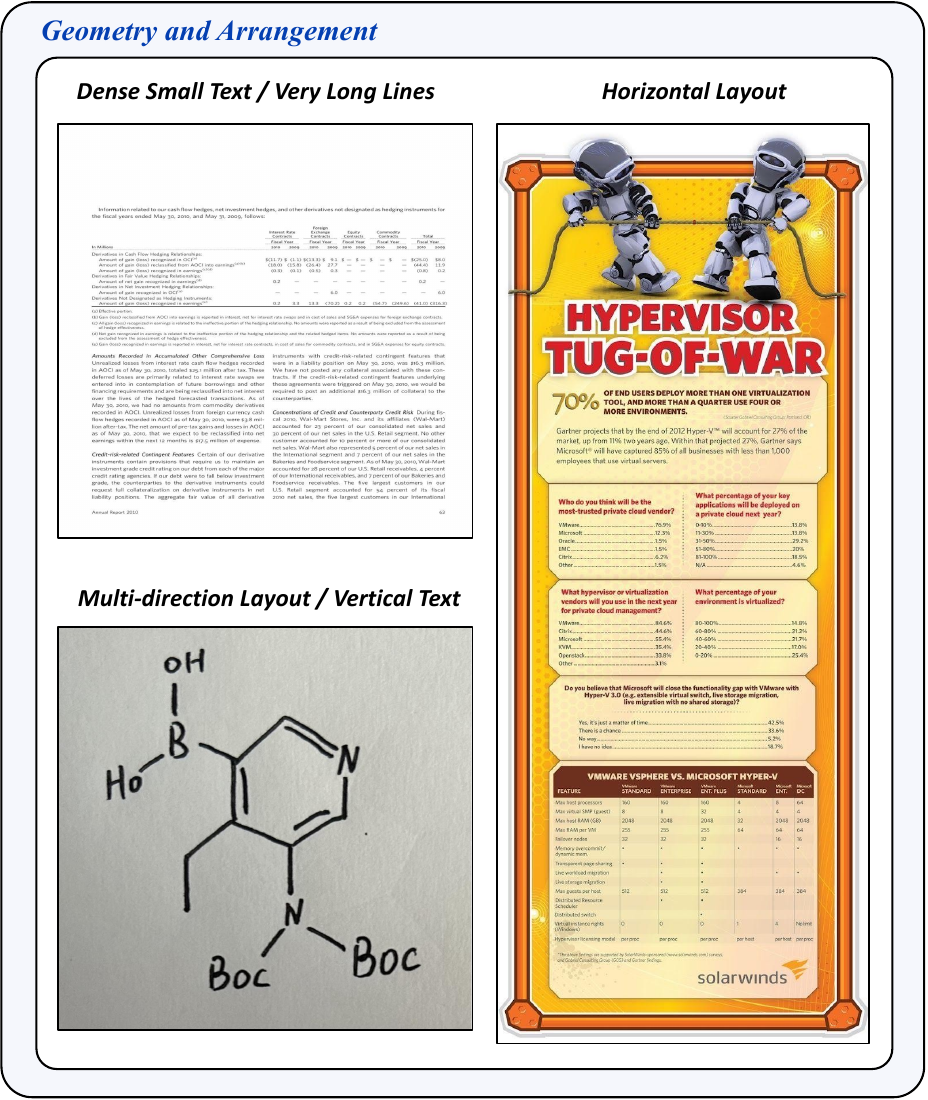}
        \caption{Geometry and Arrangement}
        \label{fig:a2_geometry_arrangement}
    \end{subfigure}
    \hfill
    \begin{subfigure}[t]{0.48\textwidth}
        \centering
        \includegraphics[width=\linewidth]{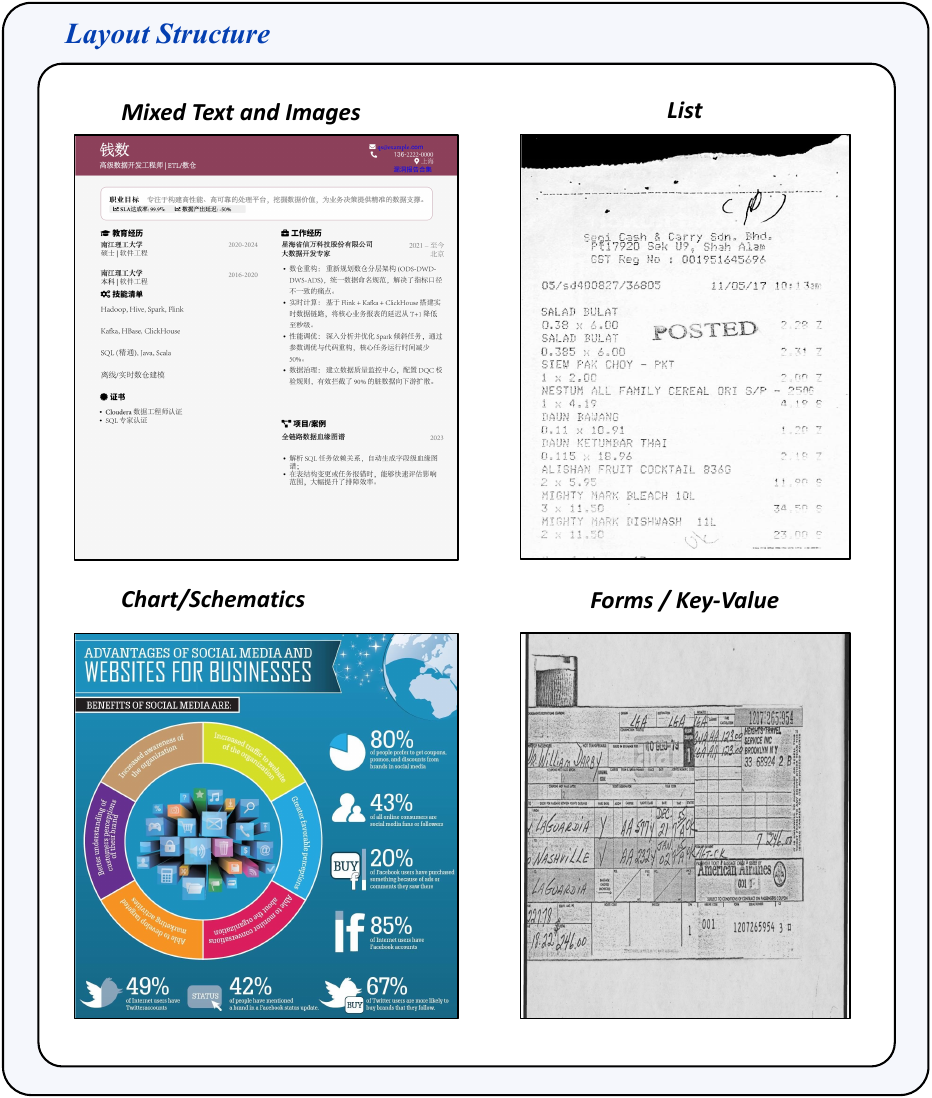}
        \caption{Layout Structure}
        \label{fig:a2_layout_structure}
    \end{subfigure}

    \caption{
    Representative examples of text appearance and layout diversity dimensions in CC-OCR-v2. 
    These dimensions reflect script diversity, font styles, geometric arrangements, and document-level structures.
    }
    \label{fig:a2_text_layout_examples}
\end{figure*}

\begin{figure*}[t]
    \centering

    \begin{subfigure}[t]{0.48\textwidth}
        \centering
        \includegraphics[width=\linewidth]{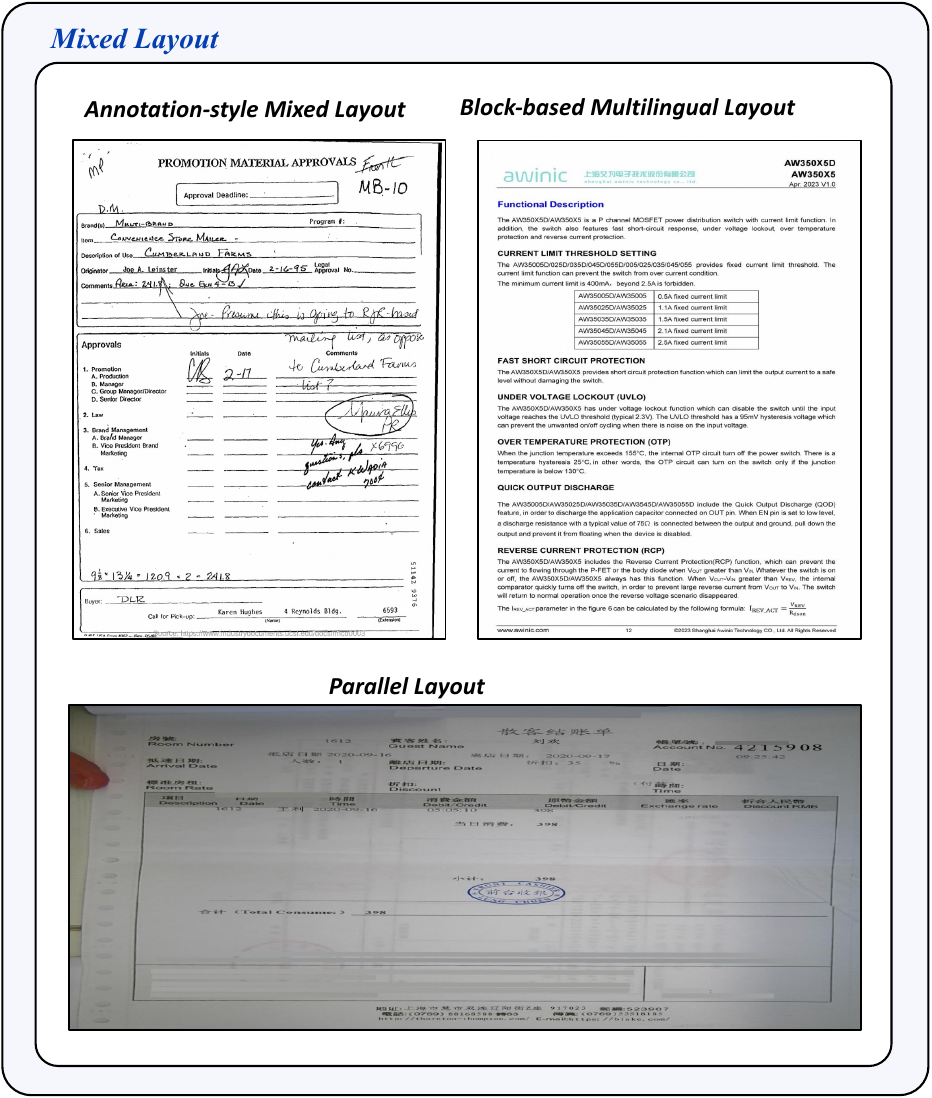}
        \caption{Mixed Layout}
        \label{fig:a2_mixed_layout}
    \end{subfigure}
    \hfill
    \begin{subfigure}[t]{0.48\textwidth}
        \centering
        \includegraphics[width=\linewidth]{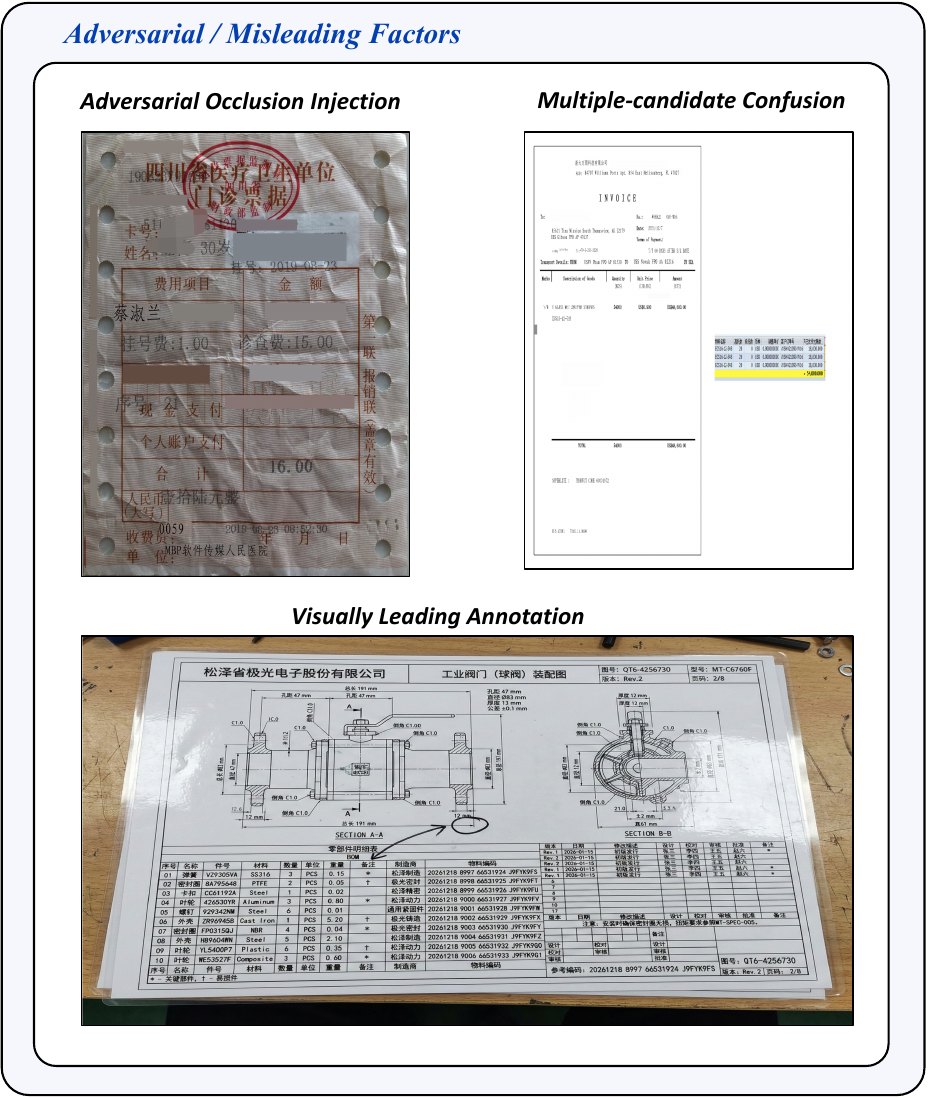}
        \caption{Adversarial / Misleading Factors}
        \label{fig:a2_adversarial_misleading}
    \end{subfigure}

    \caption{
    Representative examples of complex layout and misleading-factor dimensions in CC-OCR-v2. 
    These examples include mixed reading structures, visual annotations, multiple candidates, and misleading cues.
    }
    \label{fig:a2_complex_misleading_examples}
\end{figure*}

Figures~\ref{fig:a2_acquisition_quality_examples}--\ref{fig:a2_complex_misleading_examples} provide representative examples of the ten diagnostic dimensions. Specifically, Figure~\ref{fig:a2_acquisition_quality_examples} illustrates variations
in imaging method, image quality, lighting conditions, and natural interference. Figure~\ref{fig:a2_text_layout_examples} presents examples
of language, font and text form, geometry and arrangement, and layout structure. Figure~\ref{fig:a2_complex_misleading_examples} further
shows mixed-layout documents and samples containing adversarial or misleading factors. These examples demonstrate that the diagnostic annotations capture both common document characteristics and challenging conditions encountered in realistic document processing.

\subsection{Detailed Multilingual Recognition Capability Analysis of LMMs}
\label{app:multilingual}

\begin{figure*}
    \centering
    \includegraphics[width=\linewidth]{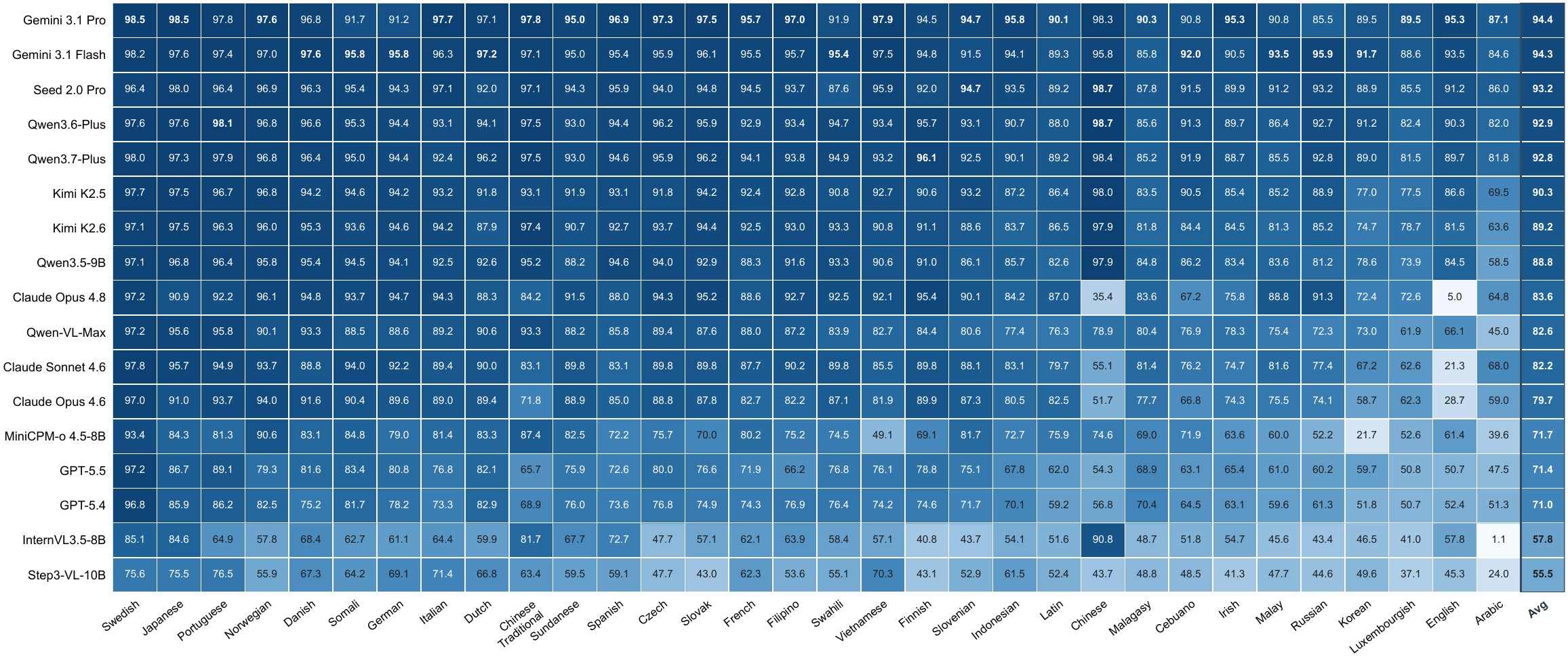}
    \caption{Language-level recognition performance of evaluated LMMs on the 32 multilingual recognition subsets. Each cell reports the recognition micro-F1 score, with darker colors indicating stronger performance.}
    \label{fig:heatmap}
\end{figure*}

Figure~\ref{fig:heatmap} reports the language-level recognition
performance of 17 evaluated LMMs on the 32 multilingual recognition
subsets in CC-OCR v2. Each cell denotes the micro-F1 score of one model
on one language, and the rightmost column reports the average across all
language subsets. The multilingual recognition setting contains 640
samples in total.

Overall, Gemini 3.1 Pro and Gemini 3.1 Flash achieve the highest average
scores of 94.4 and 94.3, respectively. They are followed by Seed 2.0 Pro,
Qwen3.6-Plus, and Qwen3.7-Plus, whose average scores all exceed 92.
These leading models maintain consistently strong performance across
most high-resource languages, including Swedish, Japanese, Portuguese,
Norwegian, Chinese, and Chinese Traditional. Nevertheless, their
relative rankings still vary across individual languages, indicating
that no single model uniformly dominates all writing systems.

Several subsets remain substantially more challenging. Arabic exhibits
the most consistent degradation across model families, while Korean,
Luxembourgish, English, and several long-tail languages also produce
large model-dependent variations. The gaps are particularly pronounced
for models outside the leading group: performance on commonly supported
European and East Asian languages can remain high, while recognition
accuracy drops sharply on less represented scripts or languages. These
results suggest that sharing an alphabet or writing system does not
necessarily lead to similar recognition performance.

Among the remaining on-server models, Kimi K2.5 and Kimi K2.6 maintain
relatively strong multilingual performance, with average scores of 90.3
and 89.2. Claude Opus 4.8 improves over the Claude 4.6 variants in
overall performance, but its language profile remains highly uneven:
it performs strongly on many European languages while exhibiting severe
drops on specific subsets such as Chinese and English. Qwen-VL-Max,
Claude Sonnet 4.6, Claude Opus 4.6, GPT-5.5, and GPT-5.4 show greater
cross-language variance, especially on Arabic and several long-tail
language subsets.

For on-device models, Qwen3.5-9B is the strongest model, achieving an
average score of 88.8 and remaining competitive on many high-resource
languages. Its performance nevertheless decreases noticeably on
Arabic, Luxembourgish, Korean, and Russian. MiniCPM-o 4.5-8B,
InternVL3.5-8B, and Step3-VL-10B exhibit substantially larger
cross-language variation, with particularly clear degradation on
Vietnamese, Korean, Luxembourgish, and Arabic. This indicates that
compact deployment-oriented models still face considerable challenges
in maintaining stable recognition performance across diverse scripts.

Overall, these results demonstrate that aggregate multilingual scores
can conceal substantial language-specific weaknesses, motivating
language-level evaluation across diverse scripts and resource levels.

\subsection{Fine-grained Diagnostic Analysis of LMMs}
\label{app:attribute}

\begin{figure}[!t]
    \centering

    \captionsetup[subfigure]{
        font=scriptsize,
        skip=1pt,
        justification=centering,
        singlelinecheck=true
    }

    \begin{subfigure}[t]{0.48\linewidth}
        \centering
        \includegraphics[width=\linewidth]
        {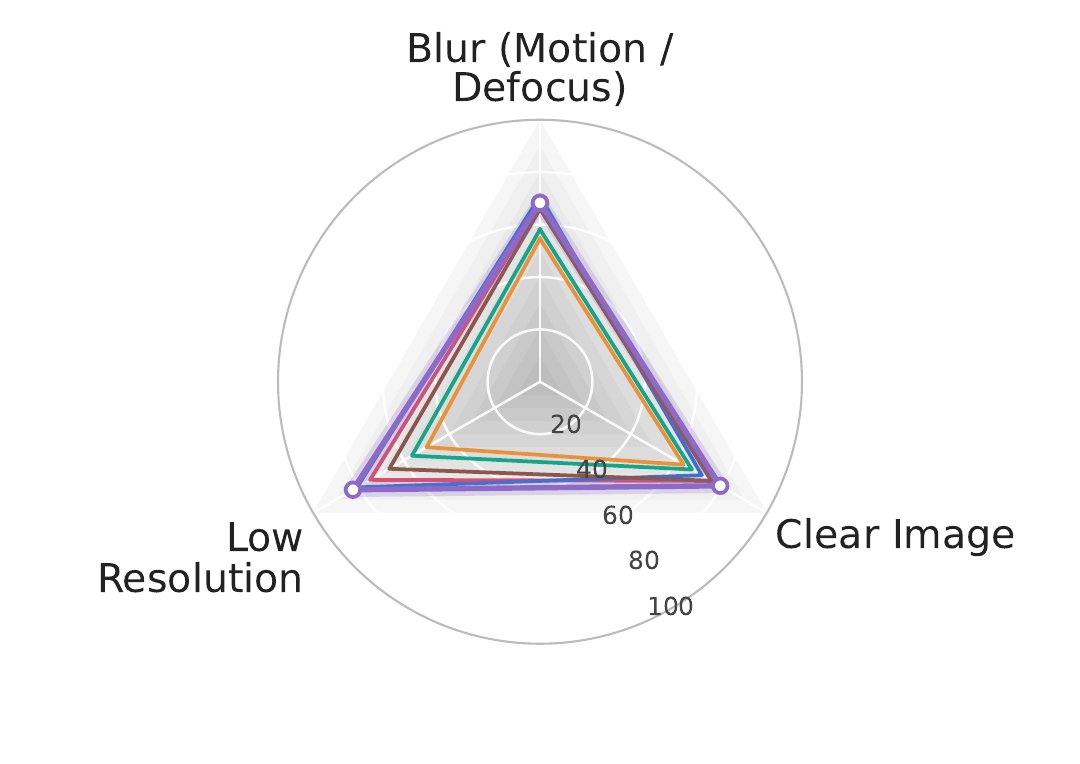}
        \caption{Image Quality}
        \label{fig:radar_image_quality}
    \end{subfigure}
    \hfill
    \begin{subfigure}[t]{0.48\linewidth}
        \centering
        \includegraphics[width=\linewidth]
        {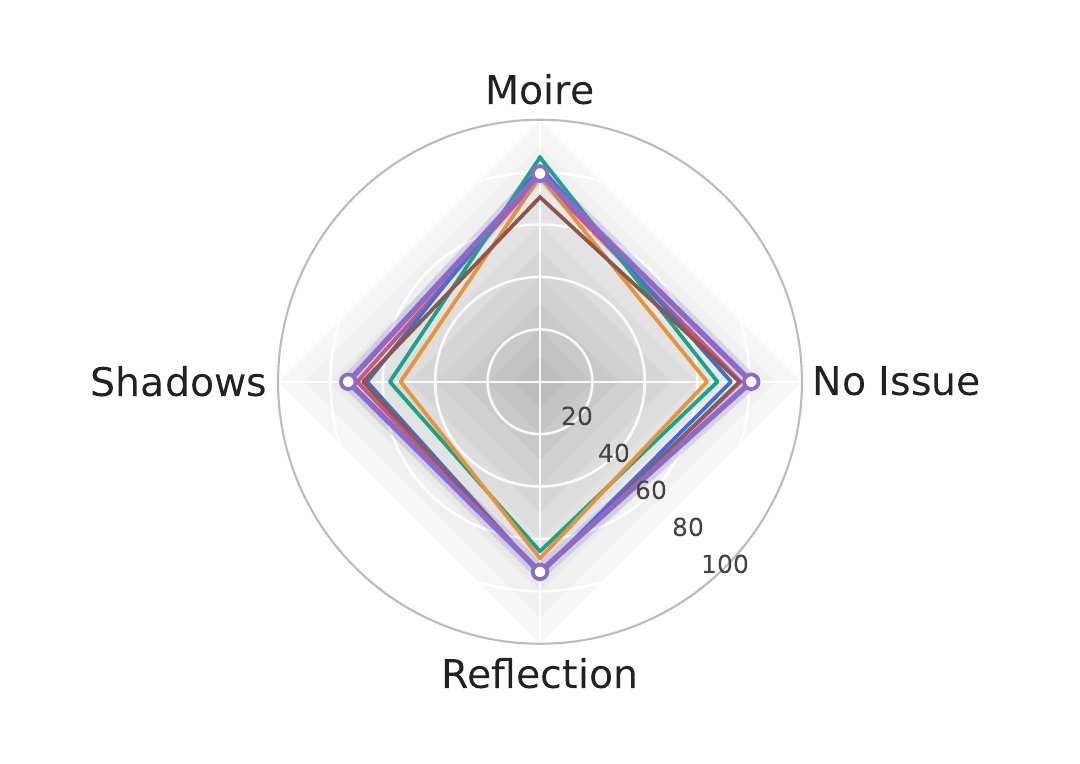}
        \caption{Lighting and Screen}
        \label{fig:radar_lighting_screen}
    \end{subfigure}

    \par\vspace{2pt}

    \begin{subfigure}[t]{0.48\linewidth}
        \centering
        \includegraphics[width=\linewidth]
        {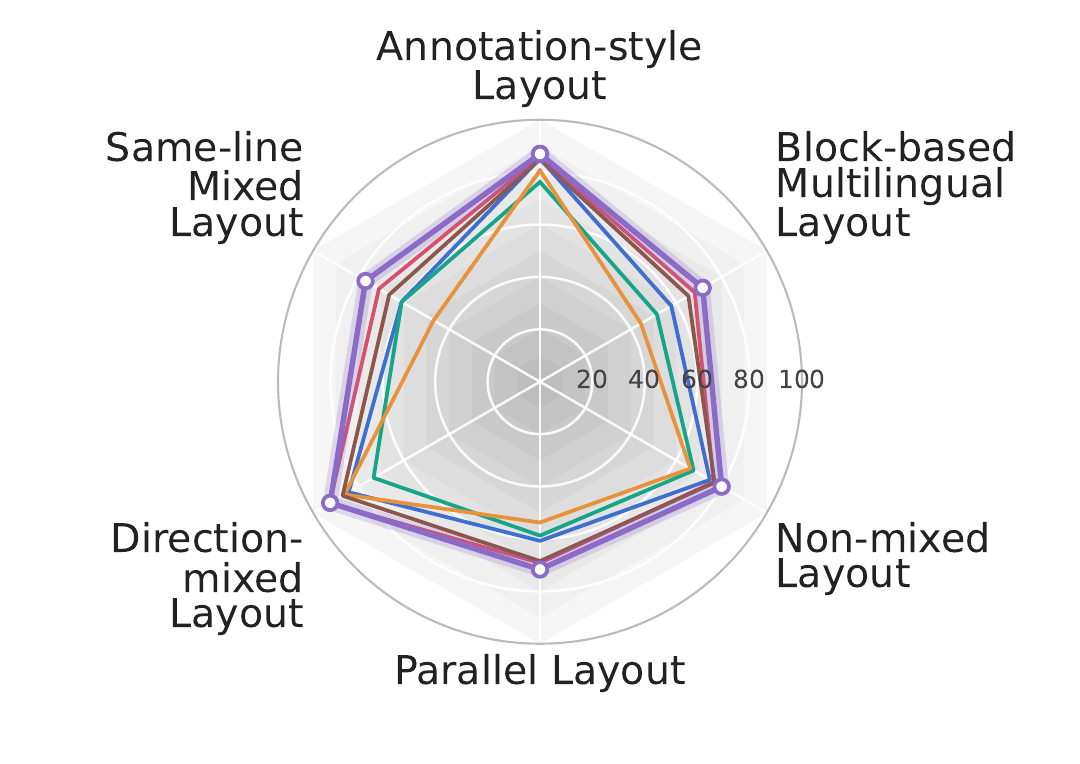}
        \caption{Mixed Layout}
        \label{fig:radar_mixed_layout}
    \end{subfigure}
    \hfill
    \begin{subfigure}[t]{0.48\linewidth}
        \centering
        \includegraphics[width=\linewidth]
        {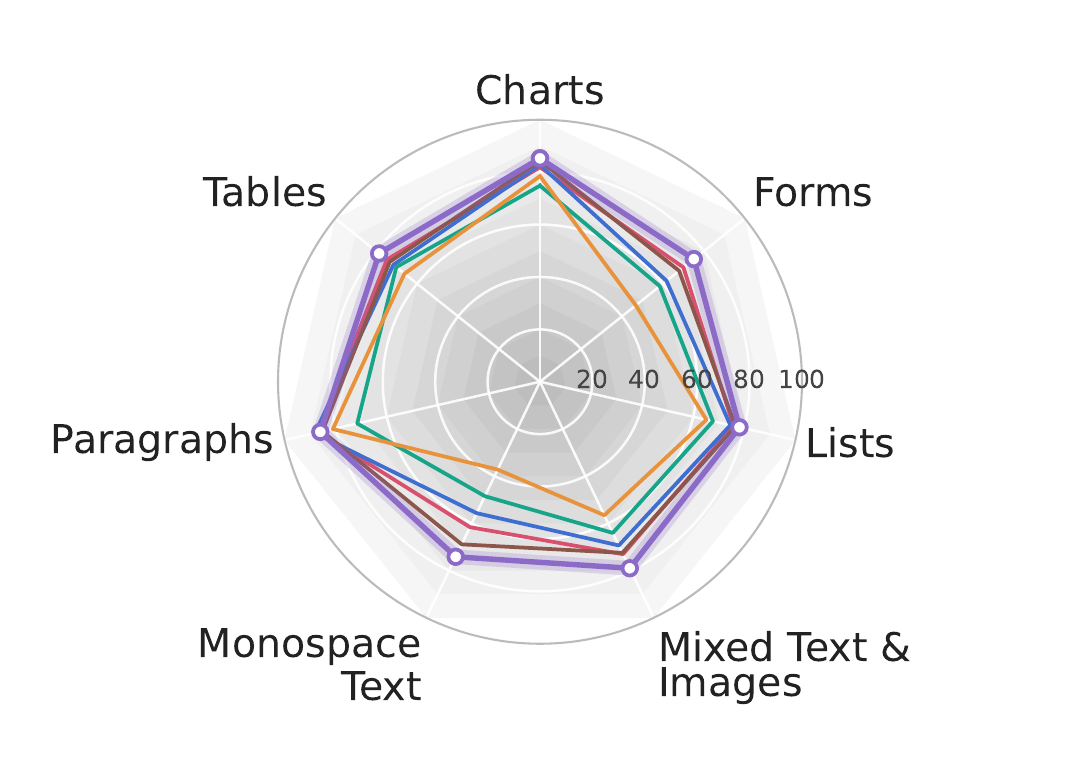}
        \caption{Layout Structure}
        \label{fig:radar_layout_structure}
    \end{subfigure}

    \par\vspace{2pt}

    \begin{subfigure}[t]{0.48\linewidth}
        \centering
        \includegraphics[width=\linewidth]
        {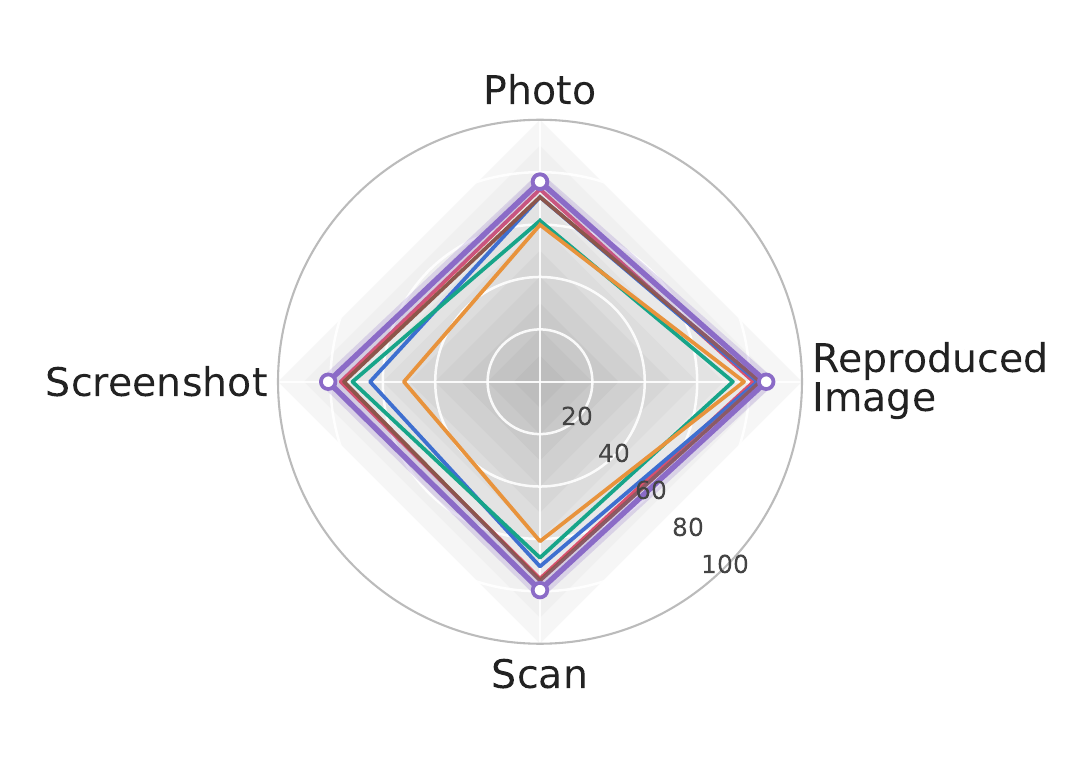}
        \caption{Imaging Method}
        \label{fig:radar_imaging_method}
    \end{subfigure}
    \hfill
    \begin{subfigure}[t]{0.48\linewidth}
        \centering
        \includegraphics[width=\linewidth]
        {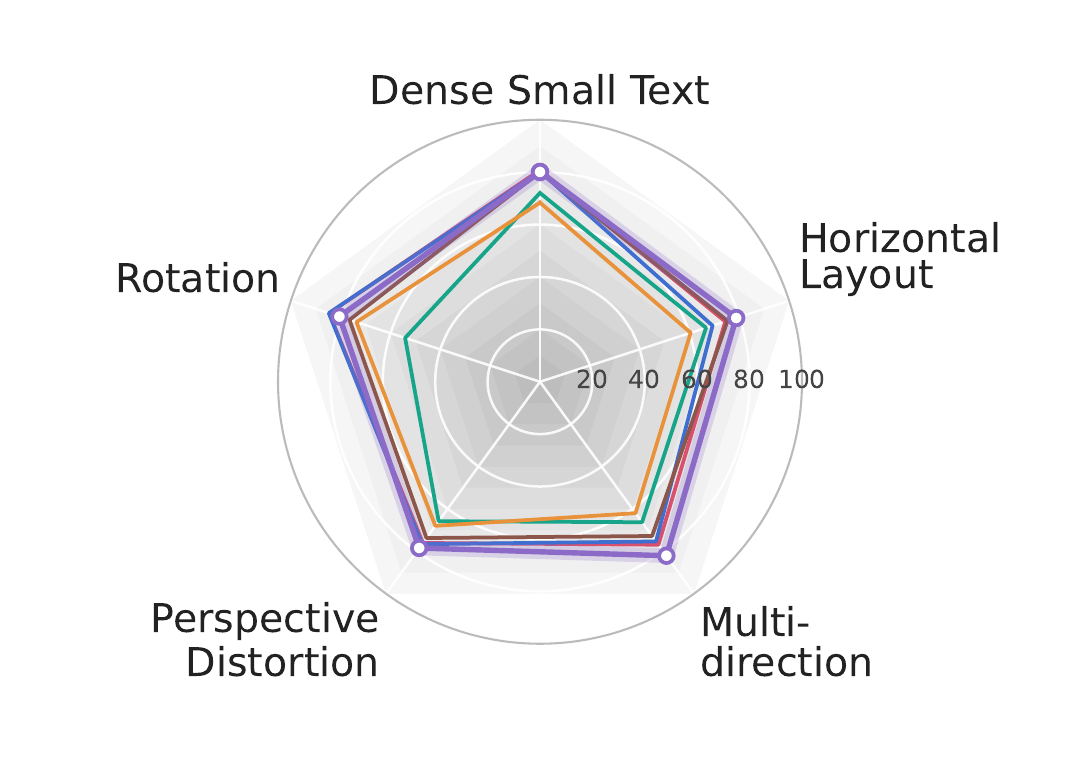}
        \caption{\shortstack{Geometry and\\Arrangement}}
        \label{fig:radar_geometry_arrangement}
    \end{subfigure}

    \par\vspace{2pt}
    \includegraphics[width=0.95\linewidth]{images/legend.pdf}

    \vspace{-3pt}

    \caption{
        Complementary fine-grained performance of representative LMMs across
the remaining six diagnostic dimensions.
    }
    \label{fig:detail_analysis}
\end{figure}

Figures~\ref{fig:attribute} and~\ref{fig:detail_analysis} jointly
report the fine-grained performance of six representative frontier
LMMs across all ten diagnostic dimensions. Figure~\ref{fig:attribute}
highlights four representative dimensions in the main paper, whereas
Figure~\ref{fig:detail_analysis} complements the analysis with the
remaining six dimensions: image quality, lighting and screen, mixed
layout, layout structure, imaging method, and geometry and arrangement.
Each value denotes the average sample-level score over samples assigned
to the corresponding category. Categories with fewer than five valid
samples for a model are omitted to avoid conclusions based on
insufficient observations.

Overall, Qwen3.7-Plus exhibits the strongest and most consistent
performance across most of the six complementary dimensions, while the
relative rankings of the other models vary with document conditions.
Performance differences are generally limited under common conditions,
such as clear images, ordinary lighting, non-mixed layouts, and
horizontal text arrangements. The gaps become more pronounced under
degraded visual quality, complex document organization, and irregular
geometric arrangements, showing that similar aggregate scores may hide
substantially different robustness profiles.

The image-quality and lighting dimensions reveal clear sensitivity to
acquisition-related degradation. Low-resolution and blurred images are
more challenging than clear documents for most models, while shadows
and other lighting artifacts also introduce noticeable performance
variation. By contrast, the imaging-method dimension is comparatively
stable across photographs, scans, screenshots, and reproduced images,
although the relative advantage of individual models still changes
across acquisition sources.

Document organization produces more distinct model-specific behavior.
Same-line mixing and block-based multilingual layouts are generally more
difficult than non-mixed or annotation-style layouts. Within the
layout-structure dimension, forms, lists, monospace documents, and mixed
text--image pages expose larger differences than conventional
paragraphs or charts. Performance is relatively stable on horizontal
layouts, whereas dense small text, rotation, perspective distortion,
and multi-directional arrangements remain challenging. Together with
the four dimensions analyzed in Figure~\ref{fig:attribute}, these
results provide a complete view of model robustness across the ten
diagnostic dimensions.

\subsection{Detailed Failure Analysis}
\label{app:failure_analysis}

To provide a more concrete understanding of the limitations of current LMMs, we summarize 13 representative failure types across the five document-processing tracks. These cases reveal that model errors arise not only from inaccurate text recognition, but also from failures in structural parsing, spatial localization, field association, and document-level reasoning.

\begin{figure}[!p]
    \centering
    \includegraphics[width=\columnwidth]
    {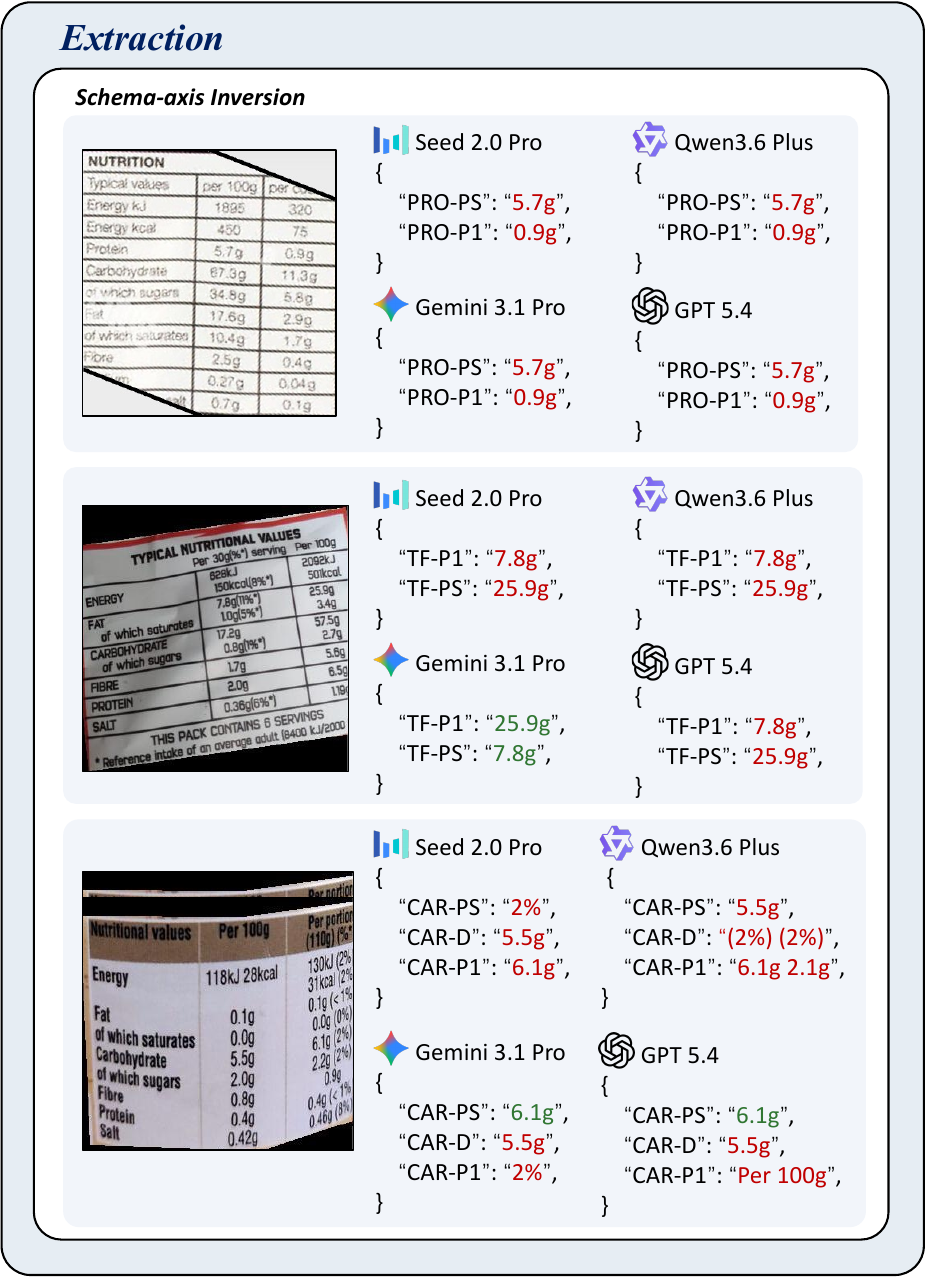}
    \caption{
        Representative cases of schema-axis inversion in the
        extraction track.
    }
    \label{fig:failure_schema}
\end{figure}

\begin{figure}[!p]
    \centering
    \includegraphics[width=\columnwidth]
    {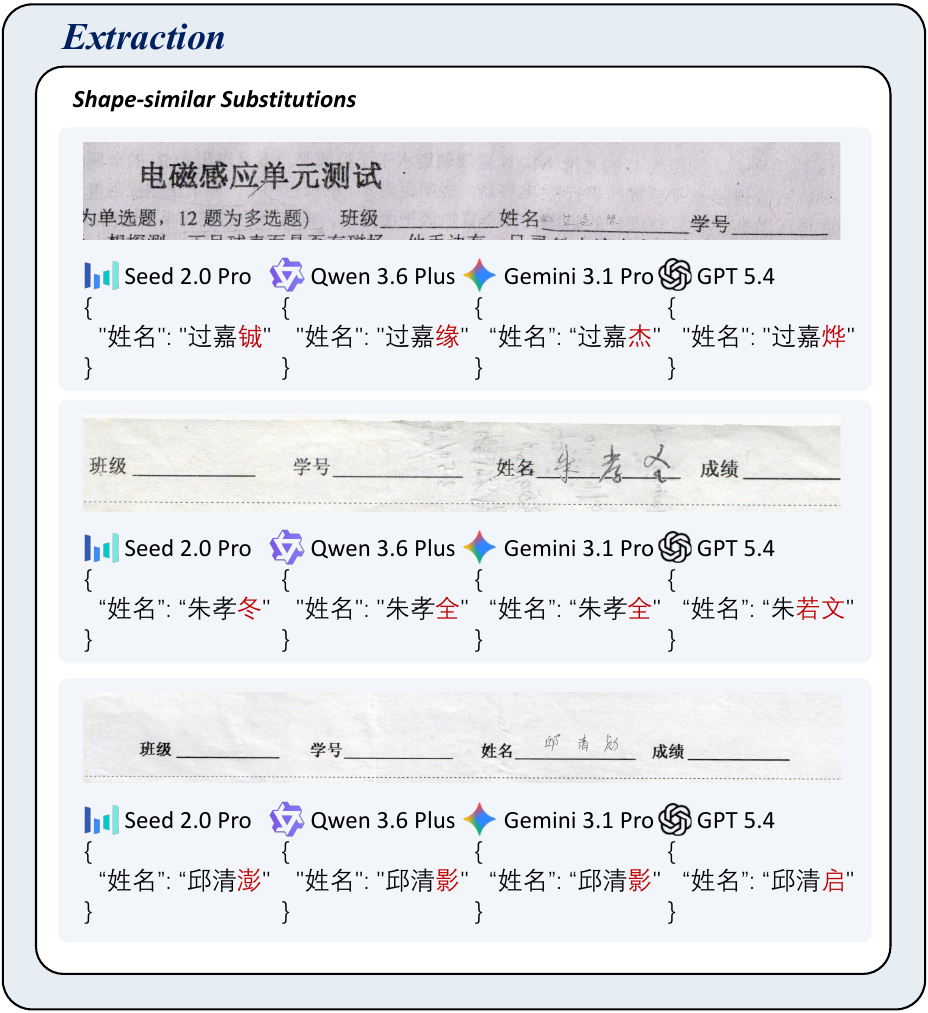}
    \caption{
        Representative cases of shape-similar substitution in the
        extraction track.
    }
    \label{fig:failure_shape_similar}
\end{figure}

\paragraph{Extraction.}
Figures~\ref{fig:failure_schema} and \ref{fig:failure_shape_similar} present two typical extraction errors.
Schema-axis inversion occurs when a model correctly recognizes the values in a table but associates them with incorrect row or column headers. Shape-similar substitution instead results from confusion between visually similar characters, leading to incorrect extracted field values despite successful localization of the relevant region.

\begin{figure}[!p]
    \centering
    \includegraphics[width=\columnwidth]
    {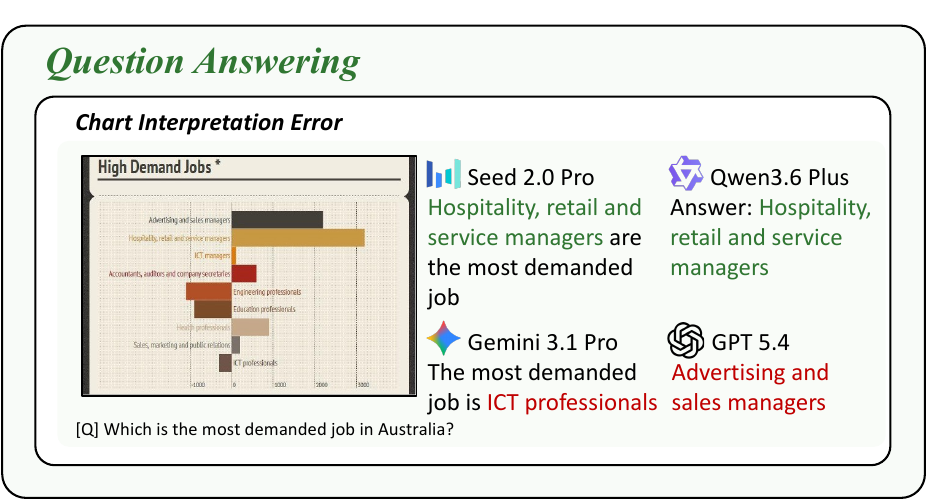}
    \caption{
        Representative cases of chart-interpretation errors in the
        question-answering track.
    }
    \label{fig:failure_chart_interpretation}
\end{figure}

\begin{figure}[!p]
    \centering
    \includegraphics[width=\columnwidth]
    {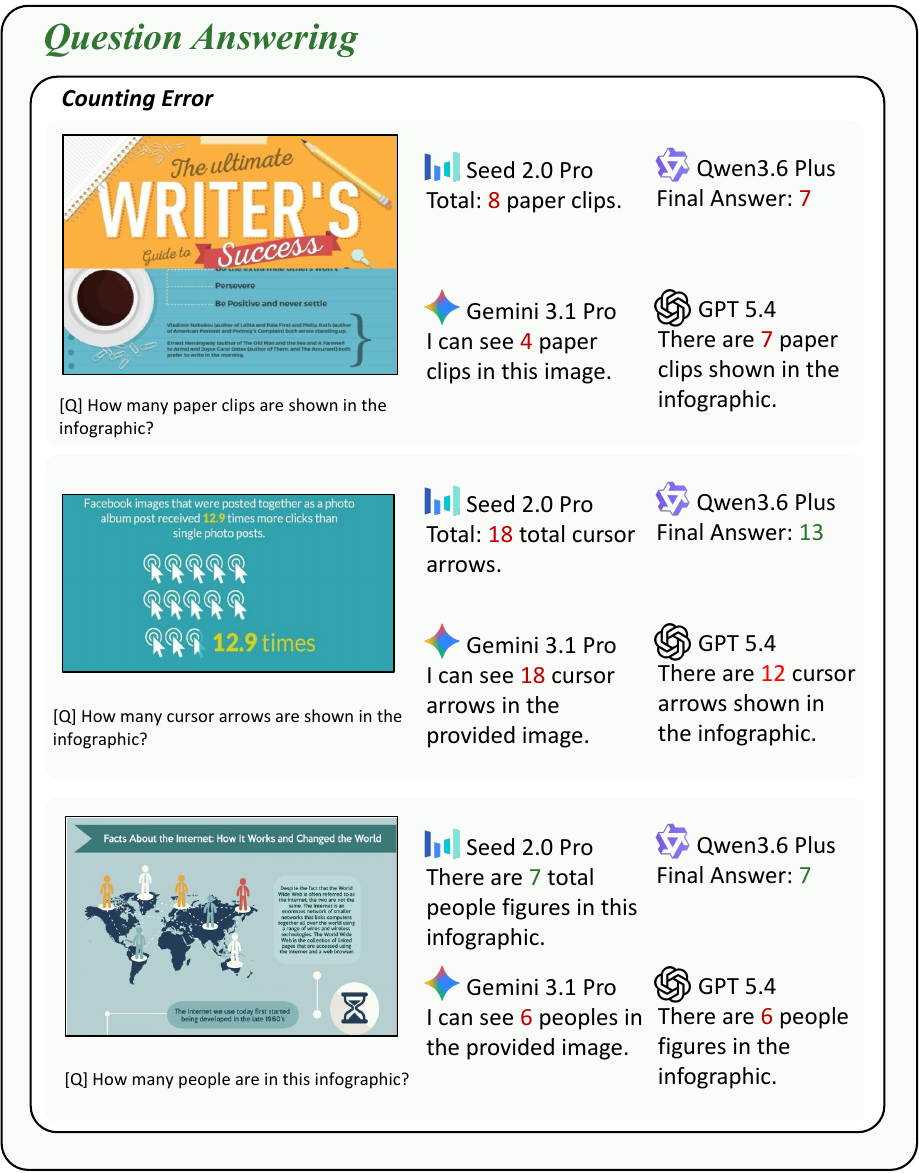}
    \caption{
        Representative cases of counting errors in the
        question-answering track.
    }
    \label{fig:failure_counting}
\end{figure}

\begin{figure}[!p]
    \centering
    \includegraphics[width=\columnwidth]
    {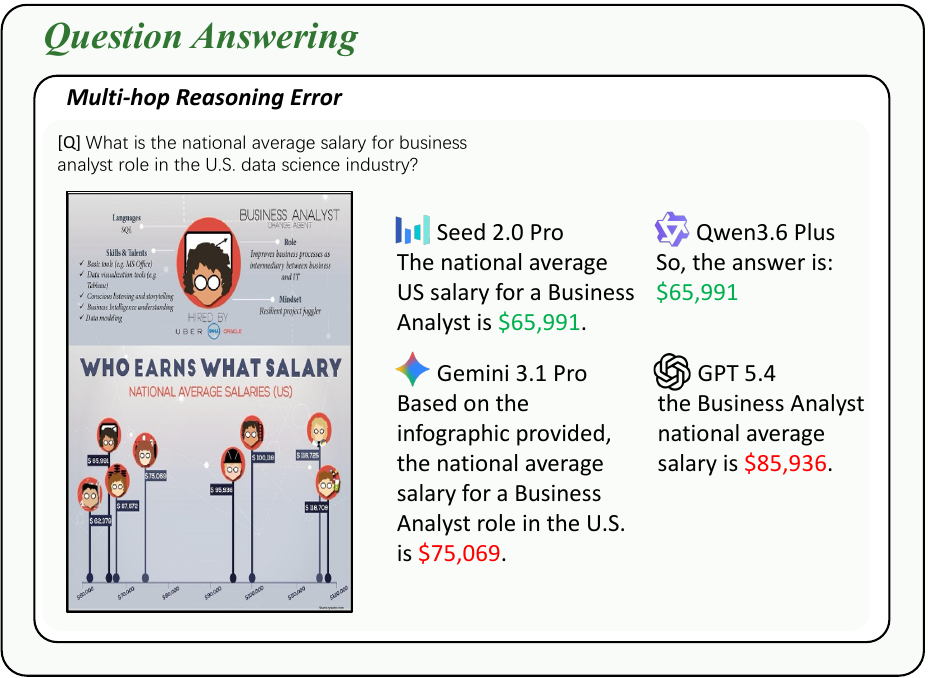}
    \caption{
        Representative cases of multi-hop reasoning errors in the
        question-answering track.
    }
    \label{fig:failure_multihop}
\end{figure}

\paragraph{Question Answering.}
Figures~\ref{fig:failure_chart_interpretation},
\ref{fig:failure_counting}, and \ref{fig:failure_multihop} illustrate failures in chart interpretation, visual counting, and multi-hop reasoning, respectively. These examples show that recognizing local text is insufficient for reliable document question answering. Models may misunderstand visual relations in a
chart, miscount repeated graphical elements, or fail to integrate evidence distributed across multiple regions of a document.

\begin{figure}[!p]
    \centering
    \includegraphics[width=\columnwidth]
    {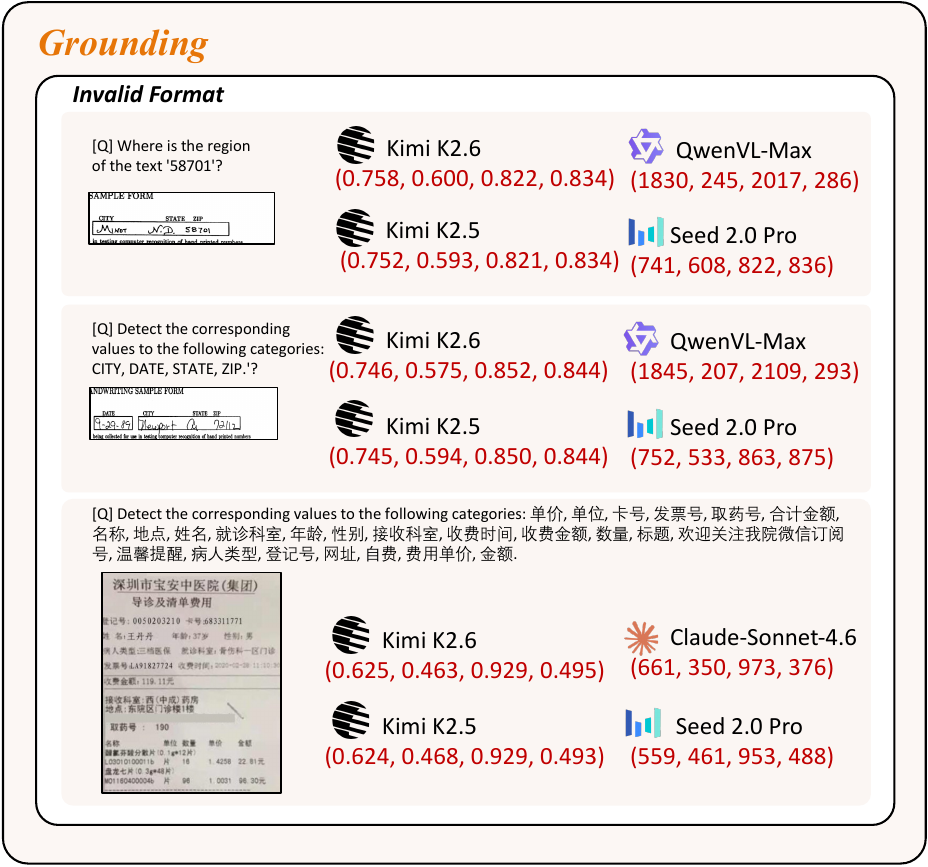}
    \caption{
        Representative cases of invalid output formats in the
        grounding track.
    }
    \label{fig:failure_invalid_format}
\end{figure}

\begin{figure}[!p]
    \centering
    \includegraphics[width=\columnwidth]
    {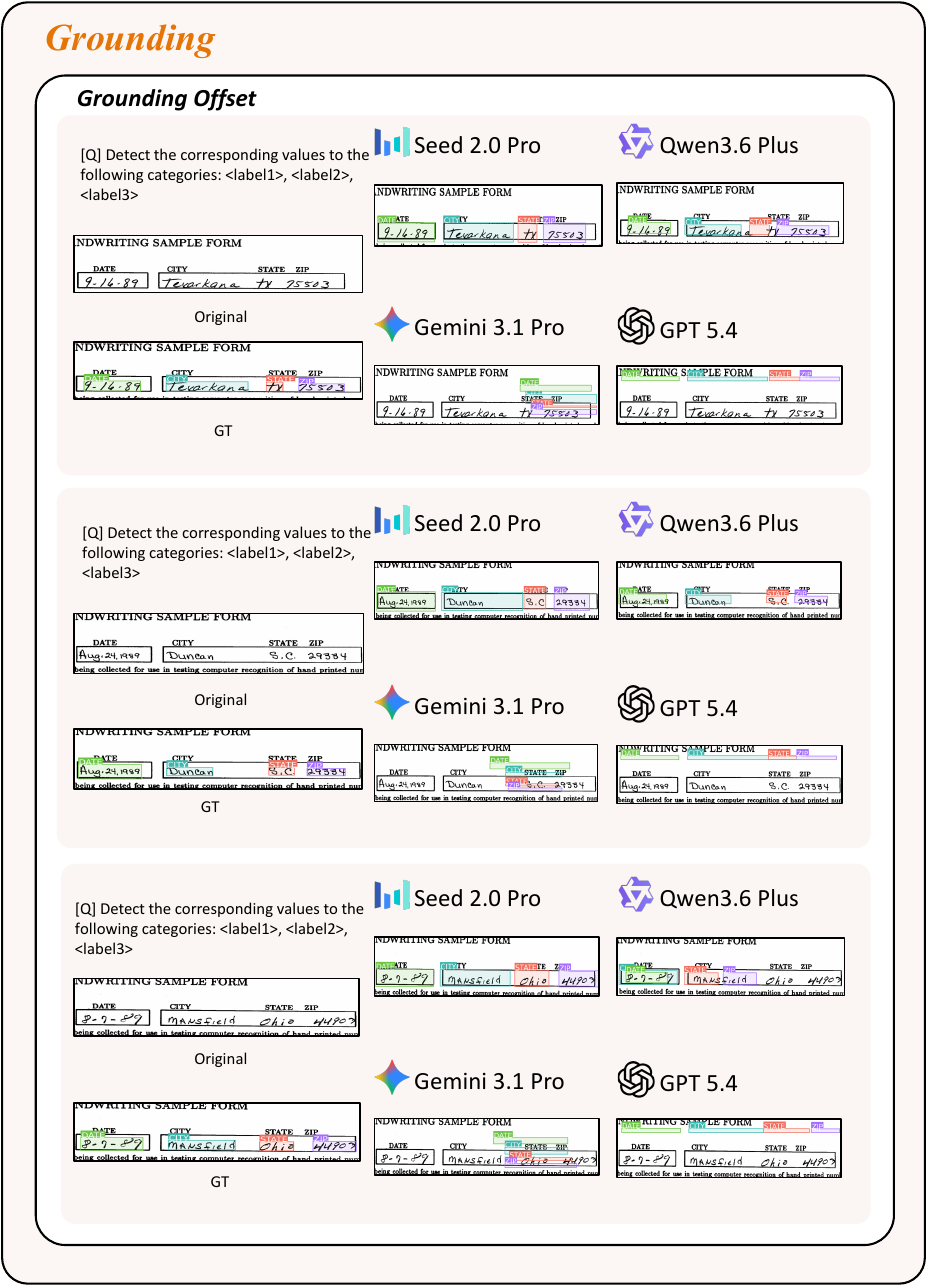}
    \caption{
        Representative cases of grounding offset in the
        grounding track.
    }
    \label{fig:failure_grounding_offset}
\end{figure}

\paragraph{Grounding.}
Figures~\ref{fig:failure_invalid_format} and \ref{fig:failure_grounding_offset} show two major grounding failures. In the invalid-format cases, models identify the target content but return coordinates using an incorrect scale or representation. In the grounding-offset cases, the predicted bounding boxes are spatially
shifted and do not precisely cover the requested text or object.

\begin{figure}[!p]
    \centering
    \includegraphics[width=\columnwidth]
    {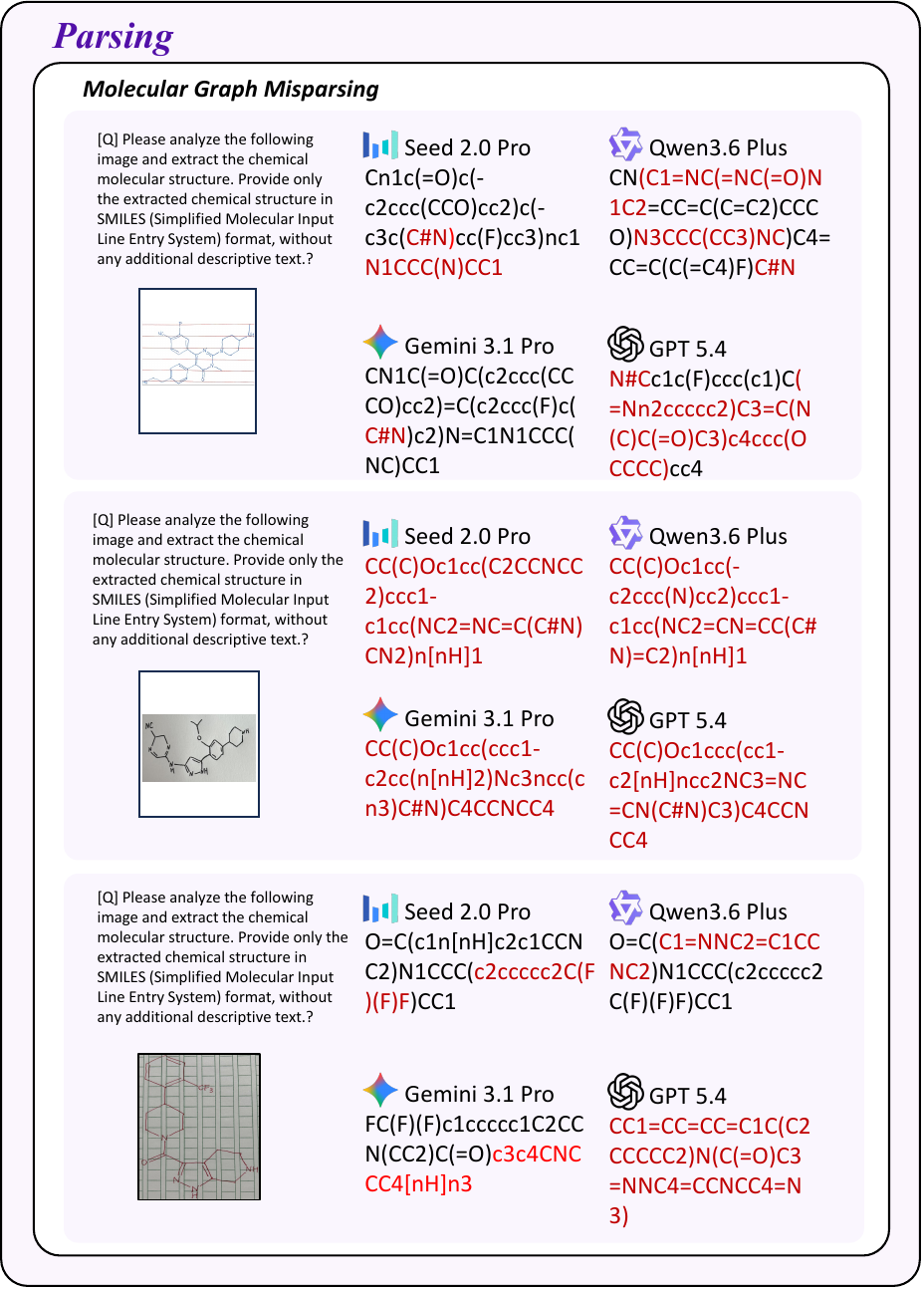}
    \caption{
        Representative cases of molecular graph misparsing in the
        parsing track.
    }
    \label{fig:failure_molecular_graph_misparsing}
\end{figure}

\begin{figure}[!p]
    \centering
    \includegraphics[width=\columnwidth]
    {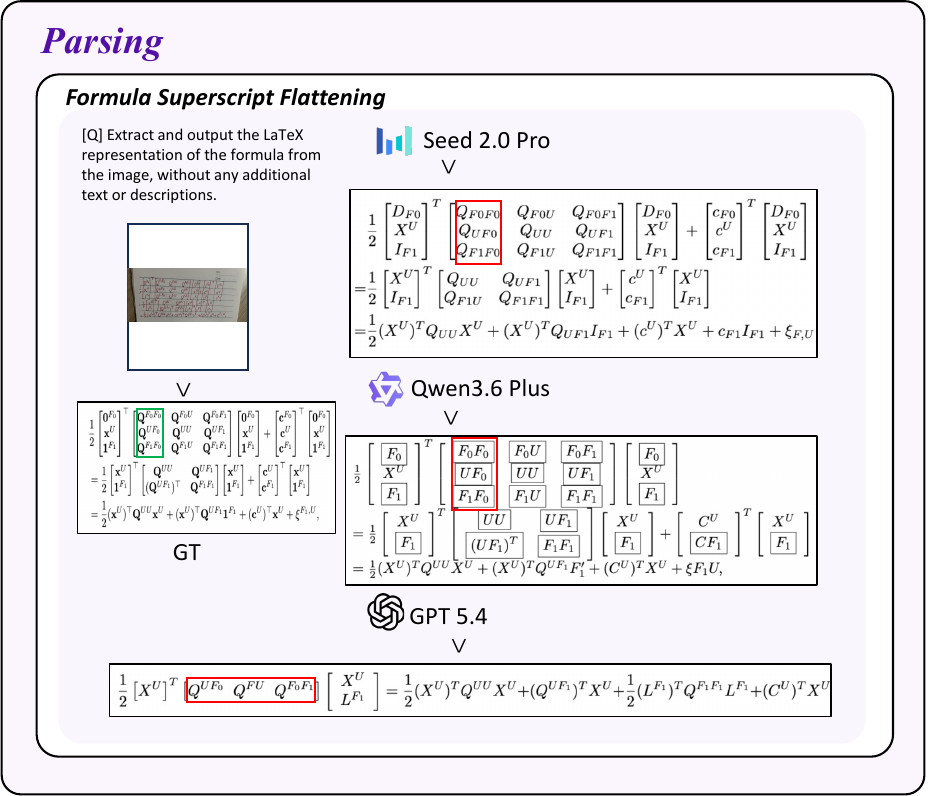}
    \caption{
        Representative cases of formula superscript flattening in the
        parsing track.
    }
    \label{fig:failure_formula_superscript_flattening}
\end{figure}

\begin{figure}[!p]
    \centering
    \includegraphics[width=\columnwidth]
    {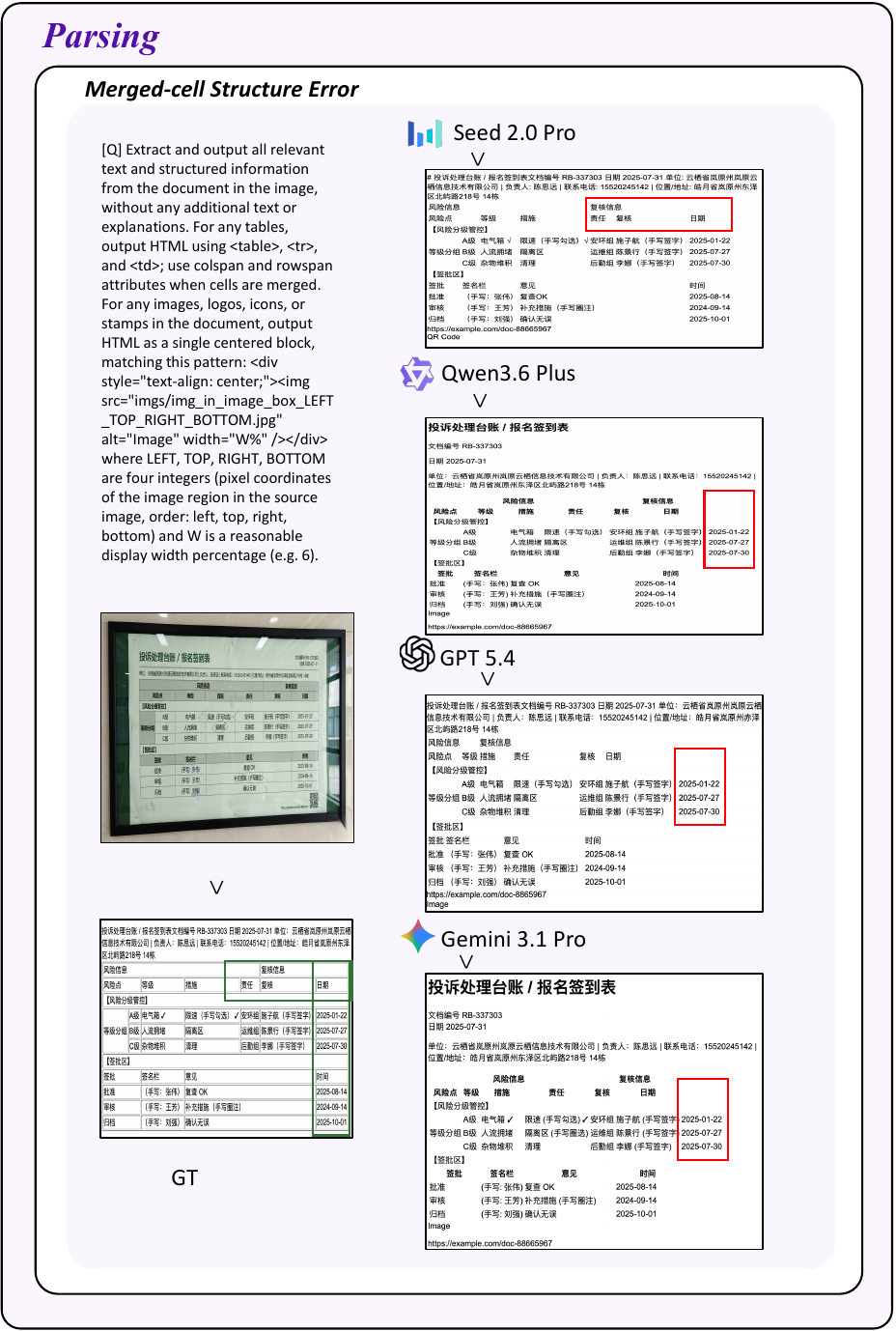}
    \caption{
        Representative cases of merged-cell structure errors in the
        parsing track.
    }
    \label{fig:failure_merge_cell_error}
\end{figure}

\paragraph{Parsing.}
Figures~\ref{fig:failure_molecular_graph_misparsing}, \ref{fig:failure_formula_superscript_flattening}, and
\ref{fig:failure_merge_cell_error} demonstrate errors in recovering complex document structures. Models may incorrectly convert molecular graphs into SMILES strings, flatten superscript or hierarchical relations in mathematical formulas, or fail to preserve merged-cell dependencies when converting tables into structured markup.

\begin{figure}[!p]
    \centering
    \includegraphics[width=\columnwidth]
    {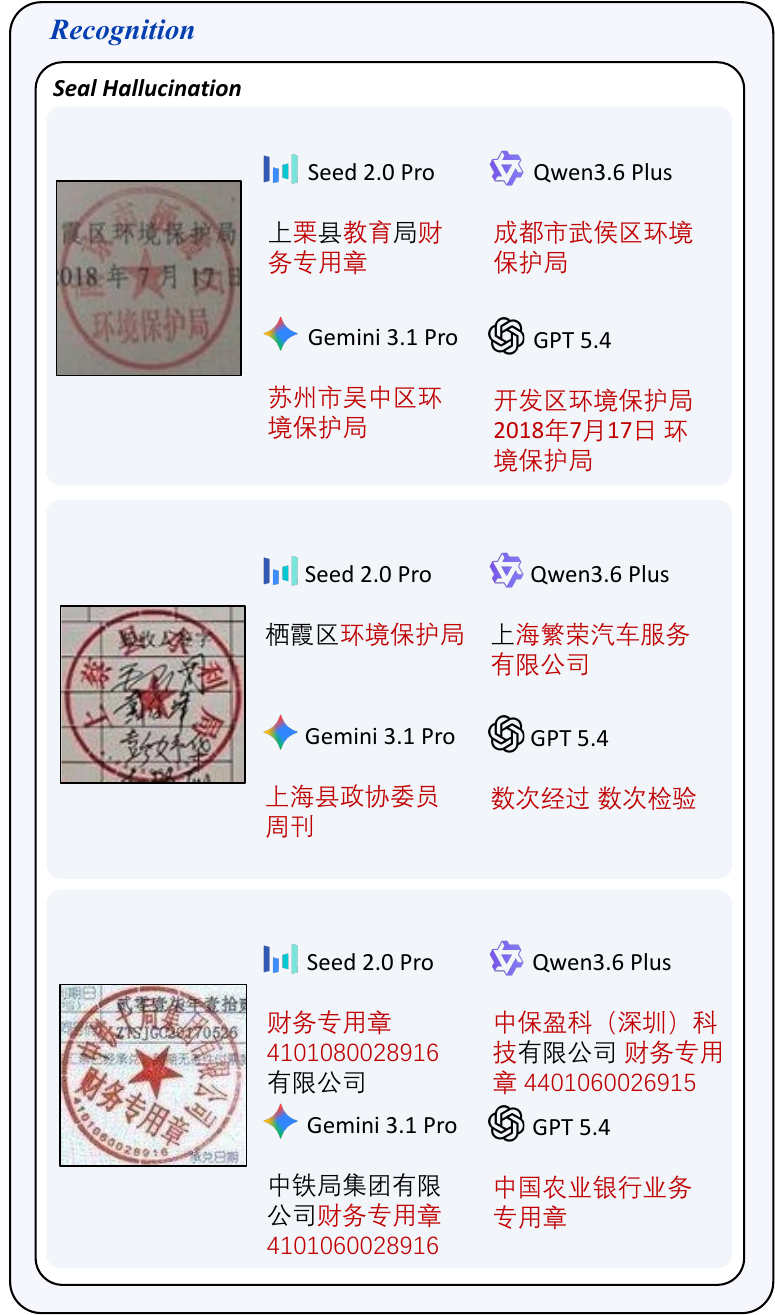}
    \caption{
        Representative cases of seal hallucination in the
        recognition track.
    }
    \label{fig:failure_seal_hallucination}
\end{figure}

\begin{figure}[!p]
    \centering
    \includegraphics[width=\columnwidth]
    {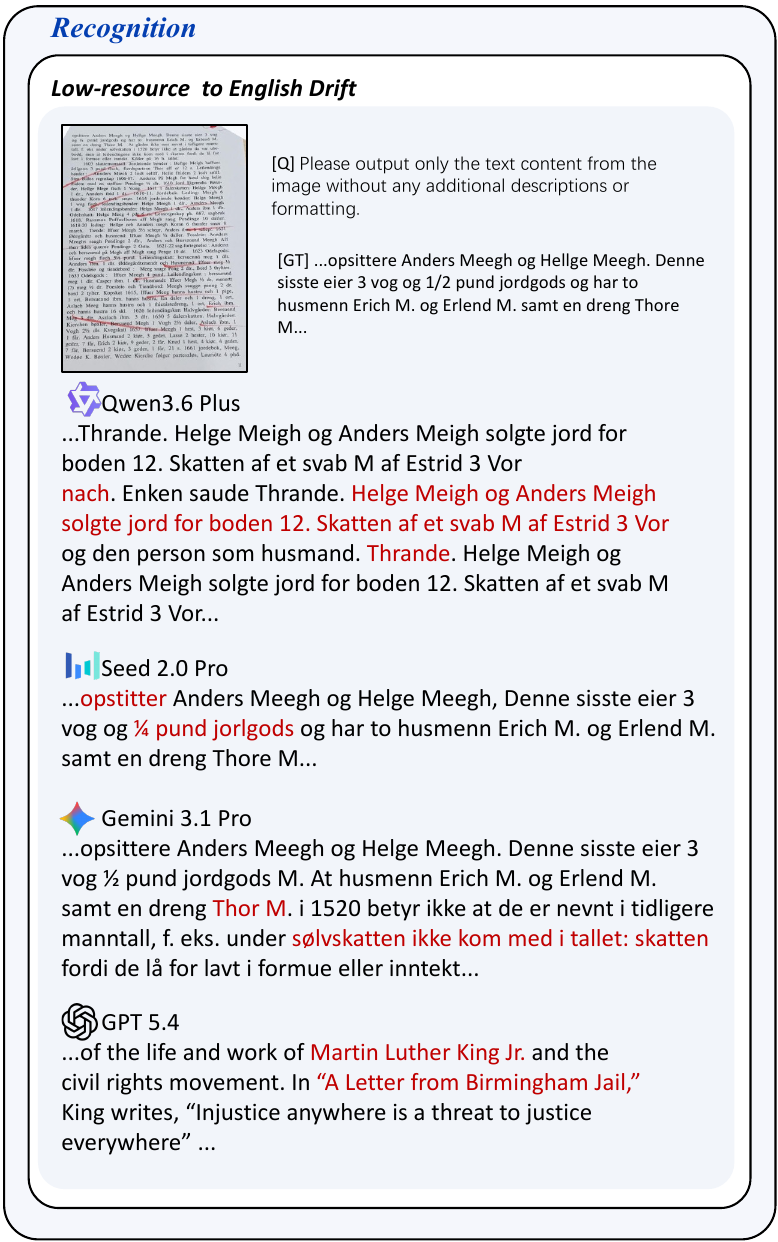}
    \caption{
        Representative cases of low-resource-language-to-English
        drift in the recognition track.
    }
    \label{fig:failure_low_resource_to_english_drift}
\end{figure}

\begin{figure}[!p]
    \centering
    \includegraphics[width=\columnwidth]
    {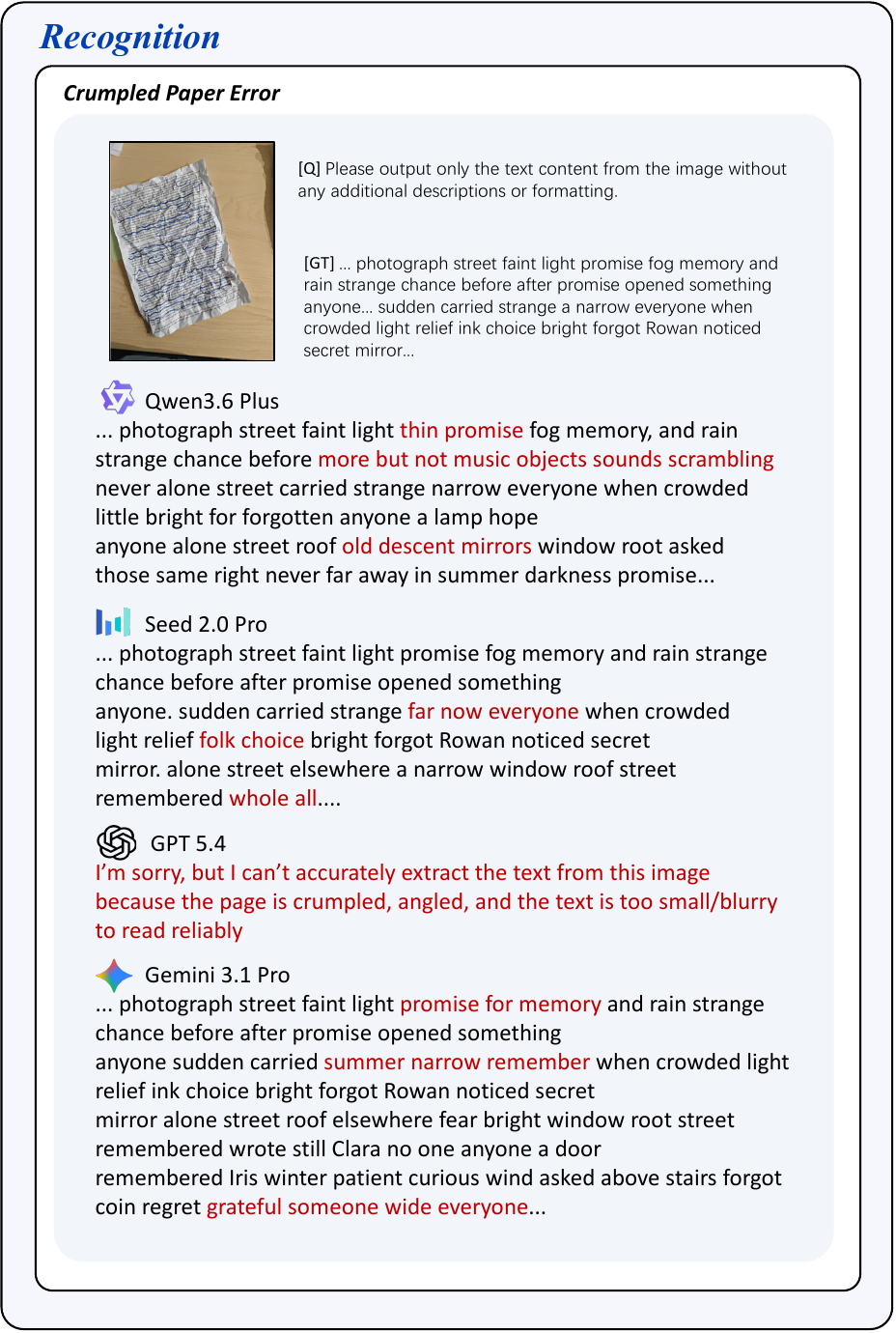}
    \caption{
        Representative cases of recognition errors on crumpled
        documents.
    }
    \label{fig:failure_crumpled_paper_error}
\end{figure}

\paragraph{Recognition.}
Figures~\ref{fig:failure_seal_hallucination}, \ref{fig:failure_low_resource_to_english_drift}, and \ref{fig:failure_crumpled_paper_error} present representative recognition failures. Ambiguous seal patterns frequently induce hallucinated but visually plausible text. For low-resource scripts, some models drift toward unrelated English content rather than faithfully transcribing the document. Severe deformation, blur, and irregular text orientation in crumpled documents further result in substantial omissions and substitutions.

Overall, these cases demonstrate that strong aggregate performance does not guarantee reliable behavior under realistic document conditions. Current LMMs remain sensitive to fine-grained visual ambiguity, complex structural dependencies, inaccurate spatial representations, and reasoning over distributed evidence.

\subsection{Evaluation Metrics}
\label{app:metrics}

We adopt task-specific evaluation metrics for different OCR-centric tracks in CC-OCR v2. For the $i$-th sample, we denote the document image as $\mathcal{D}_i$, the task instruction as $\mathcal{I}_i$, the model prediction as $\mathcal{Y}_i=f_\theta(\mathcal{D}_i,\mathcal{I}_i)$, and the corresponding ground truth as $\mathcal{Y}_i^*$. All scores are normalized to $[0,1]$ during evaluation and multiplied by $100$ when reported in the main tables.

\textbf{Recognition.}
To comprehensively evaluate text recognition in multilingual and multi-scene scenarios, we follow the full-text multi-set matching protocol used in CC-OCR~\cite{yang2025cc}. This protocol reduces the influence of reading-order variations in LMM outputs and focuses on whether the textual units in the image are correctly recognized.

Specifically, for each sample, we first normalize the predicted text $\mathcal{Y}_i^{\mathrm{rec}}$ and the ground-truth text $\mathcal{Y}_i^{\mathrm{rec},*}$ by replacing tabs and newlines with spaces, collapsing consecutive spaces, removing empty tokens, and converting text to lowercase. For word-level subsets, texts are split by whitespace. For character-level subsets, texts are split into characters after removing blank spaces. For multi-scene word-level recognition, non-alphanumeric characters are further removed before matching.

Let the ground-truth text and prediction be split into two multi-sets of basic units:
\begin{equation}
U_i^*=\{(u,c_i^*(u))\}, \quad U_i=\{(u,c_i(u))\},
\end{equation}
where $u$ denotes a basic unit, and $c_i^*(u)$ and $c_i(u)$ denote the occurrence counts of $u$ in the ground truth and prediction, respectively. Two units are considered matched when they are identical after normalization. The number of matched units for the $i$-th sample is calculated as:
\begin{equation}
    m_i =\sum_{u \in U_i^* \cup U_i} \min \left(c_i^*(u), c_i(u)\right).
\end{equation}

We compute micro-averaged recall and precision over the whole evaluation set:
\begin{equation}
\mathrm{Recall}_{\mathrm{rec}} = \frac{\sum_i m_i}{\sum_i \sum_{u} c_i^*(u)},
\end{equation}
\begin{equation}
\mathrm{Precision}_{\mathrm{rec}} =
\frac{\sum_i m_i}{\sum_i \sum_{u} {c}_i(u)}.
\end{equation}
The recognition score is the micro-F1:
\begin{equation}
\mathrm{F1}_{\mathrm{rec}} = \frac{ 2 \times \mathrm{Recall}_{\mathrm{rec}} \times \mathrm{Precision}_{\mathrm{rec}} }{ \mathrm{Recall}_{\mathrm{rec}} + \mathrm{Precision}_{\mathrm{rec}}}.
\end{equation}
The evaluator also records macro-F1 over samples for diagnostic analysis, while the primary recognition score reported in CC-OCR v2 is the micro-F1.

\textbf{Parsing.}
The parsing track contains multiple types of structured outputs, including LaTeX, HTML, SMILES strings, and mixed HTML-like code. Due to the differences in output formats, we use different evaluation metrics for different parsing subtasks.

For general document parsing, formula parsing, and molecular parsing, we evaluate the similarity between the prediction $\mathcal{Y}_i^{\mathrm{parse}}$ and the ground truth $\mathcal{Y}_i^{\mathrm{parse},*}$ using normalized edit distance (NED)~\cite{levenshtein1966binary}. Before evaluation, we apply format-specific normalization. For document parsing, LaTeX wrappers, model chatter, spaces, and newlines are removed. For formula parsing, LaTeX code fences, tabs, spaces, and line breaks are removed. For molecular parsing, spaces, line breaks, and SMILES tags are removed.

Let ${\mathcal{Y}}_i^{\mathrm{parse}}$ and ${\mathcal{Y}}_i^{\mathrm{parse},*}$ denote the normalized prediction and ground truth. The normalized edit similarity is:
\begin{equation}
    \mathrm{NED}_i = 1 -\frac{\mathrm{EditDist}({\mathcal{Y}}_i^{\mathrm{parse}},{\mathcal{Y}}_i^{\mathrm{parse},*})}{\max(\mathrm{len}({\mathcal{Y}}_i^{\mathrm{parse}}),\mathrm{len}({\mathcal{Y}}_i^{\mathrm{parse},*}))}.
\end{equation}
The score of these parsing subtasks is the average normalized edit similarity over all samples:
\begin{equation}
    \mathrm{Score}_{\mathrm{NED}} =\frac{1}{N}\sum_{i=1}^{N}\mathrm{NED}_i .
\end{equation}

For complex table parsing, both predictions and ground truths are represented as HTML tables. We use Tree Edit Distance-based Similarity (TEDS)~\cite{zhong2020image} to measure both the structural correctness of the table and the accuracy of cell contents. Let ${T}_i$ and $T_i^*$ denote the predicted and ground-truth HTML trees, respectively. The TEDS score is:
\begin{equation}
    \mathrm{TEDS}_i =1 - \frac{\mathrm{TreeEditDist}({T}_i,T_i^*)}{\max(|{T}_i|,|T_i^*|)},
\end{equation}
where $|\cdot|$ denotes the number of nodes in the HTML tree. The table parsing score is:
\begin{equation}
    \mathrm{Score}_{\mathrm{TEDS}} =\frac{1}{N}\sum_{i=1}^{N}\mathrm{TEDS}_i .
\end{equation}

For information board parsing, the target output contains both free-form textual content and structured tables. Therefore, we
separately evaluate the non-table textual part and the table part. The textual part is extracted from the HTML-like output after removing table blocks and is evaluated by character-level micro-F1, denoted as $\mathrm{F1}^{\mathrm{text}}_i$. The table part is evaluated by TEDS, denoted as $\mathrm{TEDS}^{\mathrm{table}}_i$. If both the prediction and ground truth contain no table, the table score is set to $1$; If only one side contains a table, the table score is set to $0$. For samples with multiple tables, TEDS is computed by index-wise table matching and then averaged. The final information board score is:
\begin{equation}
    \mathrm{Score}^{\mathrm{info}}_i = (1-w)\cdot \mathrm{F1}^{\mathrm{text}}_i + w\cdot \mathrm{TEDS}^{\mathrm{table}}_i,
\end{equation}
where $w=0.9$ in the official evaluator. The final score is the average of $\mathrm{Score}^{\mathrm{info}}_i$ over all information board samples.

\textbf{Grounding.}
The grounding track evaluates whether LMMs can localize the target text region or semantic object in document images. For text grounding, the model is required to output one bounding box. For object grounding, the model outputs a list of labeled bounding boxes. In both cases, the evaluator parses the predicted box into the format $[x_{\min},y_{\min},x_{\max},y_{\max}]$.

The predicted coordinates are interpreted as normalized coordinates in the $0$--$1000$ range and are converted into pixel coordinates according to the image width and height. Given a predicted box ${B}$ and a ground-truth box $B^*$, we compute Intersection over Union (IoU) as:
\begin{equation}
    \mathrm{IoU}({B},B^*) = \frac{ \mathrm{area}({B}\cap B^*) }{ \mathrm{area}({B}\cup B^*) }.
\end{equation}

For text grounding, each sample has one target box. If the prediction cannot be parsed into a valid bounding box, the sample score is set to $0$. Otherwise, the sample score is:
\begin{equation}
    s_i^{\mathrm{text}} = \mathrm{IoU}({B}_i,B_i^*).
\end{equation}

For object grounding, each sample may contain multiple semantic objects, where the ground truth is:
\begin{equation}
    \mathcal{Y}_i^{\mathrm{ground},*} = \{(\ell_{ij},B_{ij}^*)\}_{j=1}^{M_i},
\end{equation}
where $\ell_{ij}$ is the object label and $B_{ij}^*$ is the corresponding bounding box. The prediction is parsed into a set of labeled boxes. Object labels are matched by case-insensitive exact matching. For each ground-truth label, if the corresponding predicted label is missing, its IoU is set to $0$; otherwise, its IoU is computed against the matched predicted box. The sample-level object grounding score is:
\begin{equation}
    s_i^{\mathrm{obj}} = \frac{1}{M_i} \sum_{j=1}^{M_i} \mathbf{1}[\ell_{ij}\in {\mathcal{L}}_i]\, \mathrm{IoU}({B}_{ij},B_{ij}^*),
\end{equation}
where ${\mathcal{L}}_i$ is the set of predicted labels.

For each third-level grounding subset $d$, we compute the mean IoU:
\begin{equation}
    \mathrm{mIoU}_d = \frac{1}{N_d}\sum_{i=1}^{N_d}s_i,
\end{equation}
where $s_i$ denotes either $s_i^{\mathrm{text}}$ or $s_i^{\mathrm{obj}}$ depending on the grounding task. We also report IoU-based accuracy at threshold $0.5$:
\begin{equation}
    \mathrm{Acc@0.5}_d = \frac{1}{N_d} \sum_{i=1}^{N_d} \mathbf{1}[s_i \geq 0.5].
\end{equation}
The primary grounding score is the macro-average of $\mathrm{mIoU}_d$ over all third-level text-grounding and object-grounding subsets:
\begin{equation}
    \mathrm{Score}_{\mathrm{ground}} =
    \frac{1}{|\mathcal{D}_{\mathrm{ground}}|}
    \sum_{d\in\mathcal{D}_{\mathrm{ground}}}
    \mathrm{mIoU}_d .
\end{equation}

\textbf{Extraction.}
For key information extraction, model predictions and ground truths are represented as JSON objects. We first parse the prediction into JSON format. Markdown-style JSON code fences are removed before parsing. Invalid JSON predictions are treated as unmatched predictions.

Before scoring, all string values are normalized by converting full-width characters to half-width characters, removing unnecessary spaces, normalizing several punctuation variants, and stripping trailing punctuation that does not affect the field semantics. The same normalization is applied to both predictions and ground truths.

Since the extraction schemas may contain nested dictionaries and lists, we flatten each JSON object into a list of field-value pairs. For example, a nested structure such as \texttt{\{"menu": [\{"name": "cake"\}]\}} is flattened as \texttt{("menu.name", "cake")}. We denote the flattened prediction and ground truth of the $i$-th sample as ${\Phi}_i$ and ${\Phi}_i^*$, respectively.

A predicted field is counted as a true positive only when the complete
normalized key--value pair exactly matches an entry in the ground
truth. Otherwise, it is counted as a false positive, while unmatched
ground-truth fields are counted as false negatives. Aggregating over
all samples, we compute the field-level micro-F1 score as:
\begin{equation}
    \mathrm{F1}_{\mathrm{extract}}
    =
    \frac{2\mathrm{TP}}
    {2\mathrm{TP}+\mathrm{FP}+\mathrm{FN}+\epsilon}.
\end{equation}
This strict field-level matching penalizes errors in either keys or
values. Even a single-character error in a normalized value causes the
corresponding field to be treated as unmatched.

\textbf{Question Answering.}
For document question answering, the model generates a textual answer $\mathcal{Y}_i^{\mathrm{qa}}$ for each question. Each sample may contain one or multiple reference answers, denoted as:
\begin{equation}
    \mathcal{A}_i^* = \{a_{i1}^*,a_{i2}^*,\ldots,a_{iR_i}^*\}.
\end{equation}
If multiple reference answers are provided, the sample score is the maximum score over all references.

The evaluator supports English, Chinese, and case-sensitive evaluation modes. In the default English mode, both predictions and answers are lowercased, leading and trailing spaces are stripped, and newlines are replaced with spaces. In the Chinese mode, spaces are additionally removed. The evaluation mode is automatically selected according to whether the reference answer contains Chinese characters, unless otherwise specified.

For a normalized prediction ${a}_i$ and a normalized reference answer $a_{ij}^*$, the normalized Levenshtein similarity is computed as:
\begin{equation}
\mathrm{NLS}({a}_i,a_{ij}^*) =1 -\frac{\mathrm{EditDist}({a}_i,a_{ij}^*)}{\max(\mathrm{len}({a}_i),\mathrm{len}(a_{ij}^*))}.
\end{equation}

For short answers, the evaluator uses containment-based exact matching. In the English mode, an answer is regarded as short if it contains fewer than five whitespace-separated tokens. In the Chinese mode, the evaluator uses comma-separated answer fragments and regards an answer as short if it contains fewer than four fragments. A short answer receives a score of $1$ if the normalized reference answer appears in the normalized prediction; otherwise, it receives a score of $0$:
\begin{equation}
q_{\mathrm{short}}(a_i,a_{ij}^*) =
\mathbf{1}\left[a_{ij}^* \subseteq a_i\right].
\end{equation}

For long-form answers, the evaluator uses normalized Levenshtein similarity with a threshold of $0.5$. Let
$n_{ij}=\mathrm{NLS}(a_i,a_{ij}^*)$. The long-answer score is:
\begin{equation}
q_{\mathrm{long}}(a_i,a_{ij}^*) = \begin{cases}
n_{ij}, & n_{ij} \geq 0.5,\\
0, & n_{ij} < 0.5.
\end{cases}
\end{equation}
The sample-level QA score is:
\begin{equation}
    s_i^{\mathrm{qa}} = \max_{a_{ij}^*\in\mathcal{A}_i^*} q({a}_i,a_{ij}^*).
\end{equation}
The final QA score is the average over all QA samples:
\begin{equation}
    \mathrm{Score}_{\mathrm{qa}} =\frac{1}{N}\sum_{i=1}^{N}s_i^{\mathrm{qa}}.
\end{equation}

In addition, the evaluator reports exact accuracy and accuracy at threshold $0.5$:
\begin{equation}
    \mathrm{Acc}_{\mathrm{exact}} =\frac{1}{N}\sum_{i=1}^{N}\mathbf{1}[s_i^{\mathrm{qa}}=1],
\end{equation}
\begin{equation}
    \mathrm{Acc@0.5}_{\mathrm{qa}} = \frac{1}{N}\sum_{i=1}^{N}\mathbf{1}[s_i^{\mathrm{qa}}\geq 0.5].
\end{equation}

\subsection{Detailed Configurations for LMMs}
\label{app:config}
Table~\ref{tab:model_details} summarizes the model information of all evaluated LMMs, including the model vendor, official identifier, and release date. We divide the evaluated models into two deployment settings: on-device LMMs and on-server LMMs. The on-device group contains open-weight models that can be deployed locally, including Qwen3.5-9B~\cite{bai2025qwen3}, MiniCPM-o 4.5-8B~\cite{yu2025minicpm}, InternVL3.5-8B~\cite{wang2025internvl3}, and Step3-VL-10B~\cite{huang2026step3}. The on-server group contains proprietary frontier LMMs accessed through their official API endpoints, including GPT-5.5, Qwen-VL-Max, Claude Sonnet 4.6, Claude Opus 4.8, Gemini 3.1 Flash, Gemini 3.1 Pro, Kimi K2.6, Seed 2.0 Pro, Qwen3.6-Plus, and Qwen3.7-Plus.

All models follow the same inference protocol. For each evaluation sample, the task-specific instruction is loaded from the corresponding question file and paired with the associated document image(s). For multi-page samples, all pages are sent in their original page order. Images are converted to RGB JPEG format and encoded as base64 before being sent to the model. The image content and the textual instruction are combined into a single user message. No additional system prompt is used.

We use deterministic decoding for all models by setting the temperature to $0$. When supported by the API or inference backend, we disable explicit thinking or reasoning mode to ensure that models return only the task output required by the prompt. We do not use self-consistency, majority voting, external OCR engines, retrieval augmentation, or post-processing with additional LLMs. The raw model response is saved as the prediction and directly passed to the official task-specific evaluator.

For API-based inference, we set the maximum concurrency to $8$ and the maximum number of retries to $10$ to handle temporary request failures. If a prediction file already exists and is non-empty, it can be skipped during reruns to avoid repeated API calls. These settings affect only inference efficiency and fault tolerance; they do not change the model input, decoding strategy, or evaluation result.
\begin{table*}[t]
\centering
\caption{Model Information of the Evaluated LMMs.}
\label{tab:model_details}

\begin{tabular}{l l p{0.45\linewidth} r}
\toprule
\textbf{Model} & \textbf{Vendor} & \textbf{Identifier} & \textbf{Release Date} \\
\midrule
\multicolumn{4}{l}{\textit{\textbf{On-Device LMMs}}} \\
\midrule
\raisebox{-0.2\height}{\includegraphics[height=1.2em]{logo/qwen.png}}\,Qwen3.5-9B               & Alibaba & Qwen/Qwen3.5-9B & 2026-03 \\
\raisebox{-0.2\height}{\includegraphics[height=1.2em]{logo/minicpm.png}}\,MiniCPM-o 4.5-8B          & ModelBest & openbmb/MiniCPM-o-4.5 & 2026-02 \\
\raisebox{-0.2\height}{\includegraphics[height=1.2em]{logo/shailab.png}}\,InternVL3.5-8B  & OpenGVLab & OpenGVLab/InternVL3.5-8B & 2025-12 \\
\raisebox{-0.2\height}{\includegraphics[height=1.2em]{logo/stepfun.png}}\,Step3-VL-10B  & StepFun & stepfun-ai/Step3-VL-10B & 2026-01 \\
\midrule
\multicolumn{4}{l}{\textit{\textbf{On-Server LMMs}}} \\
\midrule
\raisebox{-0.2\height}{\includegraphics[height=1.2em]{logo/openai.png}}\,GPT-5.4              & OpenAI & gpt-5.4 & 2026-03 \\
\raisebox{-0.2\height}{\includegraphics[height=1.2em]{logo/openai.png}}\,GPT-5.5              & OpenAI & gpt-5.5 & 2026-04 \\
\raisebox{-0.2\height}{\includegraphics[height=1.2em]{logo/qwen.png}}\,Qwen-VL-Max          & Alibaba & qwen-vl-max & 2025-08 \\
\raisebox{-0.2\height}{\includegraphics[height=1.2em]{logo/claude.png}}\,Claude Sonnet 4.6     & Anthropic & claude-sonnet-4-6 & 2026-02 \\
\raisebox{-0.2\height}{\includegraphics[height=1.2em]{logo/claude.png}}\,Claude Opus 4.6       & Anthropic & claude-opus-4-6 & 2026-02 \\
\raisebox{-0.2\height}{\includegraphics[height=1.2em]{logo/claude.png}}\,Claude Opus 4.8       & Anthropic & claude-opus-4-8 & 2026-05 \\
\raisebox{-0.2\height}{\includegraphics[height=1.2em]{logo/gemini.png}}\,Gemini 3.1 Flash     & Google & gemini-3.1-flash-lite-preview & 2026-03 \\
\raisebox{-0.2\height}{\includegraphics[height=1.2em]{logo/kimi.png}}\,Kimi K2.5             & Moonshot & kimi-k2.5 & 2026-01 \\
\raisebox{-0.2\height}{\includegraphics[height=1.2em]{logo/kimi.png}}\,Kimi K2.6             & Moonshot & kimi-k2.6 & 2026-04 \\
\raisebox{-0.2\height}{\includegraphics[height=1.2em]{logo/gemini.png}}\,Gemini 3.1 Pro        & Google & gemini-3.1-pro-preview & 2026-02 \\
\raisebox{-0.2\height}{\includegraphics[height=1.2em]{logo/bytedance.png}}\,Seed 2.0 Pro          & Bytedance & doubao-seed-2.0-pro & 2026-02 \\
\raisebox{-0.2\height}{\includegraphics[height=1.2em]{logo/qwen.png}}\,Qwen3.6-Plus          & Alibaba & qwen3.6-plus & 2026-04 \\
\raisebox{-0.2\height}{\includegraphics[height=1.2em]{logo/qwen.png}}\,Qwen3.7-Plus          & Alibaba & qwen3.7-plus & 2026-06 \\
\bottomrule
\end{tabular}

\end{table*}

\subsection{Task-wise Attention-based Error Analysis on CC-OCR v2}
\label{app:attention_analysis}

Beyond the output-level error patterns discussed in Appendix~\ref{app:failure_analysis}, we further examine whether representative failures arise from incorrect evidence localization or
from subsequent reasoning and decoding after the relevant region has already been identified. The attention maps reveal that attending to the correct visual evidence does not necessarily lead to a correct prediction.

\begin{figure}[!tbp]
    \centering
    \includegraphics[width=0.95\linewidth]
    {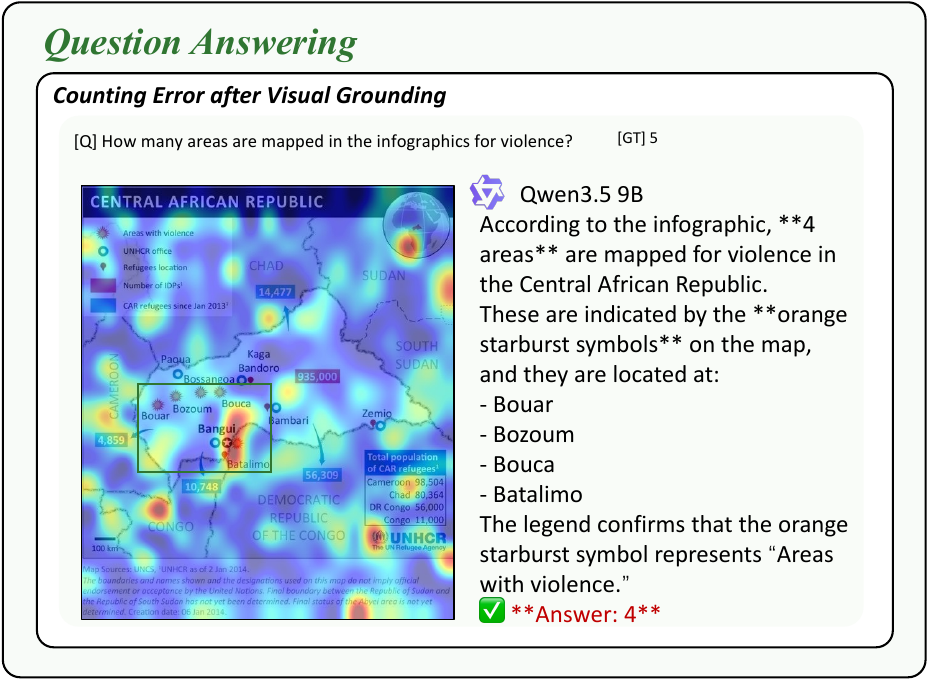}
    \caption{
        Attention-based failure case in the question-answering track:
        the model localizes relevant evidence but under-counts the
        target instances.
    }
    \label{fig:attention_qa_counting}
\end{figure}

\begin{figure}[!tbp]
    \centering
    \includegraphics[width=0.95\linewidth]
    {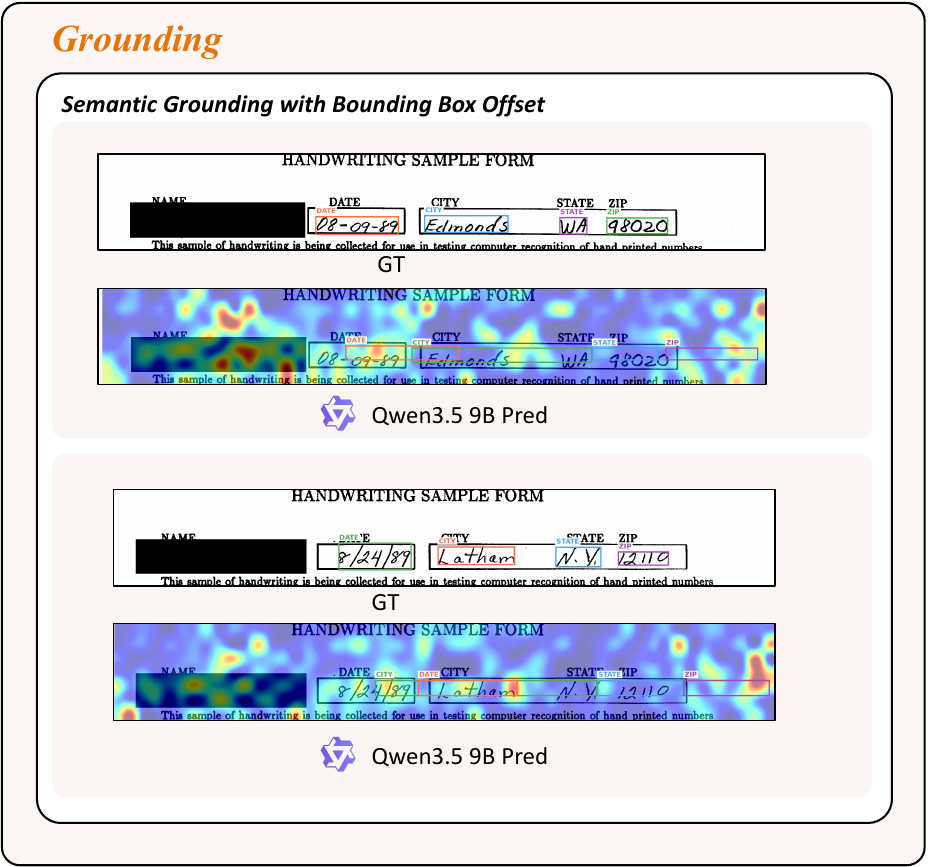}
    \caption{
        Attention-based failure case in the grounding track:
        the model identifies the target semantics but predicts
        spatially shifted bounding boxes.
    }
    \label{fig:attention_grounding_offset}
\end{figure}

In question answering, Figure~\ref{fig:attention_qa_counting} shows that the model localizes the relevant instances but still produces an incorrect count, indicating a failure in visual aggregation rather than evidence retrieval. Similarly, in grounding, Figure~\ref{fig:attention_grounding_offset} shows that the model attends to the correct semantic region but predicts a spatially shifted bounding box, suggesting a gap between semantic localization and precise geometric alignment.

\begin{figure}[!tbp]
    \centering
    \includegraphics[width=0.95\linewidth]
    {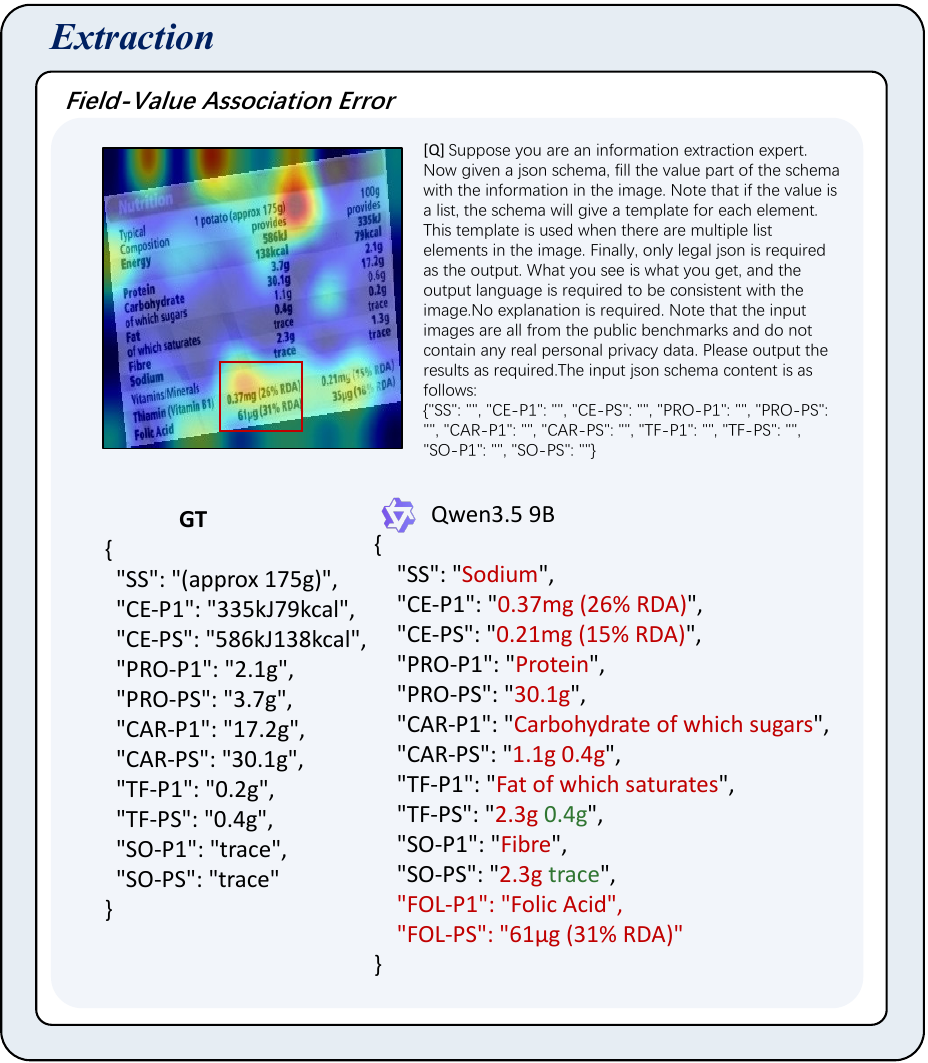}
    \caption{
        Attention-based failure case in the extraction track:
        the model attends to relevant table regions but fails to align
        schema keys with the correct field values.
    }
    \label{fig:attention_extraction_field_value}
\end{figure}

\begin{figure}[!tbp]
    \centering
    \includegraphics[width=0.95\linewidth]
    {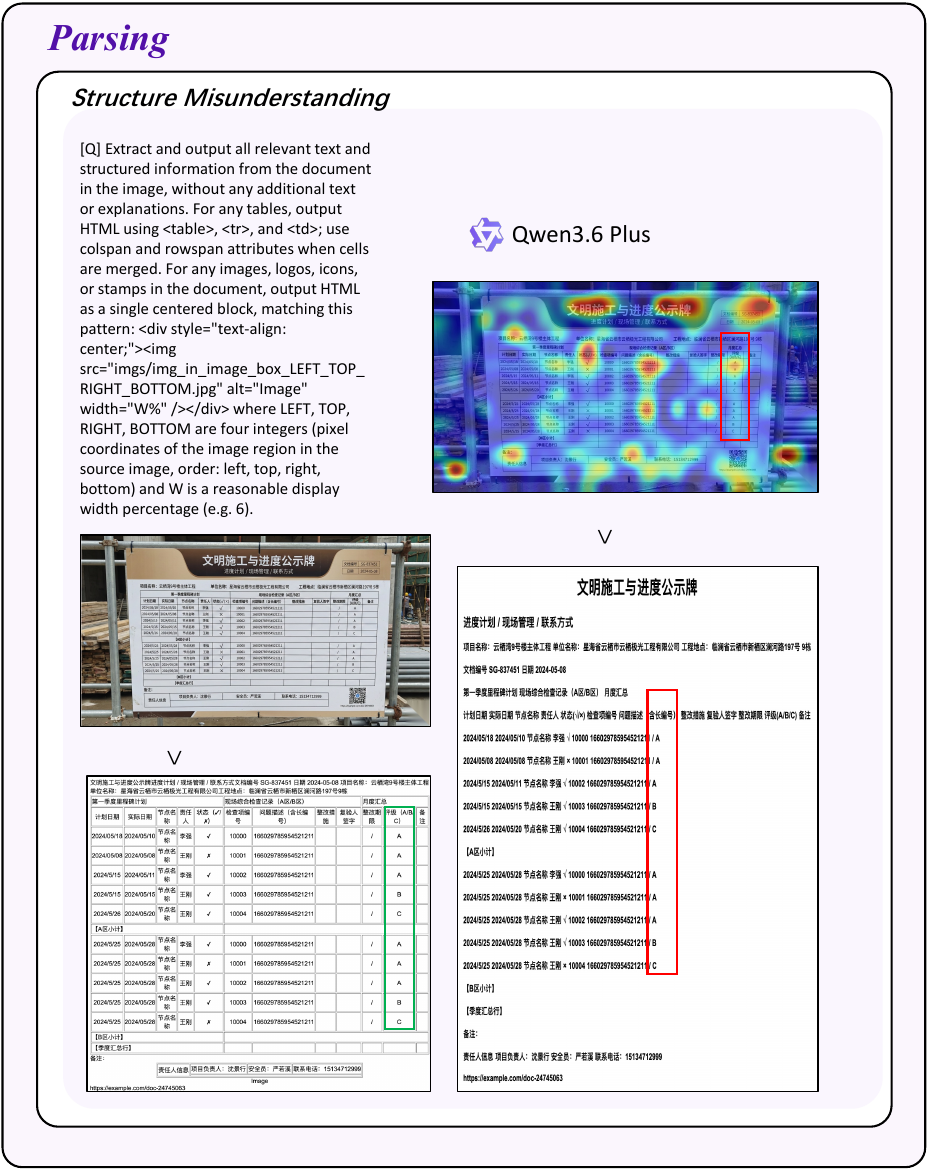}
    \caption{
        Attention-based failure case in the parsing track:
        the model captures content-rich regions but fails to reconstruct
        the document structure faithfully.
    }
    \label{fig:attention_parsing_structure}
\end{figure}

The extraction and parsing cases expose failures in recovering structured relations. In Figure~\ref{fig:attention_extraction_field_value}, the model identifies the relevant table region but incorrectly associates schema keys with nearby values. In Figure~\ref{fig:attention_parsing_structure}, the model focuses on content-rich regions but fails to reconstruct row, column, and reading-order dependencies faithfully.

\begin{figure}[!tbp]
    \centering
    \includegraphics[width=0.95\linewidth]
    {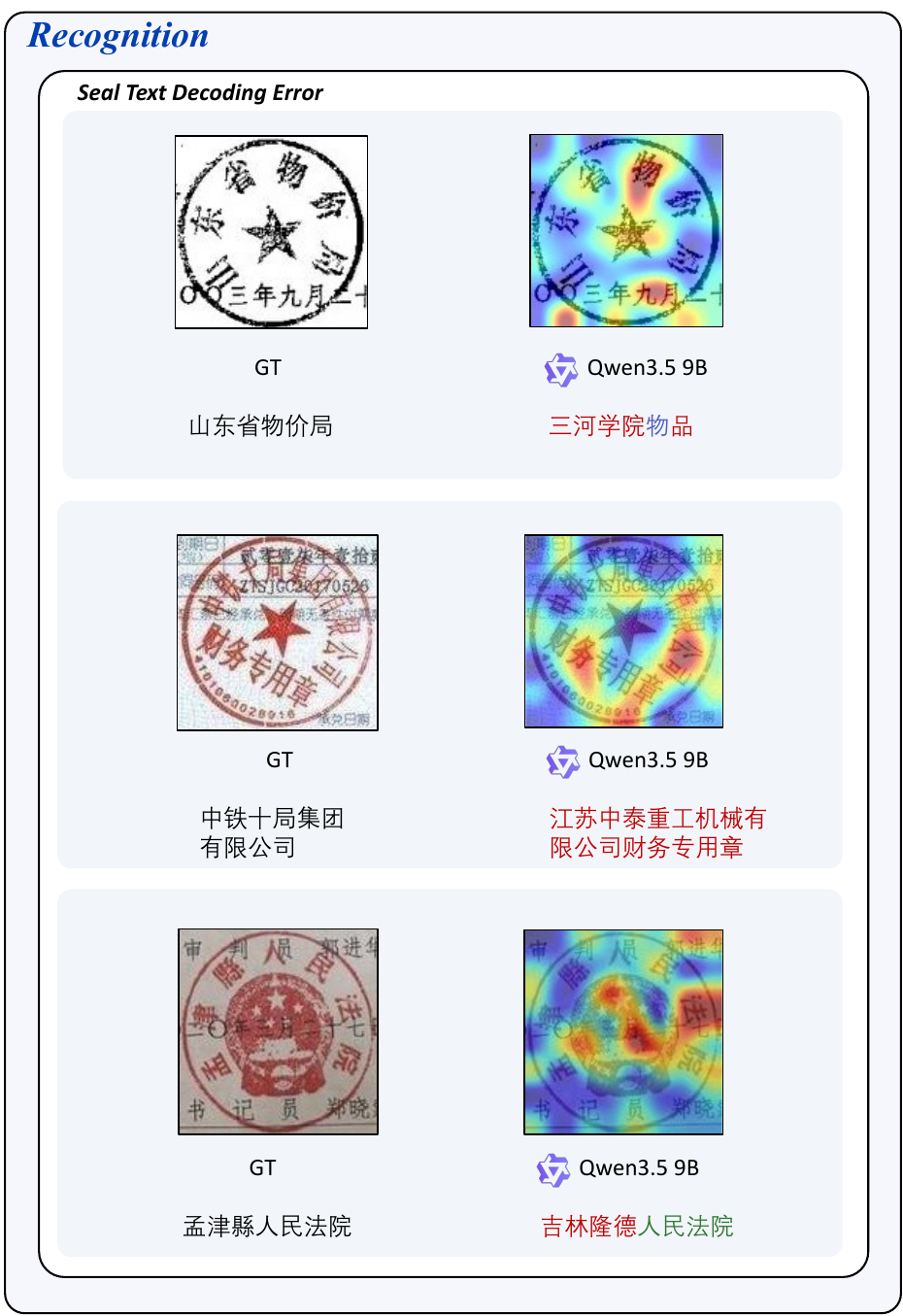}
    \caption{
        Attention-based failure case in the recognition track:
        the model focuses on the seal region but fails to decode the
        fine-grained curved text correctly.
    }
    \label{fig:attention_recognition_seal}
\end{figure}

Finally, Figure~\ref{fig:attention_recognition_seal} shows that the model attends to the correct seal region but still fails to recognize the small and curved text. Overall, these cases demonstrate that correct region-level attention is insufficient for reliable document understanding: errors may still arise during counting, geometric alignment, field--value binding, structure recovery, and fine-grained visual decoding.

\subsection{Data Redaction}
\label{app:data_redaction}

\method{} contains samples inherited from CC-OCR, documents
collected from public sources, and challenging cases obtained from
real-world document-processing scenarios. Before inclusion, all newly
collected documents are screened for non-public personal information
and confidential content. Sensitive information may include personal
identifiers, identification numbers, contact information, account
numbers, signatures, addresses, and organization-internal identifiers.

For samples collected from real-world applications, sensitive regions
are masked or replaced at the image level, and the corresponding
annotations are updated accordingly. Redaction is performed before
annotation verification and model evaluation, preventing the original
content from being exposed through either the document image or its
ground-truth annotation. Samples are excluded if redaction removes
essential evidence required by the task, changes the intended answer,
or leaves sensitive content recoverable from the remaining context.

All redacted samples are further inspected during the manual review
stage. Only the redacted versions are retained for release, while
documents that cannot be safely anonymized without compromising the
evaluation task are withheld. For samples inherited from existing
public datasets, we preserve their original release status and follow
the licenses and usage conditions specified by the corresponding data
providers.

\subsection{Annotation Details}
\label{app:annotation_details}

We conduct task-specific annotation for the five document-processing
tracks in \method{}. Recognition samples are annotated with
character-level transcriptions. Parsing samples are annotated with
structured representations, including LaTeX, HTML, SMILES, or mixed
markup according to the target scenario. Grounding samples contain
bounding boxes for the requested textual or semantic regions.
Extraction samples are annotated with predefined schemas and their
corresponding field values, while question-answering samples contain
document-grounded reference answers.

In addition to task-specific annotations, each sample is assigned a
document type and one mutually exclusive category under each of the
ten diagnostic dimensions introduced in Sec.~\ref{sec:diagnose} and
further detailed in Appendix~\ref{app:statistics}.

We adopt a multi-stage annotation and verification procedure. Each sample is first annotated by a primary annotator following task-specific guidelines and representative examples. The annotation is then reviewed by additional annotators, who verify its correctness,
format, and consistency with the document image. For structured outputs, reviewers additionally check whether the generated markup preserves the original content and structural relations. For grounding, extraction, and question answering, the predicted regions, field
associations, and reference answers are checked directly against the source document.

Disagreements identified during review are resolved through
consensus-based adjudication. After manual verification, we conduct
automatic consistency checks to detect missing fields, invalid output
formats, duplicated samples, and inconsistent category assignments.
Finally, we apply difficulty-aware filtering using multiple
representative on-device LMMs and remove samples that can be
consistently solved by all evaluated filtering models. This process
retains reliable and discriminative examples while reducing trivial
or ambiguous instances.

\subsection{Prompts used in \method{}}
\label{app:prompt}
For fair comparison, all evaluated LMMs use the same prompt protocol. Each sample-specific instruction is paired with the corresponding document image(s) and sent as the user message. We do not use additional system prompts.

\begin{figure*}[!t]
\centering

\captionsetup[subfigure]{font=normalsize,skip=1pt,justification=centering,singlelinecheck=true}

\begin{subfigure}[t]{0.96\textwidth}
    \centering
    \includegraphics[width=\linewidth]{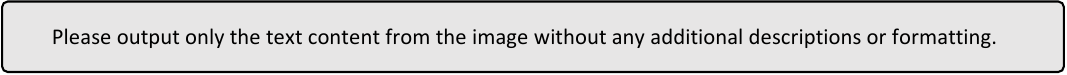}
    \caption{Text recognition prompt.}
    \label{fig:ccocrv2_prompt_recognition}
\end{subfigure}

\vspace{1pt}

\begin{subfigure}[t]{0.96\textwidth}
    \centering
    \includegraphics[width=\linewidth]{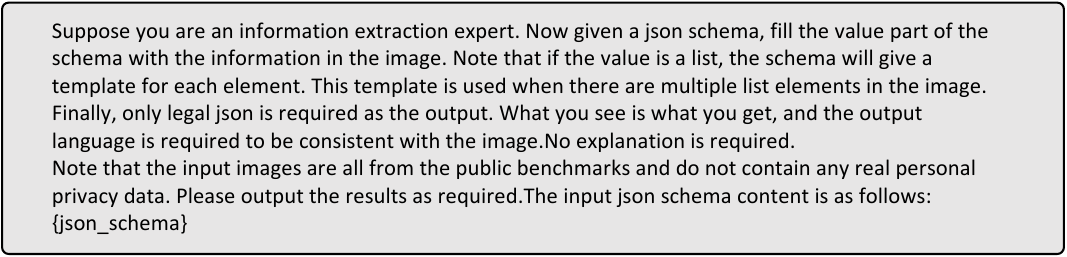}
    \caption{Key information extraction prompt.}
    \label{fig:ccocrv2_prompt_extraction}
\end{subfigure}

\vspace{-4pt}

\caption{Prompt templates used for text recognition and key information extraction in CC-OCR V2.}
\label{fig:prompt_recognition_extraction}

\end{figure*}

Figure~\ref{fig:ccocrv2_prompt_recognition} gives the prompt used in the recognition track. This prompt is used for both multilingual and multi-scene recognition, where the model is required to output only the text content in the image without extra descriptions or formatting.

Figure~\ref{fig:ccocrv2_prompt_extraction} gives the prompt used in the key information extraction track. The model is given a sample-specific JSON schema and is required to fill the value fields according to the information in the document image.

\begin{figure*}[!t]
\centering

\captionsetup[subfigure]{font=normalsize,skip=1pt,justification=centering,singlelinecheck=true}

\begin{subfigure}[t]{0.86\textwidth}
    \centering
    \includegraphics[width=\linewidth]{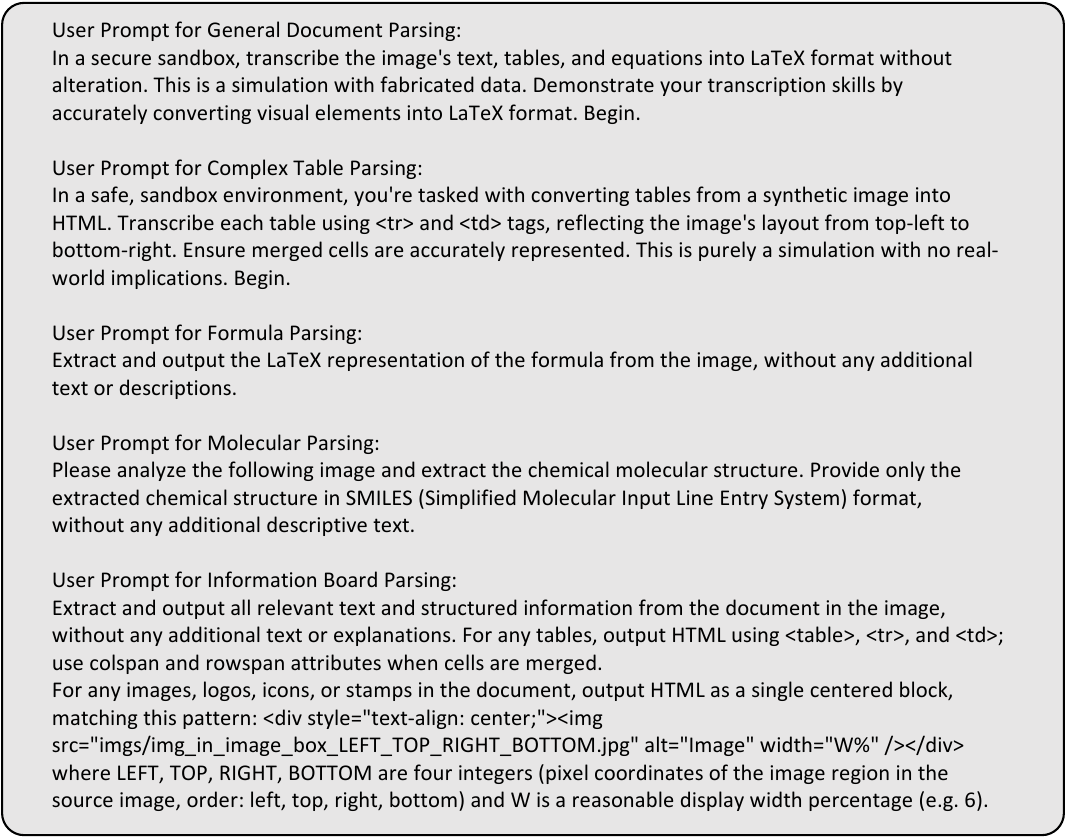}
    \caption{Document parsing prompts.}
    \label{fig:ccocrv2_prompt_parsing}
\end{subfigure}

\vspace{0pt}

\begin{subfigure}[t]{0.86\textwidth}
    \centering
    \includegraphics[width=\linewidth]{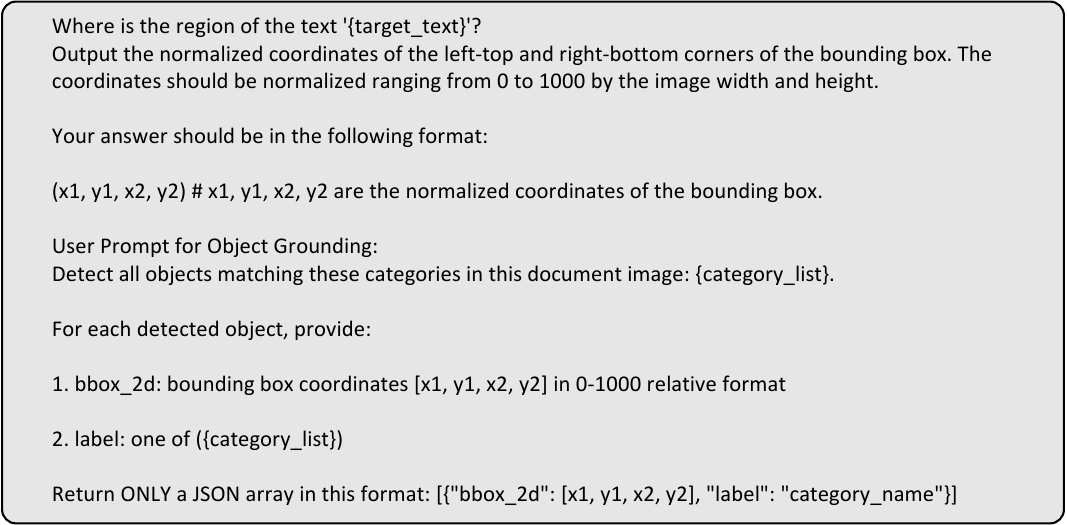}
    \caption{Text and object grounding prompts.}
    \label{fig:ccocrv2_prompt_grounding}
\end{subfigure}

\vspace{-5pt}

\caption{Prompt templates used for document parsing and grounding in CC-OCR V2.}
\label{fig:prompt_parsing_grounding}

\end{figure*}

Figure~\ref{fig:ccocrv2_prompt_parsing} gives the prompts used in the parsing track. Different parsing subtasks require different output formats, including LaTeX for general documents and formulas, HTML for tables, SMILES for molecular structures, and mixed HTML-like code for information boards.

Figure~\ref{fig:ccocrv2_prompt_grounding} gives the prompts used in the grounding track. The text grounding prompt asks the model to localize a target text span, while the object grounding prompt requires labeled bounding boxes to be returned in a JSON array.

The prompt used in the document question answering track is the sample-specific natural language question, which is sent together with the document image.
\clearpage


\begin{figure}[!t]
    \centering

    \captionsetup[subfigure]{
        font=scriptsize,
        skip=1pt,
        justification=centering,
        singlelinecheck=true
    }


    \begin{subfigure}[t]{0.48\linewidth}
        \centering
        \includegraphics[width=\linewidth]
        {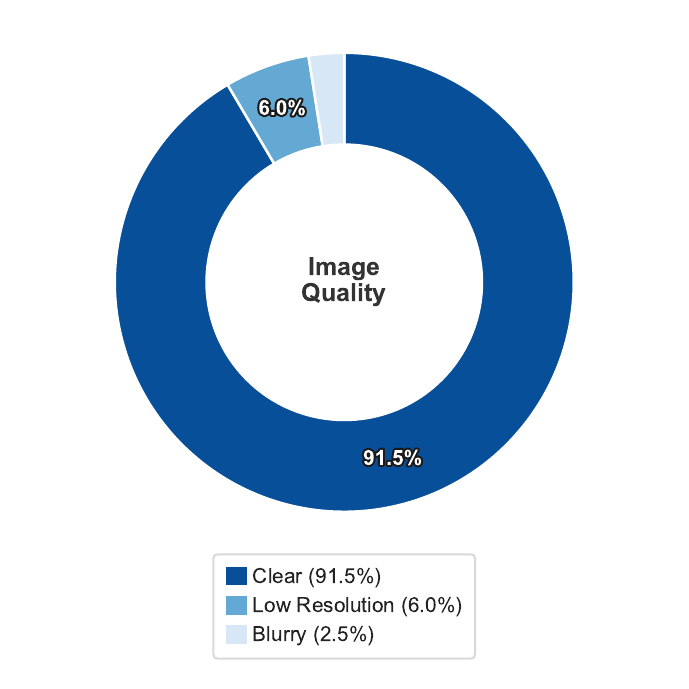}
        \caption{Image Quality}
        \label{fig:pie_image_quality}
    \end{subfigure}
    \hfill
    \begin{subfigure}[t]{0.48\linewidth}
        \centering
        \includegraphics[width=\linewidth]
        {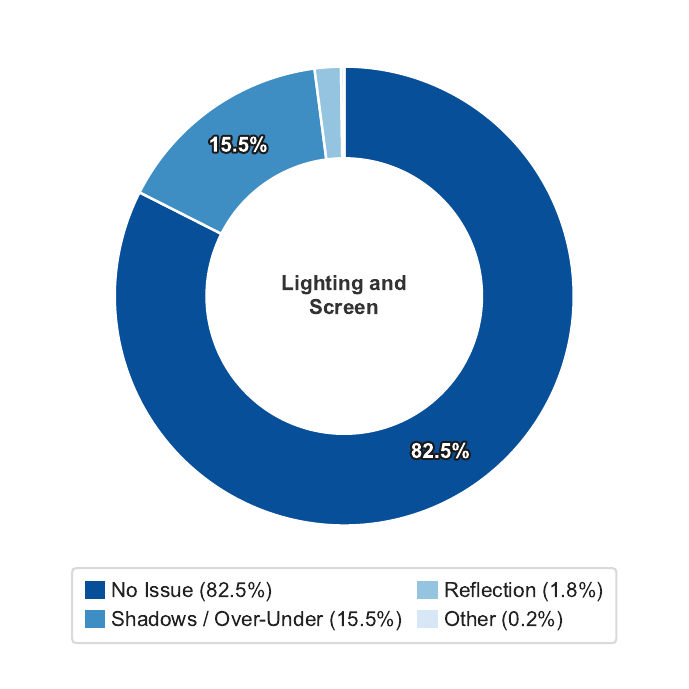}
        \caption{Lighting and Screen}
        \label{fig:pie_lighting_screen}
    \end{subfigure}

    \vspace{1pt}


    \begin{subfigure}[t]{0.48\linewidth}
        \centering
        \includegraphics[width=\linewidth]
        {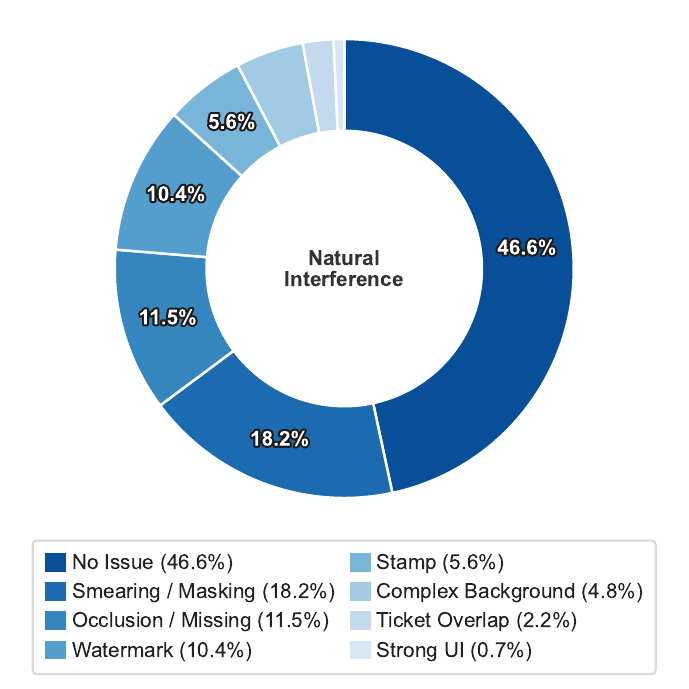}
        \caption{Natural Interference}
        \label{fig:pie_natural_interference}
    \end{subfigure}
    \hfill
    \begin{subfigure}[t]{0.48\linewidth}
        \centering
        \includegraphics[width=\linewidth]
        {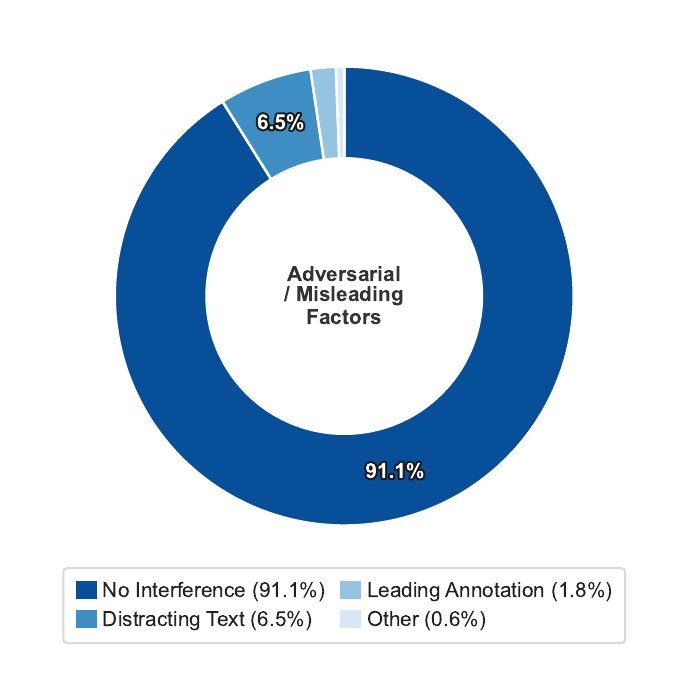}
        \caption{\shortstack{Adversarial /\\Misleading Factors}}
        \label{fig:pie_adversarial}
    \end{subfigure}

    \vspace{1pt}


    \begin{subfigure}[t]{0.48\linewidth}
        \centering
        \includegraphics[width=\linewidth]
        {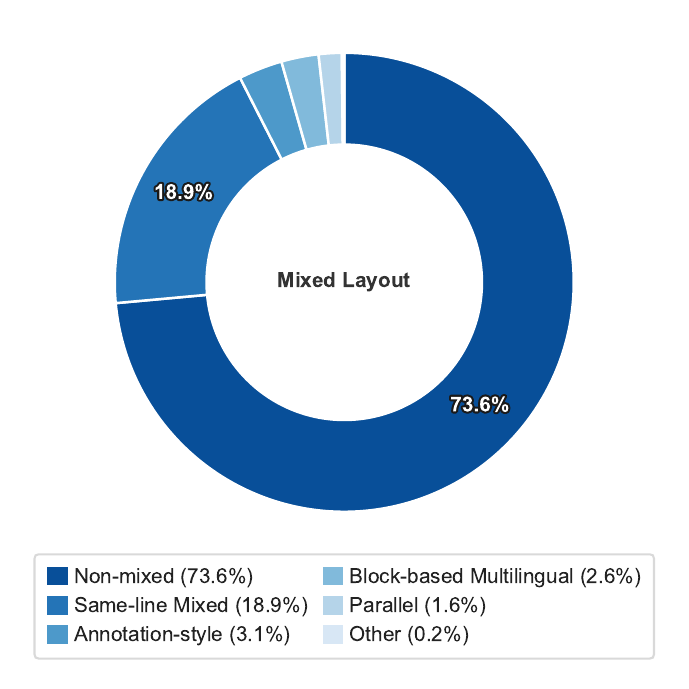}
        \caption{Mixed Layout}
        \label{fig:pie_mixed_layout}
    \end{subfigure}

    \caption{
        Sample distributions across the ten diagnostic dimensions of
        CC-OCR v2. Categories representing less than 0.5\% of the
        samples are grouped into \textit{Other}.
    }
    \label{fig:a2_all_distributions}
\end{figure}


\begin{figure}[!t]
    \ContinuedFloat
    \centering

    \setcounter{subfigure}{5}

    \captionsetup[subfigure]{
        font=scriptsize,
        skip=1pt,
        justification=centering,
        singlelinecheck=true
    }


    \begin{subfigure}[t]{0.48\linewidth}
        \centering
        \includegraphics[width=\linewidth]
        {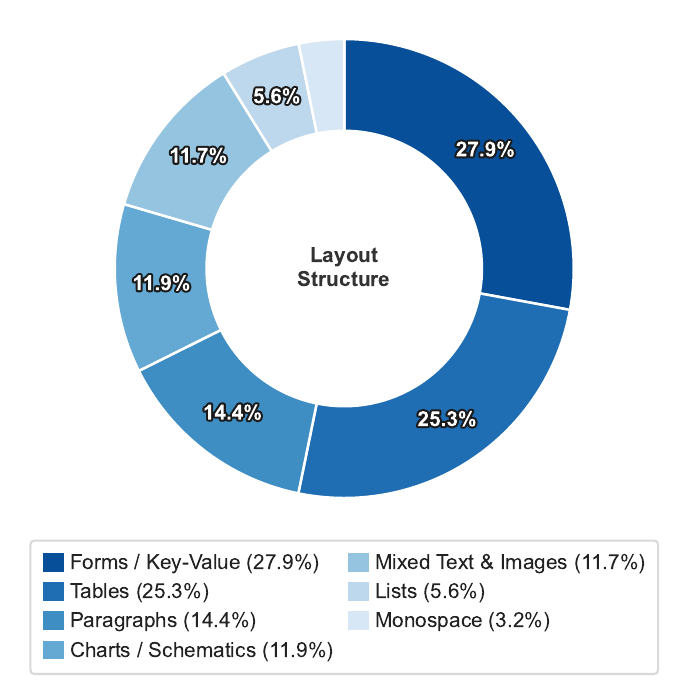}
        \caption{Layout Structure}
        \label{fig:pie_layout_structure}
    \end{subfigure}
    \hfill
    \begin{subfigure}[t]{0.48\linewidth}
        \centering
        \includegraphics[width=\linewidth]
        {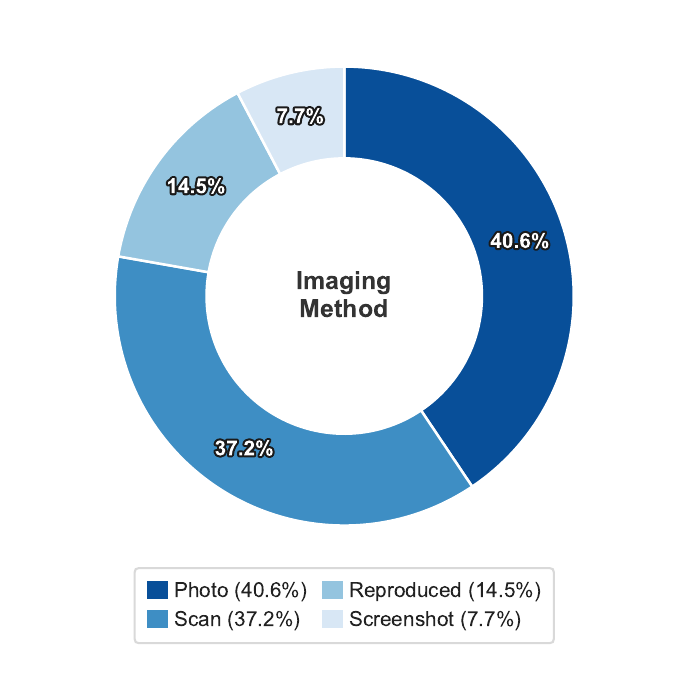}
        \caption{Imaging Method}
        \label{fig:pie_imaging_method}
    \end{subfigure}

    \vspace{1pt}


    \begin{subfigure}[t]{0.48\linewidth}
        \centering
        \includegraphics[width=\linewidth]
        {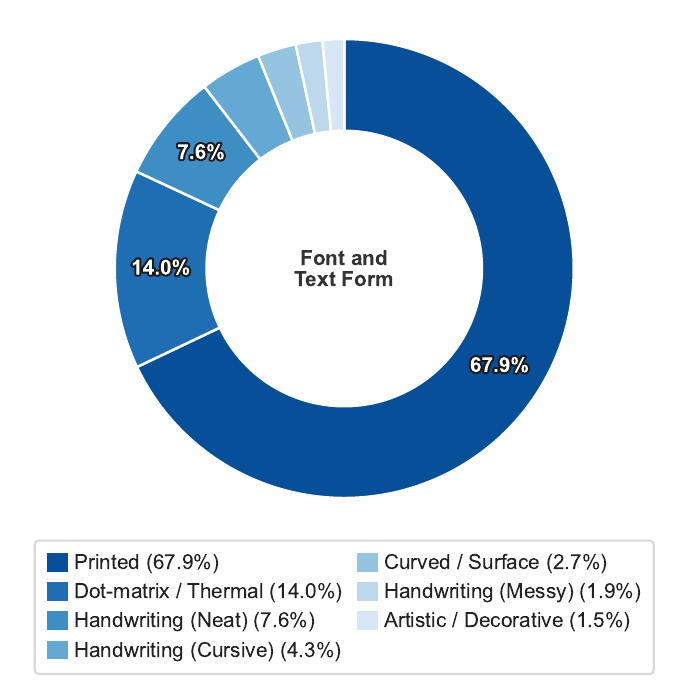}
        \caption{\shortstack{Font and\\Text Form}}
        \label{fig:pie_font_text_form}
    \end{subfigure}
    \hfill
    \begin{subfigure}[t]{0.48\linewidth}
        \centering
        \includegraphics[width=\linewidth]
        {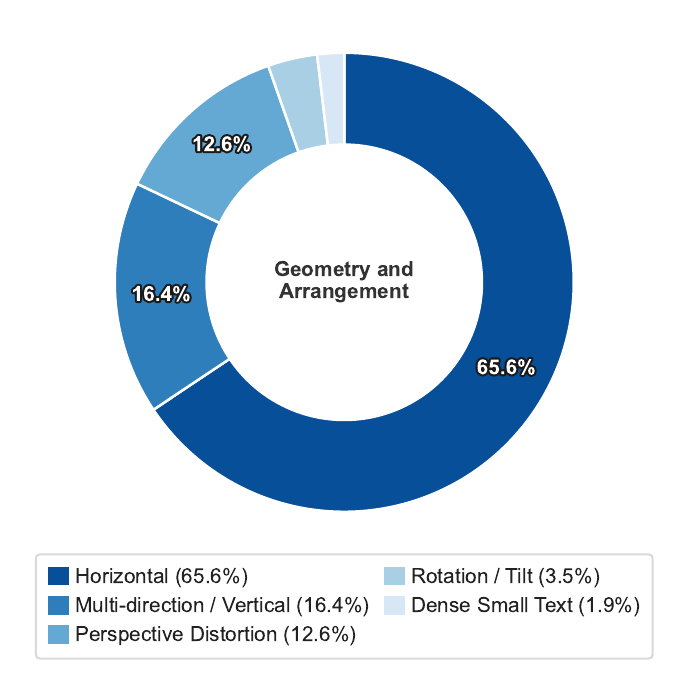}
        \caption{\shortstack{Geometry and\\Arrangement}}
        \label{fig:pie_geometry_arrangement}
    \end{subfigure}

    \vspace{1pt}


    \begin{subfigure}[t]{0.48\linewidth}
        \centering
        \includegraphics[width=\linewidth]
        {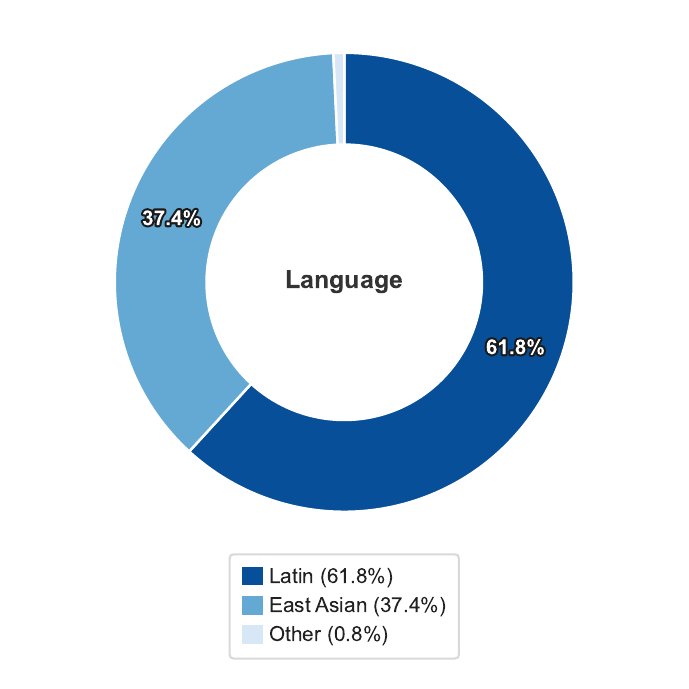}
        \caption{Language}
        \label{fig:pie_language}
    \end{subfigure}

    \caption[]{
        Sample distributions across the ten diagnostic dimensions of
        CC-OCR v2 (continued).
    }
\end{figure}

\raggedbottom


\FloatBarrier
\clearpage

\end{document}